\documentclass[preprint,12pt]{elsarticle}

\usepackage{amsmath, amssymb}

\usepackage{url}            
\usepackage{booktabs}       
\usepackage{amsfonts}       
\usepackage{nicefrac}       
\usepackage{microtype}      

\usepackage[usenames,dvipsnames]{xcolor}
\usepackage{multirow}
\usepackage{graphicx}
\usepackage{bm}
\usepackage{bbm}
\usepackage{subfig}
\usepackage{wrapfig}
\usepackage{caption}
\usepackage{amsfonts}
\usepackage{algorithm}
\usepackage{listings}

\usepackage[T1]{fontenc}
\usepackage{lmodern}

\newcommand{\xmark}{$\times$}
\newcommand{\ie}{\textit{i.e. }}

\journal{PR for Review}

\begin{document}

\begin{frontmatter}



\title{On the Pros and Cons of Momentum Encoder in Self-Supervised Visual Representation Learning}


\author[a1]{Trung Pham} \ead{trungpx@kaist.ac.kr}
\author[a1]{Chaoning Zhang} 
\author[a2]{Axi Niu}

\author[a1]{Kang Zhang}
\author[a1]{Chang D. Yoo \corref{mycorrespondingauthor}} \ead{cd_yoo@kaist.ac.kr}
%
            
\cortext[mycorrespondingauthor]{Corresponding author: Chang D. Yoo. We are with Northwestern Polytechnical University and KAIST.}

\begin{abstract}
Exponential Moving Average (EMA or momentum) is widely used in modern self-supervised learning (SSL) approaches, such as MoCo, for enhancing performance. We demonstrate that such momentum can also be plugged into momentum-free SSL frameworks, such as SimCLR, for a performance boost. Despite its wide use as a basic component in the modern SSL frameworks, the benefit caused by momentum is not well understood. We find that its success can be at least partly attributed to the stability effect. In the first attempt, we analyze how EMA affects each part of the encoder and reveal that the portion near the encoder's input plays an insignificant role while the latter parts have much more influence. By monitoring the gradient of the overall loss with respect to the output of each block in the encoder, we observe that the final layers tend to fluctuate much more than other layers during backpropagation, i.e. less stability. Interestingly, we show that using EMA to the final part of the SSL encoder, i.e. projector, instead of the whole deep network encoder can give comparable or preferable performance. Our proposed projector-only momentum helps maintain the benefit of EMA but avoids the double forward computation.
\end{abstract}




\end{frontmatter}


\section{Introduction}
\label{intro}
Self-supervised learning (SSL) is known for its capability to learn representation with unlabeled data. SSL has been widely applied in NLP \cite{brown2020language,Lan2020ALBERT,radford2019language,su2020vlbert,nie2020dc} and computer vision \cite{LIU2022108767,ZHANG2022108234,komodakis2018unsupervised,oord2018representation,he2020momentum,chen2020simple,madaan2022representational,Vasudeva_2021_ICCV,zhang2022how,zhang2022dual, BAYKAL2022108244}. We find that multiple seminal SSL frameworks, MoCo \cite{he2020momentum,chen2020improved,chen2021mocov3}, BYOL \cite{grill2020bootstrap}, DINO \cite{caron2021emerging}, and ReSSL \cite{zheng2021ressl} all use momentum to form a teacher-student paradigm where the teacher encoder is updated from the student model with exponential moving average (EMA). To avoid any confusion, we use either ``EMA'' or ``momentum'' interchangeably when mentioning the EMA \cite{tarvainen2017mean}.

In the literature, EMA encoder has been motivated mainly for two purposes: (a) to generate consistent representations for negative pairs \cite{he2020momentum,chen2020improved,zheng2021ressl}; and (b) avoid collapse in negative-free SSL frameworks \cite{caron2021emerging,grill2020bootstrap}. These motivations have been challenged by the claims in follow-up works~\cite{zhang2022dual,chen2020simple}. For example, a large number of negative samples might not be necessary, rendering the EMA unnecessary from the perspective of negative samples \cite{zhang2022dual}. SimSiam \cite{chen2020simple} also shows that EMA is not necessary for preventing a collapse in the negative-free SSL frameworks. This motivates a need for understanding EMA from a new perspective for explaining its empirical benefit in boosting performance.

We investigate and find that such a performance boost can be partly explained by the stability effect. In practice, unfortunately, this benefit of the stability effect does not come free in the sense that it doubles the forward computation resources. Note that the core of SSL lies in learning an augmentation-invariant representation, thus the optimization goal is often designed to make two images augmented from the same image have a similar representation. Specifically, the representation of another image augmented from the same image is used as the target for optimizing the network. Applying momentum on the teacher encoder requires additional forward computation resources as well as memory for obtaining the target representation.

To this end, we investigate the influence of momentum on different parts of the encoder. Specifically, we disentangle the momentum in the teacher network encoder into different parts and investigate the effect of momentum for each stage (block) of the whole encoder network. We find that the benefit of momentum differs massively at different stages of the encoder. The EMA in the early stages (close to the input) has an insignificant role, while that in the latter stages (close to the output) is much more significant. We find that this can be attributed to much higher yet fluctuating gradient values which cause more instability.

Interestingly, we find that only applying EMA to the final \textit{projector} of the encoder is sufficient enough to yield a comparable or sometimes preferable performance, especially when its momentum coefficient is set to be more aggressive. A merit of applying EMA only on the \textit{projector} is that the forward propagation on the backbone can be shared, which only causes double forward propagation on the light \textit{projector}. 

\begin{figure}[!htbp]
  \centering
  \subfloat[EMA with whole network (BYOL \cite{grill2020bootstrap})] {\includegraphics[width=0.49\linewidth]{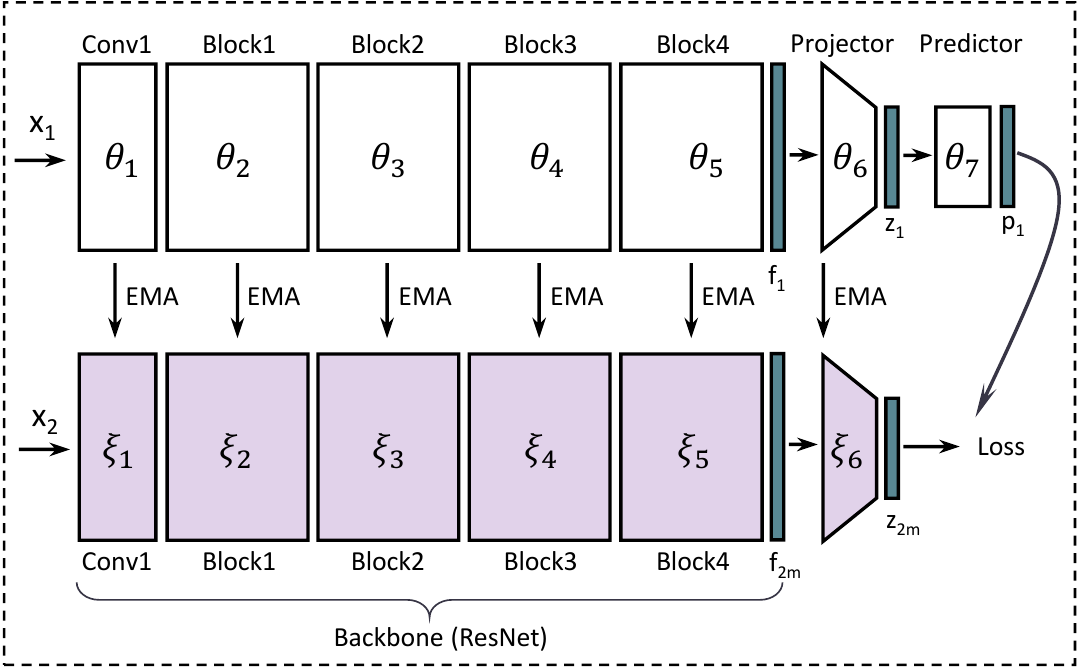}}
  \hfill
  \subfloat[EMA with projector-only (Ours)] {\includegraphics[width=0.49\linewidth]{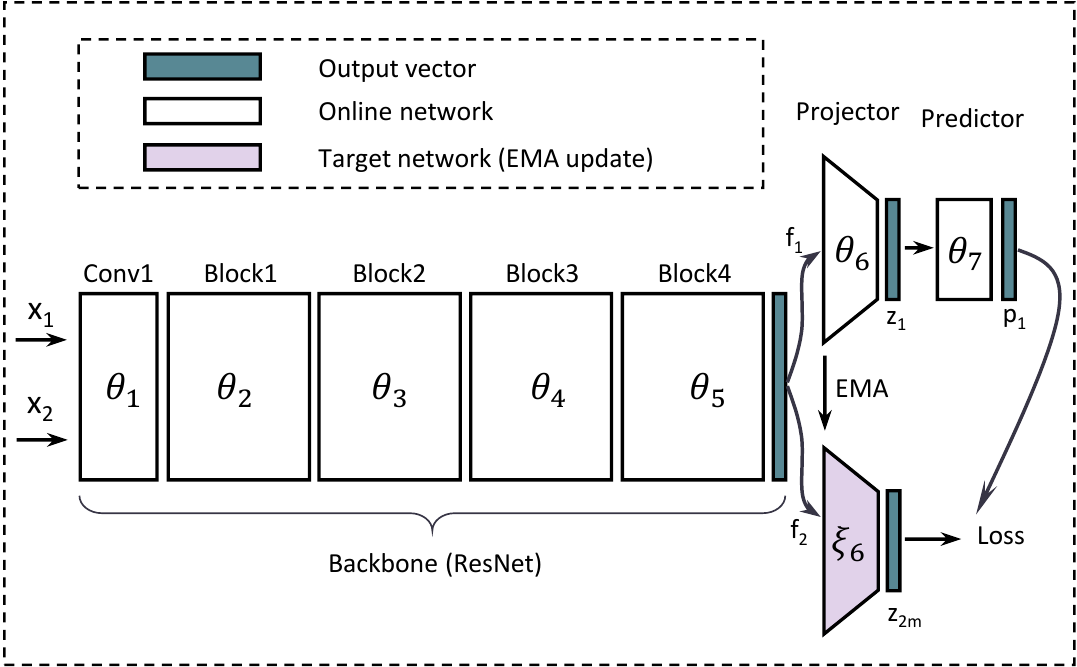}}
  \caption{Comparing traditional EMA updates to the whole target network (a) and EMA updates on a final part of the target network (b). We investigate the EMA effect on each separate part of the entire encoder. Fig. b) shows the case when EMA is applied on the \textit{projector}, all other parts are kept sharing for \textit{online} and \textit{target} network. Here, $x_1$ and $x_2$ are two random augmentations of the same image.}
  \label{fig:ema_whole_part}
\end{figure}

Compared with the forward computation on the \textit{backbone}, the computation overhead on the \textit{projector} is negligible. Therefore, in this work, we suggest projector-only momentum as Fig. \ref{fig:ema_whole_part}b) for maintaining the benefit of EMA but avoiding the double forward computation overhead. To the best of our knowledge, we are the first to investigate the impact of the momentum encoder given the recent progress of SSL.

Overall, the contributions of our work are summarized as follows:
\begin{enumerate}    
    \item Recognizing that the existing motivations for introducing momentum into the SSL frameworks are refuted by follow-up works, we investigate with experiments and show that its empirical benefit for boosting performance can be partly attributed to the stability effect.
    
    \item We explore and reveal an intriguing observation that the benefit of applying momentum in the later stages of the encoder is significantly higher than that in the early stage. Moreover, we provide a gradient analysis to justify this phenomenon.
    
    \item We note that the benefit of momentum comes at the cost of causing double forward computation. Motivated by the above observation, we propose a projector-only momentum for maintaining the pros of momentum while mitigating its cons.
\end{enumerate}

\section{Background}
\subsection{Full EMA in Self-Supervised Learning}
Given an unlabeled image $\mathbf{x}^i$, two random augmentations $\mathbf{x}^i_1$ and $\mathbf{x}^i_2$ are generated and formed as a positive pair. We consider an \textit{encoder} $E_{\theta}$ which includes \textit{backbone} $b(.)$ and an MLP \textit{projector} $g(.)$ for SSL frameworks in general \cite{chen2020simple,caron2021emerging,he2020momentum,chen2021exploring,caron2020unsupervised,zheng2021ressl}, and plus an additional MLP \textit{predictor} $h(.)$ for BYOL \cite{grill2020bootstrap} or SimSiam \cite{chen2021exploring} as illustrated in Fig. \ref{fig:ema_whole_part}a).
With the input image $\mathbf{x}_1$ and $\mathbf{x}_2$, the representations after backbone $b(.)$ are as follows:
\begin{equation}
    f_1 = b(x_1), f_2 = b(x_2).
\end{equation}
And outputs for the momentum backbone $b_m(.)$ are computed by:
\begin{equation}
    f_{1m} = b_m(x_1), f_{2m} = b_m(x_2).
\end{equation}
Note that, subscript `m' denotes the momentum version. Similarly, we feed $f_1,f_2,f_{1m},f_{2m}$ to projector $g(.),g_m(.)$ to have the projected outputs:
\begin{equation}
    z_{1} = g(f_1), z_{2} = g(f_2), z_{1m} = g_m(f_{1m}), z_{2m} = g_m(f_{2m}). 
\end{equation}
Finally, the predictor is applied to the projections of the online network $h(.)$ only, \ie no momentum version for the predictor:
\begin{equation}
    p_1 = h(z_1), p_2 = h(z_2).
\end{equation}
For other SSL frameworks such as Moco, DINO, and ReSSL the projections $z_1,z_2$ will be used for the objective function. Here BYOL loss is taken into account for example, where additional predictor outputs are used for computing the objective loss.
With the distance between the \textit{anchor} $p^i_1$ (or $p^i_2$) and the \textit{target} $z^{i}_{2m}$ (or $z^{i}_{1m}$) measured by \textit{cosine similarity} with operator ($\cdot$), the objective function of BYOL \cite{grill2020bootstrap} within a mini-batch $N$ samples is defined as follows:
\begin{equation}
    \label{eq:byol_loss}
    \mathcal{L} = - \frac{1}{2} ( \frac{1}{N} \sum_{i=1}^{N} (p^i_1 \cdot z^{i}_{2m}) + \frac{1}{N} \sum_{i=1}^{N} (p^i_2 \cdot z^{i}_{1m}) ),
\end{equation}
where $p^i_1$ and $z^{i}_{1m}$ denote the output of the predictor $h(.)$ and output at the momentum projector $g(.)$ of the augmented image $\mathbf{x}^i_1$, respectively. In the same way, $p^i_2$ and $z^{i}_{2m}$ denote the notations for the second augmented image $\mathbf{x}^i_2$. Note that here each augmented image is fed into the full momentum encoder, \ie both momentum \textit{backbone ($b_m$)} and \textit{projector ($g_m$)}.

It is also worth noticing that all $p_1^i,p_2^i,z^i_{1m},z^i_{2m}$ are $l_2$-normalized, and BYOL uses a symmetric loss as default. It is also worth to note that $z^{i}_{1m}$ and $z^{i}_{2m}$ here are the outputs of the \textit{momentum} encoder in BYOL (or stop gradient in SimSiam \cite{chen2021exploring}), the subscript $m$ is put beside index $i$ to indicate it is momentum output. 

Obviously, the BYOL loss only allows the gradient through $p^i_1$ and $p^i_2$, not $z^i_1$ and $z^i_2$. Follow \cite{grill2020bootstrap}, we term \textit{online} network for the branch that allows the gradient update \cite{NEURIPS2019_pytorch} and \textit{target} network for the branch that is updated with EMA. It can also be referred to as \textit{student} and \textit{teacher} network \cite{caron2021emerging}, respectively. 
The \textit{online} network is specified by the parameters $\theta$ which allows updating gradient with back-propagation during training \cite{NEURIPS2019_pytorch}, and the \textit{target} network is modeled by the parameters $\xi$. The momentum update rule for the parameters $\xi$ as follows \cite{grill2020bootstrap}:
\begin{equation}
    \label{eq:ema_update}
    \xi \leftarrow \beta\xi + (1-\beta)\theta,
\end{equation}
where a constant $\beta\in[0,1]$ is the momentum coefficient, which is often chosen with value of 0.99 for short training (200 epochs). In longer training, \ie 1000 epochs, SSL methods often use the higher momentum value, \ie $\beta=0.996$ for BYOL \cite{grill2020bootstrap,chen2021exploring} and DINO \cite{caron2021emerging}, or $\beta=0.999$ for MoCo \cite{he2020momentum,chen2020improved}. 

\subsection{EMA Projector in Self-Supervised Learning}
In this new setting, instead of using backbone momentum outputs $f_{1m},f_{2m}$, the outputs of the online backbone $f_{1},f_{2}$ are used to feed into the momentum projector $g_m(.)$:
\begin{equation}
    z_{1m} = g_m(f_1), z_{2m} = g_m(f_2).
\end{equation}
The final objective loss formula is kept the same as the baselines, \ie Eq. \ref{eq:byol_loss}. For the other frameworks, the \textit{EMA projector} is performed the same as described above.

\section{Towards Understanding Momentum in SSL}
\subsection{Momentum Brings Performance Boost}
We bring four widely used momentum-based SSL methods including BYOL \cite{grill2020bootstrap}, ReSSL \cite{zheng2021ressl}, MoCo v2 \cite{chen2020improved}, and DINO \cite{caron2021emerging} to conduct comprehensive experiments to analyse the EMA encoder effect. Furthermore, to generalize our findings, we also inject the momentum on two non-momentum-based frameworks, SimCLR \cite{chen2020simple} and SimSiam \cite{chen2021exploring}.

We conduct two experiments to compare the influence of momentum on the quality of the representations learned by the SSL encoders. \textit{First}, the experiments are conducted for each method that uses the momentum with optimal parameters, and \textit{second}, we remove the momentum by setting the momentum coefficient $\beta = 0$ given \textit{stop gradient} \cite{chen2021exploring} which means sharing the weight between \textit{online} and \textit{target} network.

We measure the performance using top-1 accuracy (\%) on the test set as the common protocol in SSL, \ie BYOL \cite{grill2020bootstrap}. Tab. \ref{tab:effect_of_momentum} shows that without momentum encoder, all momentum-based methods dramatically drop performance (BYOL, DINO, MoCo v2, and ReSSL). Besides, non-momentum-based frameworks, \ie SimSiam and SimCLR get improvement with the existence of momentum. 

Specifically, when the momentum of the \textit{target} network is removed, \ie $\beta=0$, MoCo v2, DINO, and ReSSL are not able to train, their accuracy is trivial ($\sim 2\%$, not better than a random guess) while BYOL can be able to train but with the visible degradation on accuracy (43.48\%) comparing to the baseline that uses EMA (62.01\%).
\begin{table}[!htbp]
    \caption{Effect of EMA in various SSL frameworks. We use ResNet-18 on CIFAR-100, train for 200 epochs. Plugging momentum consistently improves the performance compared to the version without it. We report top-1 accuracy. Notations, CE: cross-entropy, CL: contrastive learning.} 
    \label{tab:effect_of_momentum}
    \smallskip
    \centering
    \resizebox{1.0\hsize}{!}{
    \begin{tabular}{ccccccc}
    \toprule
    Method & Architecture / Loss & Momentum & Coefficient $\beta$ & Remarks & Acc (\%) & $\Delta_{\text{acc}}$ (\%)  \\
    \noalign{\smallskip}
    \midrule
    BYOL \cite{grill2020bootstrap} & \multirow{2}{*}{Asymmetric / Non-CL} & \checkmark & $\beta=0.99$ & Baseline & \textbf{62.01} & - \\
    - &  & \xmark & $\beta=0$ & Remove & \textcolor{red}{43.48} & \textcolor{red}{$-18.53$} \\
    \midrule
    SimSiam \cite{chen2021exploring} & \multirow{2}{*}{Asymmetric / Non-CL} & \xmark & $\beta=0$ & Baseline & 51.41 & - \\
    - &  & \checkmark & $\beta=0.99$ & Add & \textbf{53.66} & \textcolor{blue}{$+2.25$} \\
    \midrule
    DINO \cite{caron2021emerging} & \multirow{2}{*}{Symmetric / CE} & \checkmark & $\beta=0.99$ & Baseline & \textbf{58.12} & - \\
    - &  & \xmark & $\beta=0$ & Remove & \textcolor{red}{1.08} & \textcolor{red}{$-57.04$} \\
    \midrule
    ReSSL \cite{zheng2021ressl} & \multirow{2}{*}{Symmetric / CL} & \checkmark & $\beta=0.99$ & Baseline & \textbf{54.66} & - \\
    - &  & \xmark & $\beta=0$ & Remove & \textcolor{red}{1.56} & \textcolor{red}{$-53.10$} \\
    \midrule
    MoCo v2 \cite{chen2020improved} & \multirow{2}{*}{Symmetric / CL} & \checkmark & $\beta=0.99$ & Baseline & \textbf{62.34} & - \\
    - &  & \xmark & $\beta=0$ & Remove & \textcolor{red}{2.22} & \textcolor{red}{$-60.12$} \\
    \midrule
    SimCLR \cite{chen2020simple} & \multirow{2}{*}{Symmetric / CL} & \xmark & $\beta=0$ & Baseline & 60.45 & - \\
    - &  & \checkmark & $\beta=0.90$ & Add & \textbf{61.97} & \textcolor{blue}{$+1.52$} \\
    \bottomrule
    \end{tabular}}
\end{table}

Note that, BYOL without momentum can be considered same as SimSiam framework given some different configurations in \textit{projector} and \textit{predictor} \cite{chen2021exploring}. With a careful design, SimSiam \cite{chen2021exploring} presented a special case of BYOL where $\beta=0$ that works fine however it cannot be comparable with BYOL in performance \cite{grill2020bootstrap,chen2021exploring}.
We ran SimSiam using \cite{turrisi2021sololearn} and found its accuracy is 51.41\% which significantly outperforms the version without momentum in the BYOL framework whose accuracy is 43.48\% (Tab. \ref{tab:effect_of_momentum}). However, compared to the BYOL baseline that uses momentum (accuracy 62.01\%), SimSiam is far lower to catch up.
\textit{By default}, SimSiam \cite{chen2021exploring} and SimCLR \cite{chen2020simple} do not have EMA in its architecture. We performed experiments by adding EMA to their \textit{target} encoder and find that EMA boosts SimSiam from 51.41\% to \textbf{53.66\%} ($+2.25$\%) and improves SimCLR from 60.45\% to \textbf{61.97\%} ($+1.52$\%) as shown in Tab. \ref{tab:effect_of_momentum}.

The analysis in this section demonstrates the significance of EMA in guaranteeing learning and boosting the performance of a wide range of network architectures. But, \textit{how can EMA help in SSL?}

\subsection{How Does Momentum Improve Performance?}
\textbf{Prior motivations.} The use of EMA in the CL-based frameworks \cite{he2020momentum,chen2020improved,zheng2021ressl} is for creating consistent negative-pairs representations, and avoiding collapsing in the Non-CL-based frameworks \cite{caron2021emerging,grill2020bootstrap}. Their motivations have been refuted by the follow-up works \cite{zhang2022dual,chen2020simple} as the earlier discussion. This motivates us with a new perspective for the empirical benefit of EMA as the \textit{hypothesis} below.

\textbf{Hypothesis.} Momentum helps the encoder learn from the past to make the training more stable.

\textbf{Empirical Analysis.} To demystify the hypothesis that momentum can help to stabilize the training by learning from the past, we monitor and visualize the \textit{student} and \textit{teacher} weight when using momentum compared to the weight that is not trained with momentum. BYOL \cite{grill2020bootstrap} is chosen for experiments. We use the ResNet-18 backbone, CIFAR-100, to train in 200 epochs. By default, with BYOL is implemented in \cite{turrisi2021sololearn}, the \textit{projector} in the encoder has two FC layers: FC1(512,4096), BN, ReLU, FC2(4096,256). To be able to visualize the weight, we design the final FC layer of the encoder in BYOL to have \textit{two dimensions}. To this end, we change the hidden layer from 4096 to 2, \ie FC1(512,2), BN, ReLU, FC2(2,256), and all other settings are kept intact.

\begin{figure}[!htbp]
  \vspace{-12pt}
  \centering
  \subfloat[Filter 60] {\includegraphics[width=0.33\linewidth]{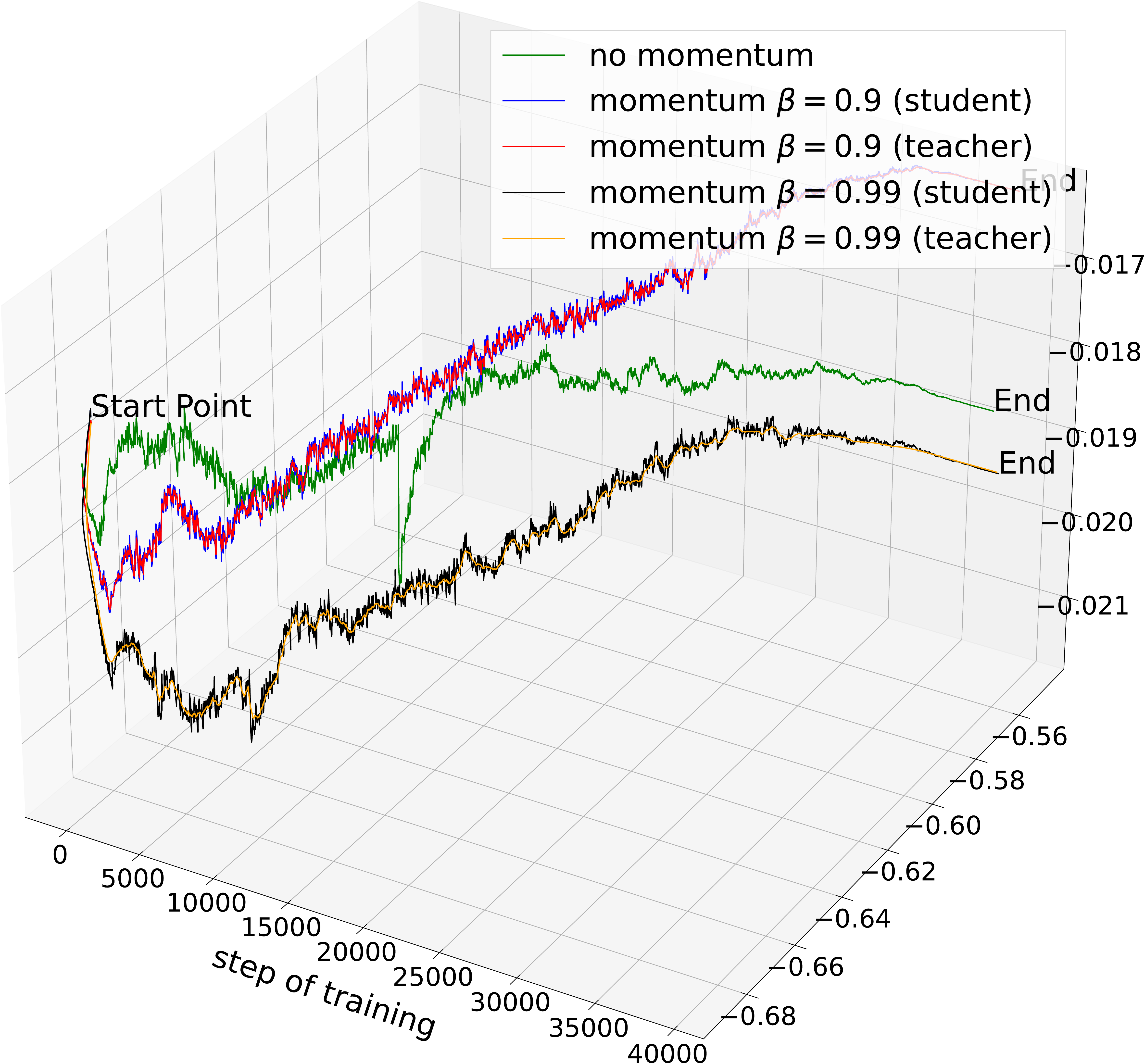}}
  \hfill
  \subfloat[Filter 103] {\includegraphics[width=0.33\linewidth]{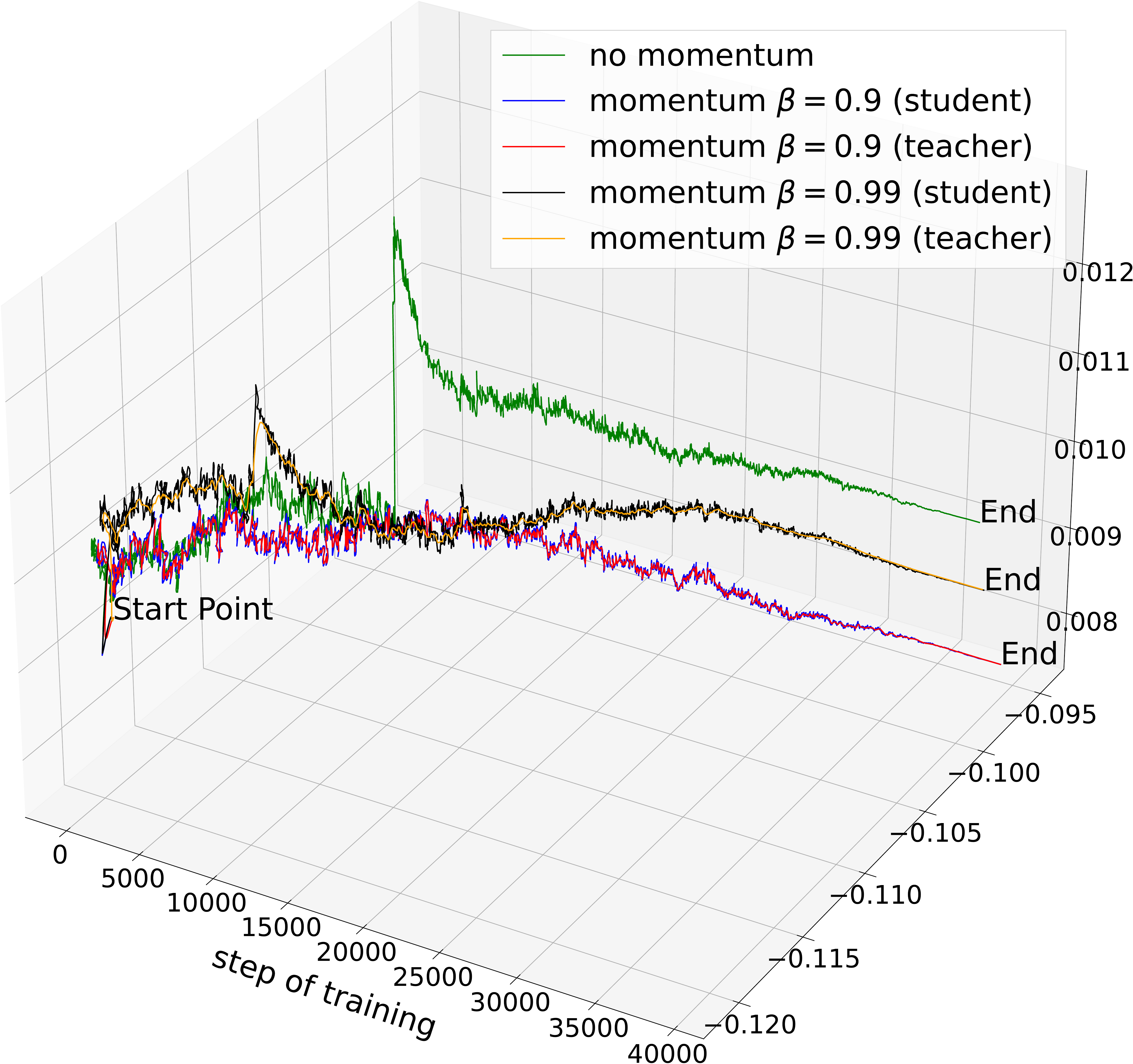}}
  \hfill
  \subfloat[Filter 133] {\includegraphics[width=0.33\linewidth]{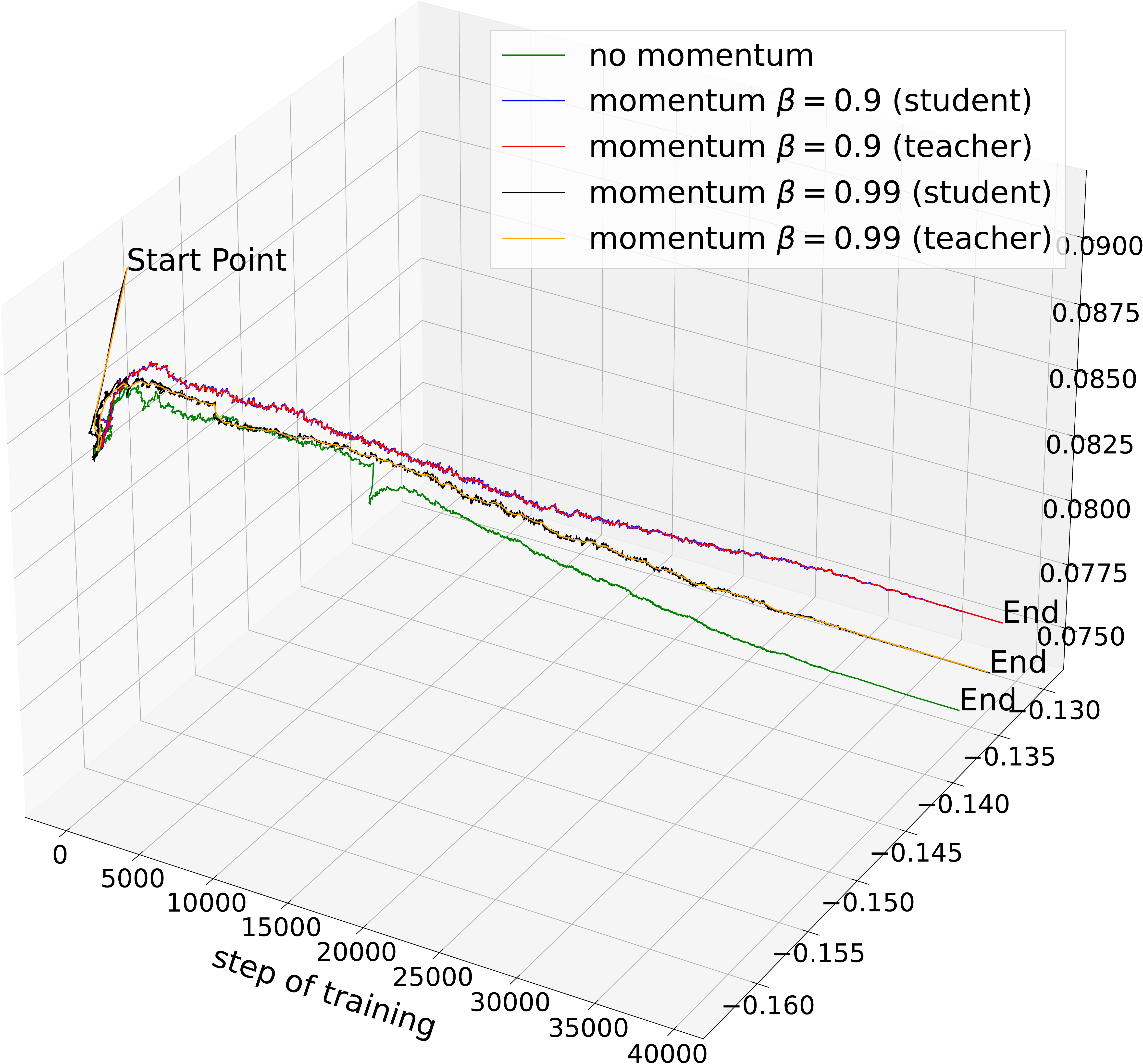}}
  \caption{Effect of the momentum to the weights (filters) of FC2. Filters are visualized in 3D space. Without momentum, we observe the ``abnormal point'' at step around 10k out of 40k (\textcolor{Green}{\textit{green line}}). The axis y, and z are weight 1 and weight 2, respectively.}
  \label{fig:filter3d}
  \vspace{-10pt}
\end{figure}
\begin{figure}[!htbp]
  \vspace{-12pt}
  \centering
  \subfloat[Linear train accuracy] {\includegraphics[width=0.33\linewidth]{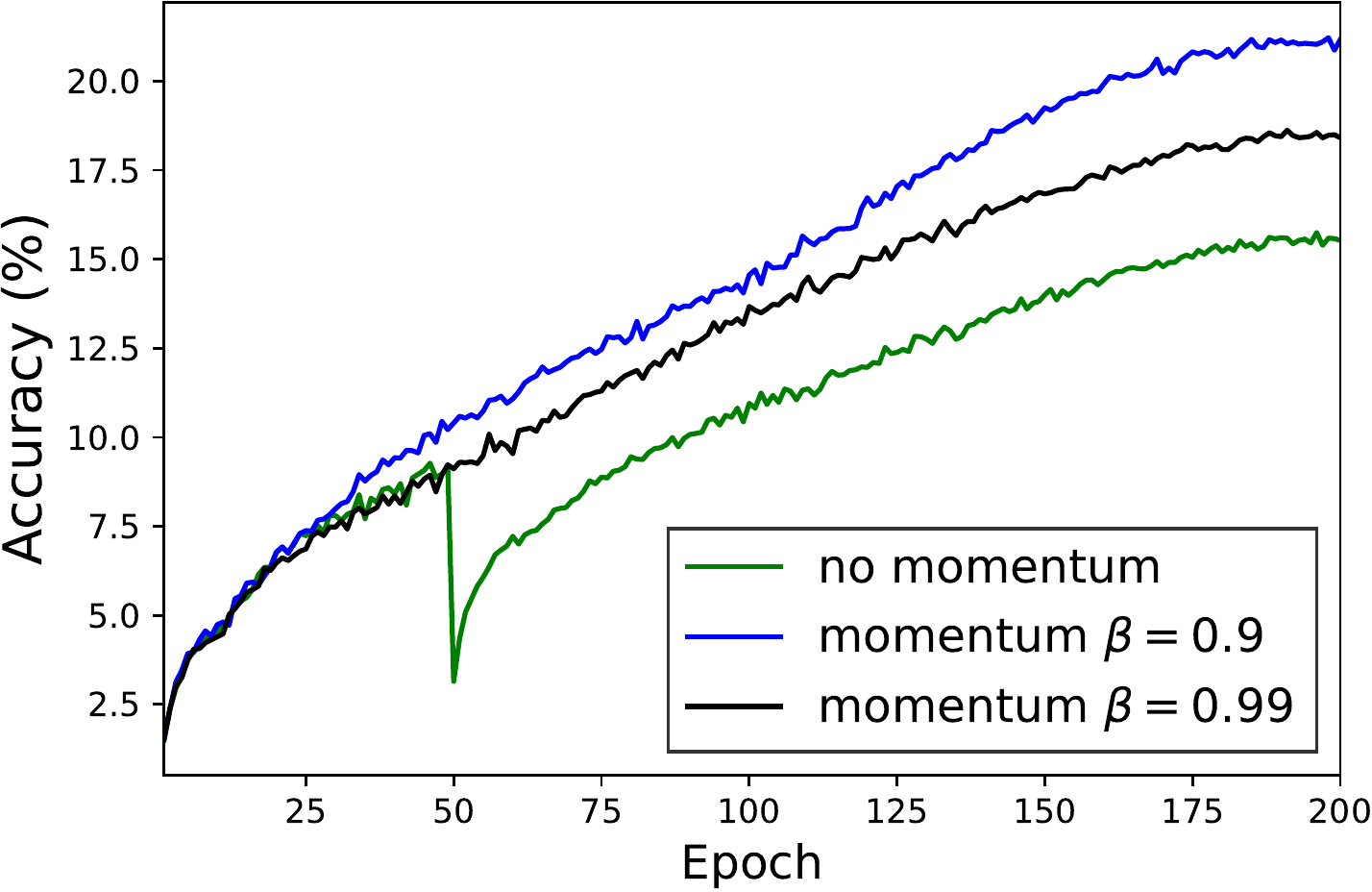}}
  \hfill
  \subfloat[Linear test accuracy] {\includegraphics[width=0.33\linewidth]{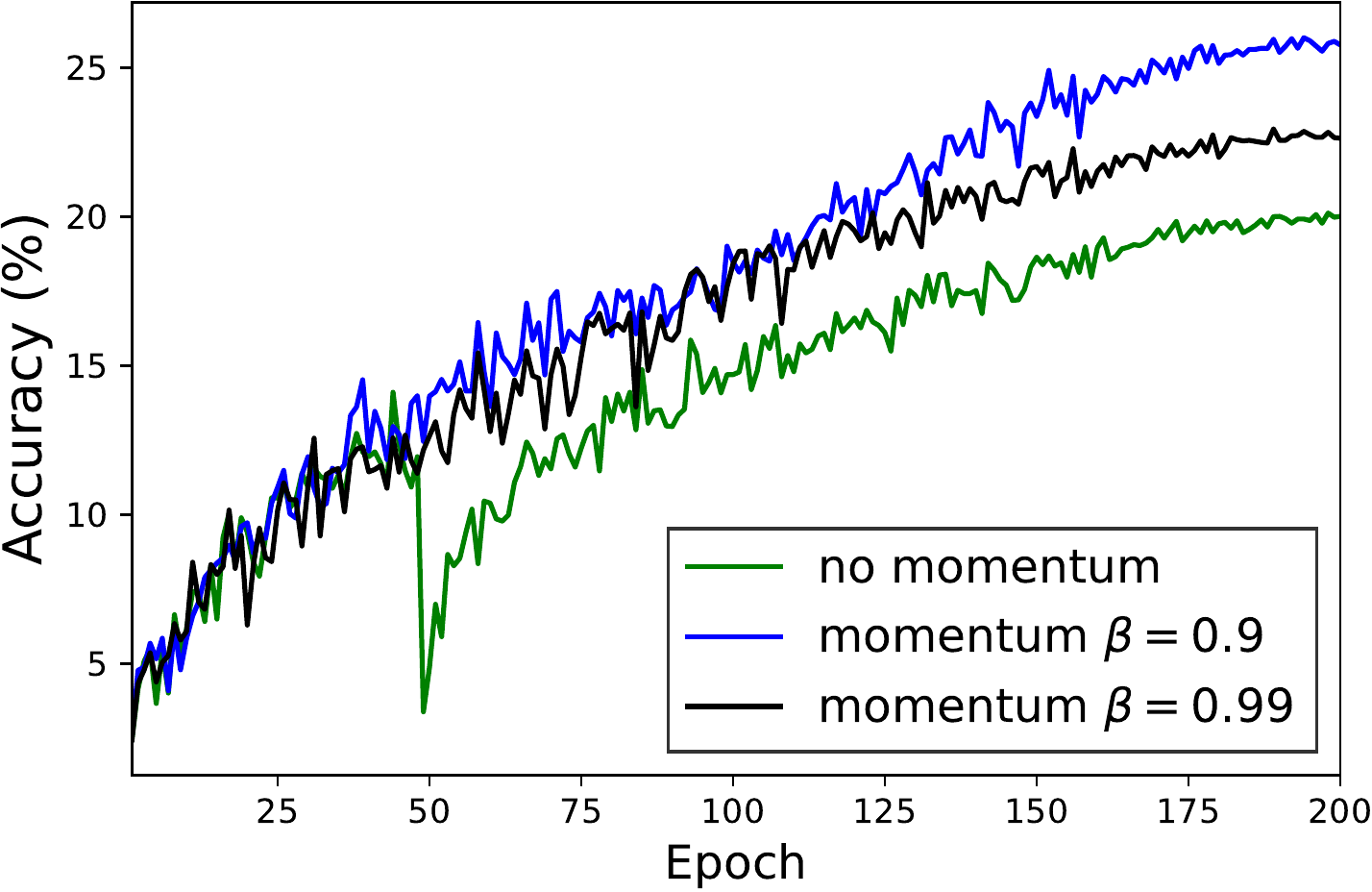}}
  \hfill
  \subfloat[KNN-1 test accuracy] {\includegraphics[width=0.33\linewidth]{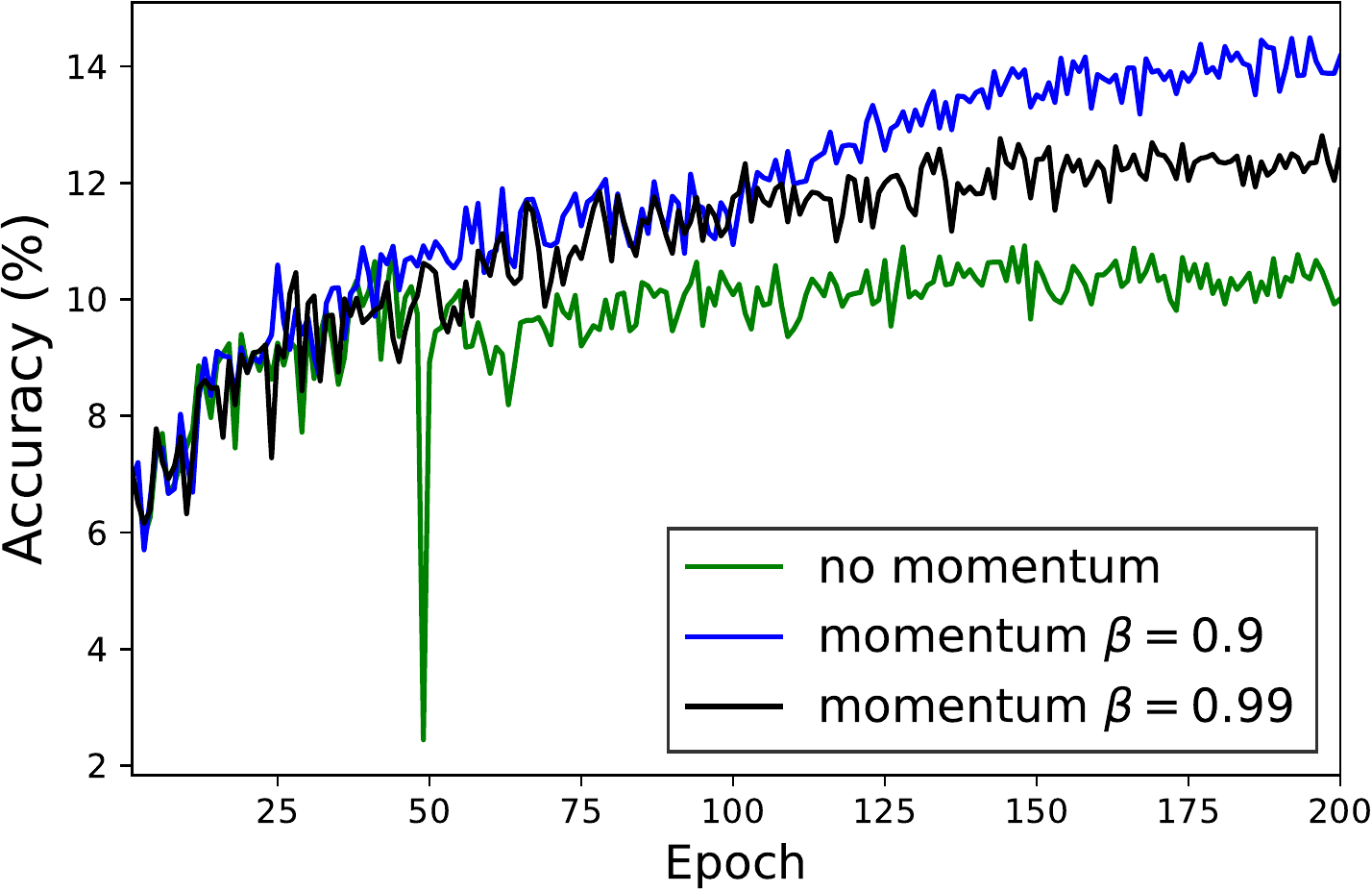}}
  \vspace{-5pt}
  \caption{Performance with EMA effects on CIFAR-100. The abnormal accuracy around epoch 50 out of 200 (w/o momentum) happens when weights (filters) have suddenly fluctuated at step 10k in Fig. Fig. \ref{fig:filter3d}. With momentum, this instability is successfully prevented (\textit{black} and \textit{\textcolor{blue}{blue lines}}). }
  \label{fig:accuracy}
  \vspace{-10pt}
\end{figure}

\textit{At first}, we conduct experiment without momentum ($\beta=0$). \textit{Second}, we perform EMA ($\beta=0.9$, and $\beta=0.99$) to only FC2 of the \textit{target} projector (Fig. \ref{fig:ema_whole_part}b).
Our goal is to observe what happens to the network weight w/ and w/o momentum in the part that can be visualized. Note that, FC2(2,256) in \textit{pytorch} \cite{NEURIPS2019_pytorch} means $input\_dim=2$, $output\_dim=256$, and weight $w$ has shape of $256\times 2$. In other words, FC2 has a total of 256 filters with $dim=2$ that can be visualized. 

Tab. \ref{tab:why_momentum} shows that without momentum, the top-1 linear and KNN-1 test accuracy are 20.01\% and 10.01\%, respectively. With momentum, it can significantly boost the top-1 linear accuracy to \textbf{25.76\%} and KNN-1 to \textbf{14.17\%}. 
Interestingly, it reveals that momentum actually has an effect even though EMA is applied to a tiny part (FC2) of the \textit{target} encoder. \textit{How can it help?}
\begin{wraptable}[8]{ht}{0.5\textwidth}
    \caption{The EMA is applied to the FC2.}
    \label{tab:why_momentum} 
    \centering
    \resizebox{1.0\hsize}{!}{
    \begin{tabular}{ccc}
    \toprule
    Setup & Top-1 (\%) & KNN-1 (\%) \\
    \midrule
    Fixed random encoder & \textcolor{red}{15.21} & \textcolor{red}{6.85} \\
    w/o momentum & 20.01 & 10.01 \\
    w/ momentum, $\beta=0.9$ & \textbf{25.76} & \textbf{14.17} \\
    w/ momentum, $\beta=0.99$ & \textbf{22.63} & \textbf{12.56} \\
    \bottomrule
    \end{tabular}}
\end{wraptable}
We find that with EMA ($\beta=0.9$), the weight of the \textit{online} network is much more stable than training w/o EMA as clearly shown in Fig. \ref{fig:filter3d}. Here we plot five lines, \textcolor{Green}{\textit{green}} line denotes \textit{no momentum $\beta=0$}; \textcolor{blue}{\textit{blue}} and \textcolor{red}{\textit{red}} lines show the \textit{student} and \textit{teacher} with momentum $\beta=0.9$; \textit{black} and \textcolor{orange}{\textit{orange}} lines illustrate the \textit{student} and \textit{teacher} with momentum $\beta=0.99$.

We visualize FC2 weights in 3D Fig. \ref{fig:filter3d} where these weights are recorded in 200 epochs with a total of 40k steps. Fig. \ref{fig:accuracy} shows the corresponding top-1 linear accuracy with \textit{no momentum, momentum $\beta=0.9$, and momentum $\beta=0.99$}. We observe a sudden change of the weight in Fig. \ref{fig:filter3d} which strongly corresponds to the accuracy drop in Fig. \ref{fig:accuracy}. 

More visualizations of filters (weights) are provided in \textbf{Appendix}.
These experiments show that using EMA into FC2 (shape $256\times 2$) whose weights have only two dimensions can have a great influence on the stability in training and thus increases performance by a large margin ($+5.75\%$). Note that in optimal setting \cite{turrisi2021sololearn}, FC2 has $4096$ dimensions ($256\times 4096$) which is much more powerful than that in our visualization experiments.

This phenomenon motivates us to investigate the EMA effect in the other parts of the encoder which have different capacities and parameters. To this end, we dig into the main concern of the paper with two questions: \textit{1) How does the EMA affect each part of the network encoder?} and \textit{2) Does the EMA update to the whole network or EMA update in some certain parts of the network perform the best?}


\section{Impact of Partial EMA Update on Different Encoder Stages}

\textbf{Partial EMA Update.} ResNet-18 \cite{he2016deep} is a convolutional neural network with 18 layers whose design has four blocks. Note that, in ResNet-18, there has a convolutional layer before the four residual blocks as illustrated in Fig. \ref{fig:ema_whole_part}a). We inject EMA to each separate block to see how it influences the final performance. BYOL \cite{grill2020bootstrap} is chosen to investigate. All these experiments are conducted in CIFAR-100 with 200 epochs and all evaluation protocols are followed as \cite{grill2020bootstrap}. 

The whole network of BYOL includes 1) \textit{backbone} (ResNet-18), 2) \textit{Projector}, and 3) \textit{Predictor}. By default, the \textit{target} encoder network of BYOL is updated EMA for entire encoder, \ie both \textit{backbone} and \textit{projector}. 

We split \textit{backbone} into different blocks as its design: \textit{Conv1}, \textit{Block1}, \textit{Block2}, \textit{Block3}, and \textit{Block4} as shown in Fig. \ref{fig:ema_whole_part}a). We decompose the parameters set $\theta$ of the whole \textit{online} network into a set of consequential parameters $\theta_1,\theta_2,\theta_3,\theta_4,\theta_5,\theta_6,\theta_7$ and $\xi$ in the \textit{target} network to separate parameters $\xi_1,\xi_2,\xi_3,\xi_4,\xi_5,\xi_6$ for different parts of the encoder in Fig. \ref{fig:ema_whole_part}a). 

Note that each block contains four convolutional layers followed by a batch normalization layer and ReLU \cite{he2016deep}. 
We use EMA with $\beta=0.99$ as default library \cite{turrisi2021sololearn} for each block of the \textit{target} network while keeping all other parts sharing weight with the corresponding parts of the \textit{online} network, \ie $\beta=0$.

We also consider a case of applying EMA to a very small part at the beginning of the whole network, here we choose the first convolutional layer of \textit{backbone} namely \textit{Conv1}. Interestingly, we find that \textit{the portion near the input of the encoder gives the least impact on performance, while the latter parts of the encoder have much more influence}. 

\begin{figure}[!htbp]
  \centering
  \subfloat[Momentum in separate part] {\includegraphics[width=0.49\linewidth]{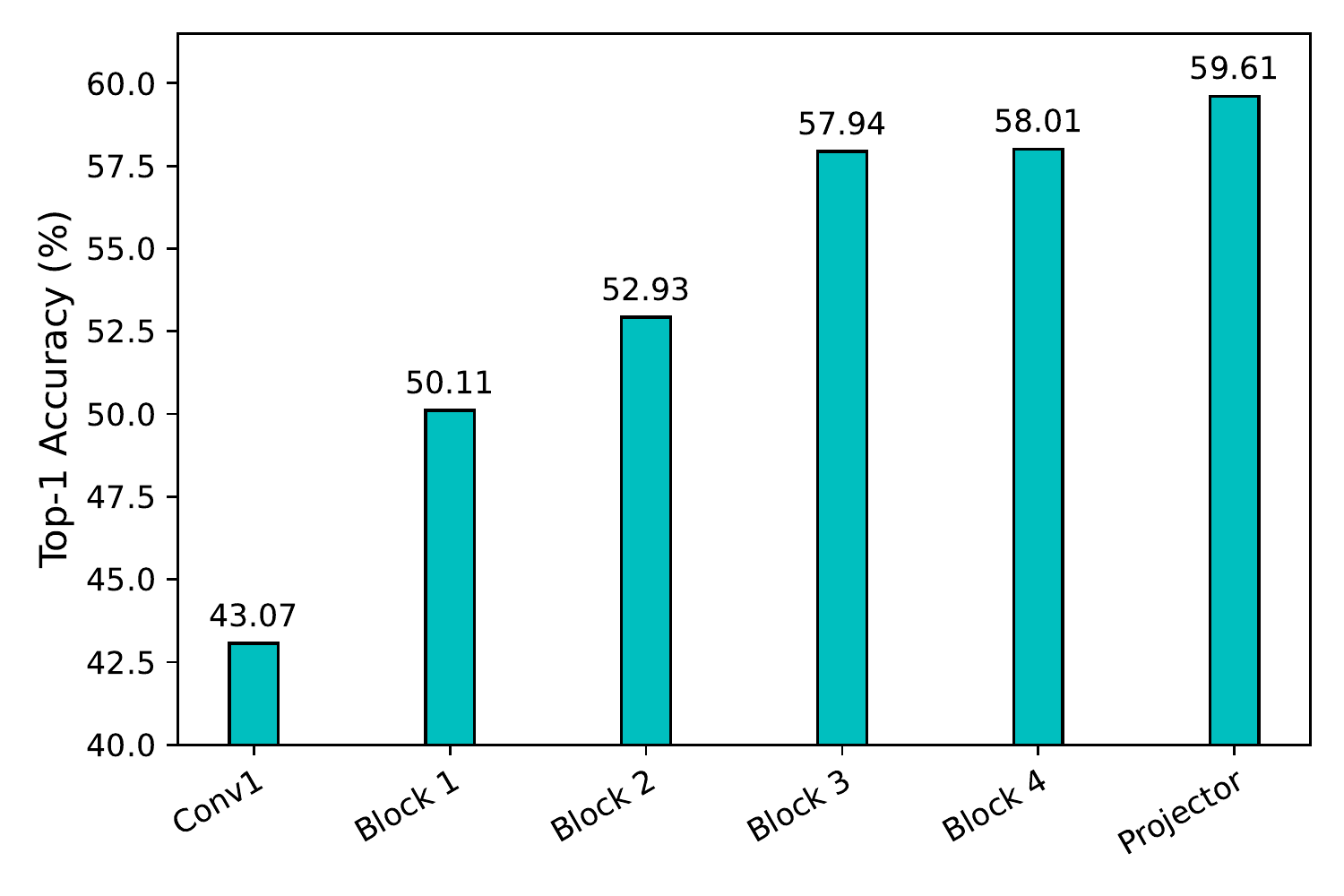}}
  \hfill
  \subfloat[Momentum in each combination] {\includegraphics[width=0.49\linewidth]{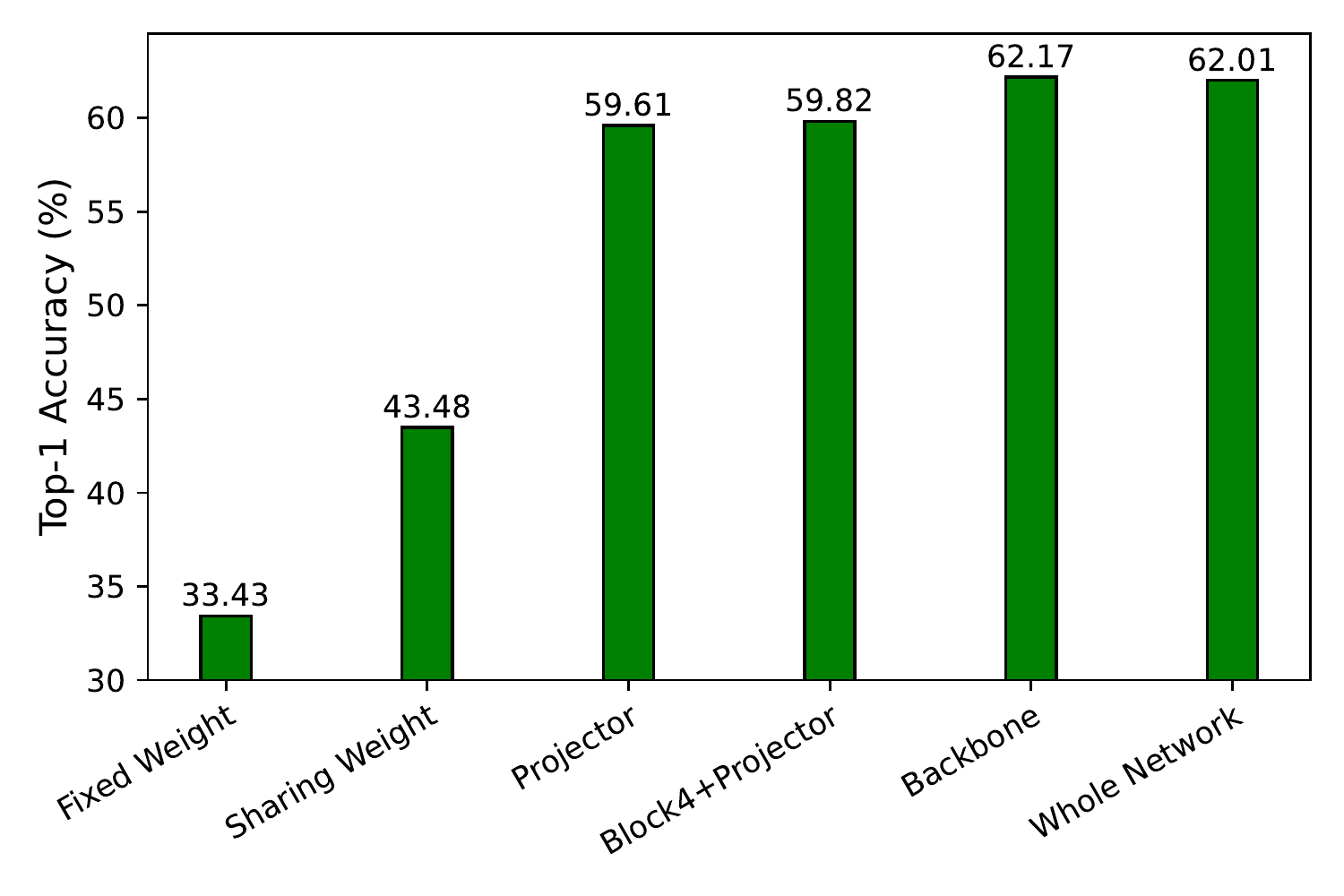}}
  \vspace{-5pt}
  \caption{ Effects of momentum in each individual part in the whole network encoder BYOL. We report top-1 accuracy and train 200 epochs on CIFAR-100. Here \textit{backbone} is the ResNet-18 which has 4 blocks in its architecture. \textit{Conv1} means that we take the first convolutional layer in the backbone. When applying momentum to each part, all other parts will be sharing weight (\ie $\beta = 0$). }
  \label{fig:ema_compare_parts}
\end{figure}

As shown in Fig. \ref{fig:ema_compare_parts} a), EMA in \textit{Conv1}, the first part of the network gives an accuracy of $43.07\%$ which is almost similar to the case not applying EMA with the accuracy of $43.48\%$. 
The impact of EMA keeps increasing to the final layers. Here, applying EMA to the final layer, \ie \textit{projector} which has only two FC layers can bring significant improvement from $43.48\%$ to $59.61\%$ which is nearly comparable to baseline which updates EMA in the whole deep encoder with $62.01\%$.

This phenomenon can be attributed to the fact that the earlier parts of the network share the weight with the online network which updates faster to help learn the latest information from the input and the latter final parts need to slow down the network update for training more stable. 
We elaborate on this phenomenon by analyzing the gradient for each block below.
We also investigate the EMA influence of each combination of some parts of the encoder. Applying EMA on \textit{Block4+projector} works comparable to EMA on \textit{projector}. Combining \textit{Conv1} and four blocks (equal to \textit{backbone}) can give the comparable accuracy to the baseline (\textit{whole Network}), \ie $62.17\%$ vs. $62.01\%$, respectively, Fig. \ref{fig:ema_compare_parts} b).

Overall, momentum/EMA on the final part of the SSLs encoder (\ie \textit{projector}) has the most important influence on the performance in the classification downstream task. For the SSL methods such as DINO, MoCo v2, and ReSSL, EMA in the final part of the encoder is even more critical compared to the BYOL as shown in Tab. \ref{tab:ame_partly_update} and Fig. \ref{fig:ema_compare_parts}.

A natural question is raised: \textit{why does EMA apply in the final parts has more influence compared to the earlier parts of the encoder?}

\textbf{Gradient Analysis.} Amongst SSL frameworks, we take a negative-free method, such as BYOL \cite{grill2020bootstrap} to perform gradient analysis for simplicity.
We monitor the gradient to each part of the network encoder to find some clues to elaborate on why EMA has no or minor effect on the earlier but latter parts. For visual observation, we record the $l_\infty-$norm gradient magnitude of the overall loss with respect to the output of each block for the case shown in Fig. \ref{fig:filter3d} with $\beta=0$ where the abnormal point shows the strong instability during the training.

\begin{wrapfigure}{r}{0.5\textwidth}
    \centering
    \includegraphics[width=0.48\textwidth]{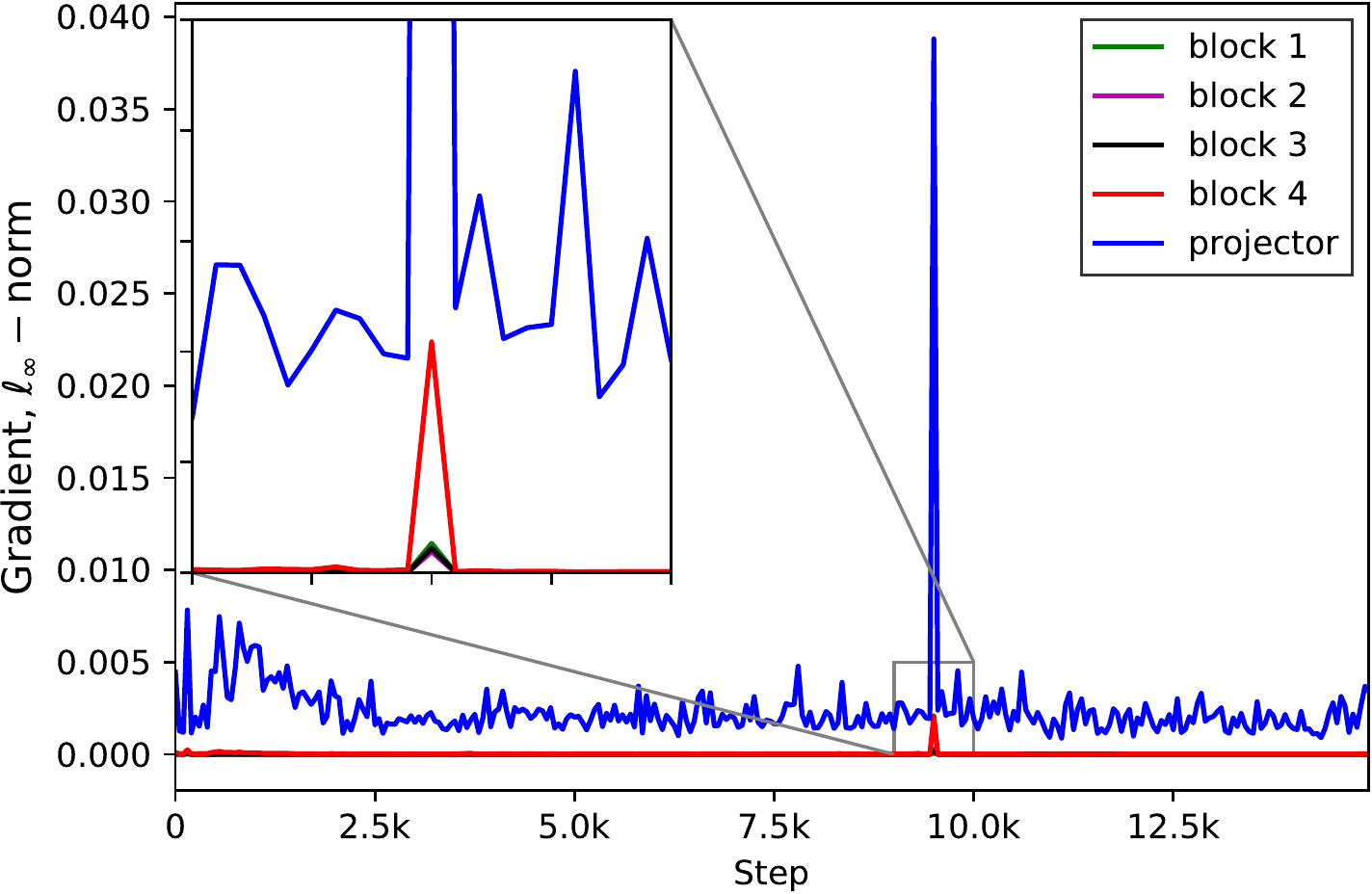}
    \caption{Gradient magnitude of each layer. It shows the gradients of latter parts such as \textit{block 4} and \textit{projector} of the encoder are more fluctuated.}
    \label{fig:gradient_inf_norm}
\end{wrapfigure}

Fig. \ref{fig:gradient_inf_norm} clearly shows that the magnitude gradient in the final part of the encoder \ie \textit{projector} has the most significance and is most fluctuated. The earlier part such as \textit{block 4} has a much smaller gradient magnitude and is more stable. 
And, the very beginning blocks such as \textit{block 1}, \textit{block 2}, and \textit{block 3} have the smallest magnitude gradients and are most stable.

This empirical observation explains why applying EMA (Eq. \ref{eq:ema_update}) to the final part of the network encoder has the most influence compared to the previous stages of the network encoder.
In other words, making the weight in the final stages learned from the past to stabilize the training can have the most effects on the entire deep network encoder.

Based on the above observation, we can also use additionally 1) a simple technique such as BN \cite{santurkar2018does} to the output of the \textit{projector} and \textit{block 4} or 2) freezing the target projector to stabilize the training to improve performance \cite{chen2021mocov3}. We show the results when using EMA in the final stage projector (without EMA to the deep backbone) with BN in the output of \textit{projector} and/or \textit{block 4} can match the baseline BYOL \cite{grill2020bootstrap} in the Appendix.

\section{Projector-only Momentum in SSL}
\textbf{Double Propagation of Momentum.} The benefit of momentum comes at the cost of causing double forward computation \cite{grill2020bootstrap,he2020momentum,caron2021emerging,zheng2021ressl} compared to non-momentum-based frameworks \cite{chen2020simple,chen2021exploring}. Motivated by our previous analysis, we propose the \textit{projector-only momentum} for SSL frameworks which applies EMA only to the projector, while sharing parameters for deep backbone as an example in Fig. \ref{fig:ema_whole_part} b) for BYOL. Our method causes only double propagation on the light projector whose computation is negligible compared to that of the backbone (\ie ResNet-18 and ResNet-50).

\textbf{Implementation Details.}
The implementation is based on the open library solo-learn for SSL \cite{turrisi2021sololearn}. To investigate the momentum/EMA, we consider different SSL methods, datasets, and ResNet backbone architectures. Three widely used benchmark datasets are considered including: CIFAR-100 \cite{krizhevsky2009learning}, ImagetNet-100 \cite{tian2019contrastive}, and ImageNet-1K \cite{krizhevsky2012imagenet}. Different deep backbones (ResNet-18, ResNet-50) are considered. We follow the common protocol for evaluation the SSL \cite{grill2020bootstrap}.

\textbf{Training and Evaluation.} The encoder is pre-trained in an unsupervised manner on the training set of each dataset without labels. The transfer capability is measured by the classification downstream task \cite{grill2020bootstrap,he2020momentum}. To be specific, we evaluate the pre-trained representations by training a linear classifier on frozen representations with the corresponding test set \cite{turrisi2021sololearn}. Classification accuracy is used as the performance metric. More details and setups for SSL are provided in the Appendix.

\begin{wraptable}[15]{ht}{0.6\textwidth}
    \caption{The EMA is applied to either \textit{backbone} or \textit{projector} for each SSL framework. We report top-1 accuracy with ResNet-18 backbone on CIFAR-100. ``-'' denotes sharing weight ($\beta=0$).} 
    \label{tab:ame_partly_update}
    \vspace{-5pt}
    \centering
    \resizebox{1.0\hsize}{!}{
    \begin{tabular}{cccc}
    \toprule
    Method & EMA Backbone & EMA Projector & Acc (\%) \\
    \midrule
    BYOL \cite{grill2020bootstrap} (baseline) & \checkmark & \checkmark & 62.01 \\
     & \checkmark & - & \textcolor{red}{62.17} \\
     & - & \checkmark & \textcolor{blue}{59.61} \\
    \midrule
    DINO \cite{caron2021emerging} (baseline) & \checkmark & \checkmark & 58.12 \\
     & \checkmark & - & \textcolor{red}{8.11} \\
     & - & \checkmark & \textcolor{blue}{58.14} \\
     \midrule
    ReSSL \cite{zheng2021ressl} (baseline) & \checkmark & \checkmark & 54.66 \\
     & \checkmark & - & \textcolor{red}{29.79} \\
     & - & \checkmark & \textcolor{blue}{53.51} \\
    \midrule
    MoCo v2 \cite{chen2020improved} (baseline) & \checkmark & \checkmark & 62.34 \\
     & \checkmark & - & \textcolor{red}{45.84} \\
     & - & \checkmark & \textcolor{blue}{60.39} \\
    \bottomrule
    \end{tabular} }
\end{wraptable}
\textbf{Projector-only Momentum SSLs.} We investigate the effect of EMA to each part of the EMA-based frameworks including DINO \cite{caron2021emerging}, MoCo v2 \cite{chen2020improved}, ReSSL \cite{zheng2021ressl}, and BYOL \cite{grill2020bootstrap}. 
Tab. \ref{tab:ame_partly_update} shows that, unlike BYOL, all other frameworks perform poorly if EMA updates on the backbone only. By contrast, Tab. \ref{tab:ame_partly_update} reveals that EMA applying on the final part of the target network, \ie \textit{projector}, all SSL frameworks perform competitively with baselines.

Interestingly, with DINO, performing EMA only to the projector (3 layers) can give the same performance as the baseline which needs to \textit{update EMA for every layer} in the whole deep encoder, 21 and 53 layers for ResNet-18, ResNet-50, respectively. It suggests that EMA to the final part of the DINO encoder is sufficient and it helps to reduce half of the parameter needed to store in the memory for the deep momentum backbone and save half forwards while still exploiting the benefit of the momentum via a shallow projector (3 layers). Different from BYOL \cite{grill2020bootstrap}, EMA updates only to \textit{backbone} of the other SSLs witnesses the significant drop on performance as shown in Tab. \ref{tab:ame_partly_update}.

\label{sec:experiment_detail}


\begin{table}[!htbp]
    \caption{DINO with ImageNet-1K, 200 epochs using ResNet-50 backbone. Here ``ep'' stands for the epoch, and ``Fwd'' denotes the number of forward to ResNet-50 of each image. }
    \label{tab:imagenet1k}
    \centering
    \resizebox{1.0\hsize}{!}{
    \begin{tabular}{ccccccc}
    \toprule
    Method & \# EMA layers & Params (MB) & \#Fwd & Time (min/ep) & Top-1 (\%) & Top-5 (\%) \\
    \midrule
    DINO \cite{caron2021emerging} & \textcolor{red}{53} & 47.0 & 2 & 56.5 & 67.9 & 88.3 \\
    EMA \textit{projector} & \textcolor{blue}{\textbf{3}} & \textcolor{blue}{\textbf{23.5}} & \textcolor{blue}{\textbf{1}} & \textcolor{blue}{\textbf{38.5}} & 67.8 & 88.2 \\
    \bottomrule
    \end{tabular} }
\end{table}

\textbf{Large-Scale Dataset.} We conduct the experiments of partial update EMA in the larger scale dataset, \ie ImageNet-100 \cite{tian2020contrastive} (Tab. \ref{tab:ame_partly_update_IN100}) and ImageNet-1K \cite{krizhevsky2012imagenet} (Tab. \ref{tab:imagenet1k}) whose image size is $224 \times 224$. All methods are trained in 200 epochs. ResNet-18 and ResNet-50 are used for ImageNet-100 and ImageNet-1K, respectively. Tab. \ref{tab:ame_partly_update_IN100} shows that results are consistent as the ones in CIFAR-100 dataset. Specifically, we observe that for the large-scale dataset of ImageNet-100, with $\beta=0.99$ applying EMA to only \textit{projector} gives slightly a decrease in accuracy while keeping competitive performance with $\beta=0.9995$ for all SSL frameworks.

\begin{wraptable}[14]{ht}{0.5\textwidth}
    \caption{Partial update EMA for SSLs frameworks with ResNet-50. We train 200 epochs on CIFAR-100. All methods is set with $\beta=0.99$. We report top-1 accuracy, ``-'' denotes sharing weight.} 
    \label{tab:ame_cifar100_res50}
    \centering
    \resizebox{1.0\hsize}{!}{
    \begin{tabular}{cccc}
    \toprule
    Method & EMA Backbone & EMA Projector & Acc (\%) \\
    \noalign{\smallskip}
    \midrule
    BYOL \cite{grill2020bootstrap} & \checkmark & \checkmark & 63.71 \\
     & - & \checkmark & 59.88 \\
    \midrule
    MoCo v2 \cite{chen2020improved} & \checkmark & \checkmark & 65.12 \\
     & - & \checkmark & 61.76 \\
    \midrule
    ReSSL \cite{zheng2021ressl} & \checkmark & \checkmark & 57.07 \\
     & - & \checkmark & 56.10 \\
    \midrule
    DINO \cite{caron2021emerging} & \checkmark & \checkmark & 53.04 \\
     & - & \checkmark & \textbf{56.54} \\
    \bottomrule
    \end{tabular} }
\end{wraptable}
Applying EMA to the projector in DINO keeps the competitive performance of $74.22\%$ compared to the baseline of $74.16\%$. Consistently for the larger scale dataset, with highly aggressive EMA, \ie $\beta=0.9995$, all methods witness a dramatic drop whereas EMA projector-only has a slight drop and still keeps the competitive performance. Result on ImageNet-1K is consistent for DINO whose accuracy is maintained without EMA on deep backbone (Tab. \ref{tab:imagenet1k}). This suggests that for DINO with ResNet, EMA on the \textit{projector} is sufficient. 
\begin{table}[!htbp]
    \caption{The EMA is applied to whole \textit{encoder} or \textit{projector} for each SSL framework with baseline $\beta=0.99$ and more aggressive update with $\beta=0.9995$. We report top-1 accuracy (\%). All SSL frameworks are run for 200 epochs with ResNet-18 on ImageNet-100, ``-'' denotes sharing weight. Our method is less sensitive to the change of $\beta$ compared to baselines.}
    \label{tab:ame_partly_update_IN100}
    \smallskip
    \centering
    \resizebox{1.0\hsize}{!}{
    \begin{tabular}{cccccc}
    \toprule
    Method & EMA Backbone & EMA Projector & $\beta=0.99$ (\%) & $\beta=0.9995$ (\%) & $\Delta_{\text{acc}}$ (\%) \\
    \noalign{\smallskip}
    \midrule
    BYOL \cite{grill2020bootstrap} & \checkmark & \checkmark & 76.60 & 71.18 & $\color{red}-5.42$ \\
     & - & \checkmark & 75.02 & \textbf{75.52} & $\color{blue}+0.50$ \\
    \midrule
    DINO \cite{caron2021emerging} & \checkmark & \checkmark & 74.16 & 63.78 & $\color{red}-10.38$ \\
     & - & \checkmark & \textbf{74.22} & \textbf{72.02} & $\color{blue}-2.20$ \\
    \midrule
    ReSSL \cite{zheng2021ressl} & \checkmark & \checkmark & 74.02 & 61.48 & $\color{red}-12.54$ \\
     & - & \checkmark & 69.12 & \textbf{65.81} & $\color{blue}-3.31$ \\
    \midrule
    MoCo v2 \cite{chen2020improved} & \checkmark & \checkmark & 76.48 & 69.86 & $\color{red}-6.62$ \\
     & - & \checkmark & 74.31 & \textbf{72.06} & $\color{blue}-2.25$ \\
    \bottomrule
    \end{tabular} }
\end{table}
Based on the learning curve, if training longer, \ie 1000 epochs for ImageNet-1K, the gap between DINO baseline and DINO with EMA on projector reaches zero. We provide the training trend SSL frameworks in Appendix.

\textbf{Deeper Backbone with ResNet-50.} We keep the same \textit{projector} and consider a deeper backbone, \ie ResNet-50 (3 times deeper than ResNet-18). Two datasets are used: CIFAR-100, and ImageNet-1K. All methods are trained with 200 epochs, $\beta=0.99$. With ImageNet-1K, we conduct experiments for DINO \cite{caron2021emerging}. As shown in Tab. \ref{tab:ame_cifar100_res50} for a deeper backbone (ResNet-50), SSL frameworks such as BYOL \cite{grill2020bootstrap}, MoCo v2 \cite{chen2020improved}, and ReSSL \cite{zheng2021ressl} show the consistent results as the backbone ResNet-18 which are investigated above where EMA to the \textit{projector} can give the competitive performance no matter how deep backbone is (18 and 50 layers).

Interestingly, for DINO \cite{caron2021emerging} in CIFAR-100, baseline updates EMA to whole deep encoder (backbone + projector) performs inferior than the case updates EMA just to the projector, \ie $53.04\%$ vs. \textbf{$56.54\%$}. In ImageNet-1K, Tab. \ref{tab:imagenet1k} shows their performance are almost comparable ($67.9\%$ vs. $67.8\%$ in top-1 accuracy) where DINO needs to update EMA for total of \textit{53} layers (backbone + projector) compared to EMA on \textit{projector} with only \textit{three} layers in depth.


\section{Computation and Processing Time}
Amongst SSL frameworks, we find that DINO is the most stable method when training with the EMA projector. Therefore, we focus on DINO in the setting with ImageNet-1K, and the common use backbone is ResNet-50. We measure the processing time per epoch based on two GPUs NVIDIA RTX A6000.

As shown in Tab. \ref{tab:imagenet1k}, our method \textit{EMA projector} just needs half of the parameters for the backbone (just reduce half forward computational overhead) and shows significant efficiency in terms of process time per epoch while achieving the same performance compared to DINO \cite{caron2021emerging}. 

Concretely, the proposed \textit{EMA projector} updates EMA only for 3 layers of the projector and gets rid of the need for the forward to the backbone \ie ResNet-50 (with 50 layers in depth) which is an advantage compared to the traditional DINO. Note that the core of SSL lies in learning representation with multiple augmentations, and the total processing time may quickly become a burden. Our analysis suggests a potential paradigm for future research in self-supervised learning to solve the computational bottleneck, especially for the labs with low hardware resources.

\section{Discussion}
\textbf{Momentum Coefficient.}
It is expected that SSL frameworks give undesired performance if the EMA update is very slow. However, we show that even with the extremely slow update, EMA still can be maximally exploited to keep competitive performance. 
\begin{figure}[!htbp]
  \centering
  \subfloat[$\beta=0.99$]{\includegraphics[width=0.46\linewidth]{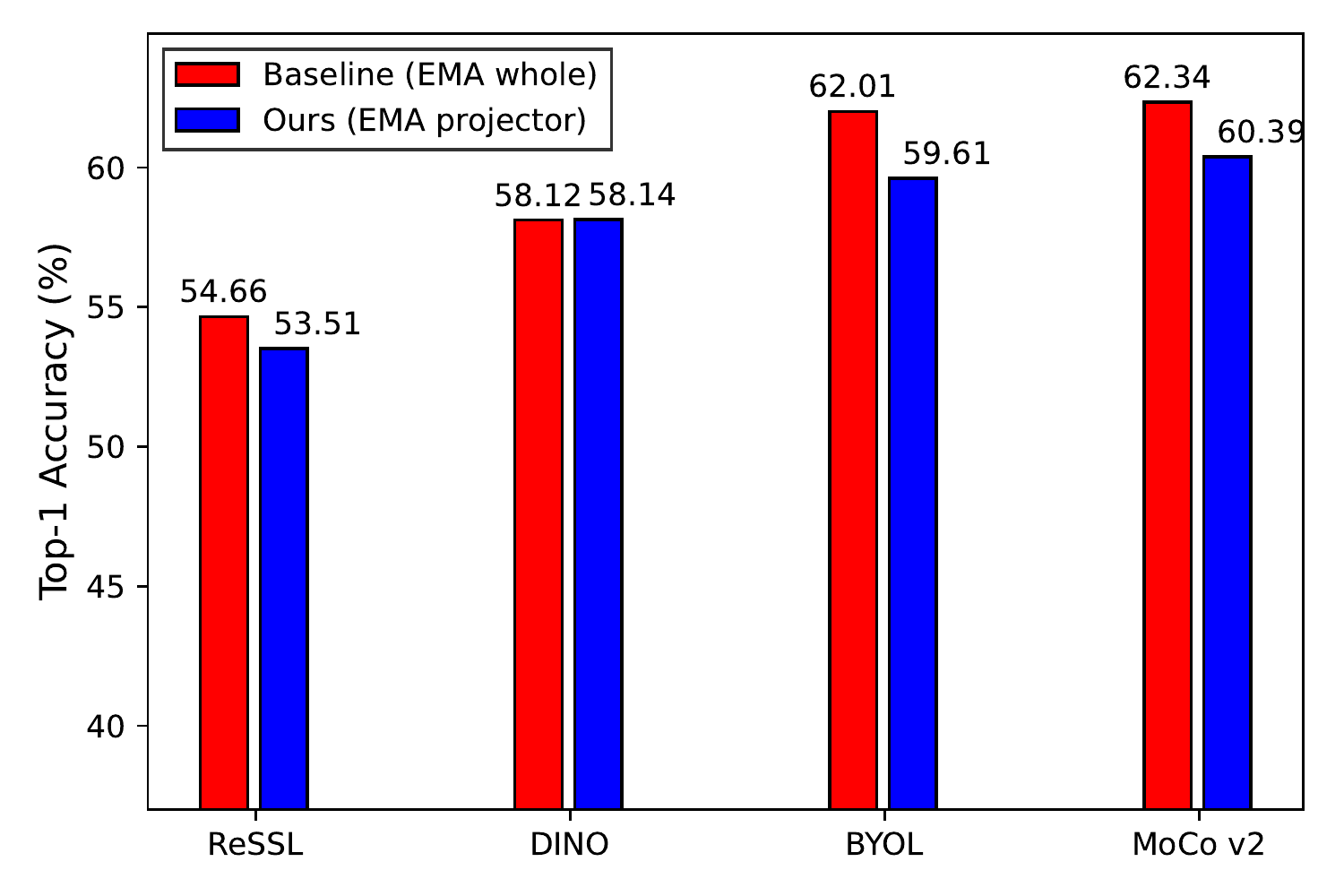}}
  \subfloat[$\beta=0.9995$]{\includegraphics[width=0.46\linewidth]{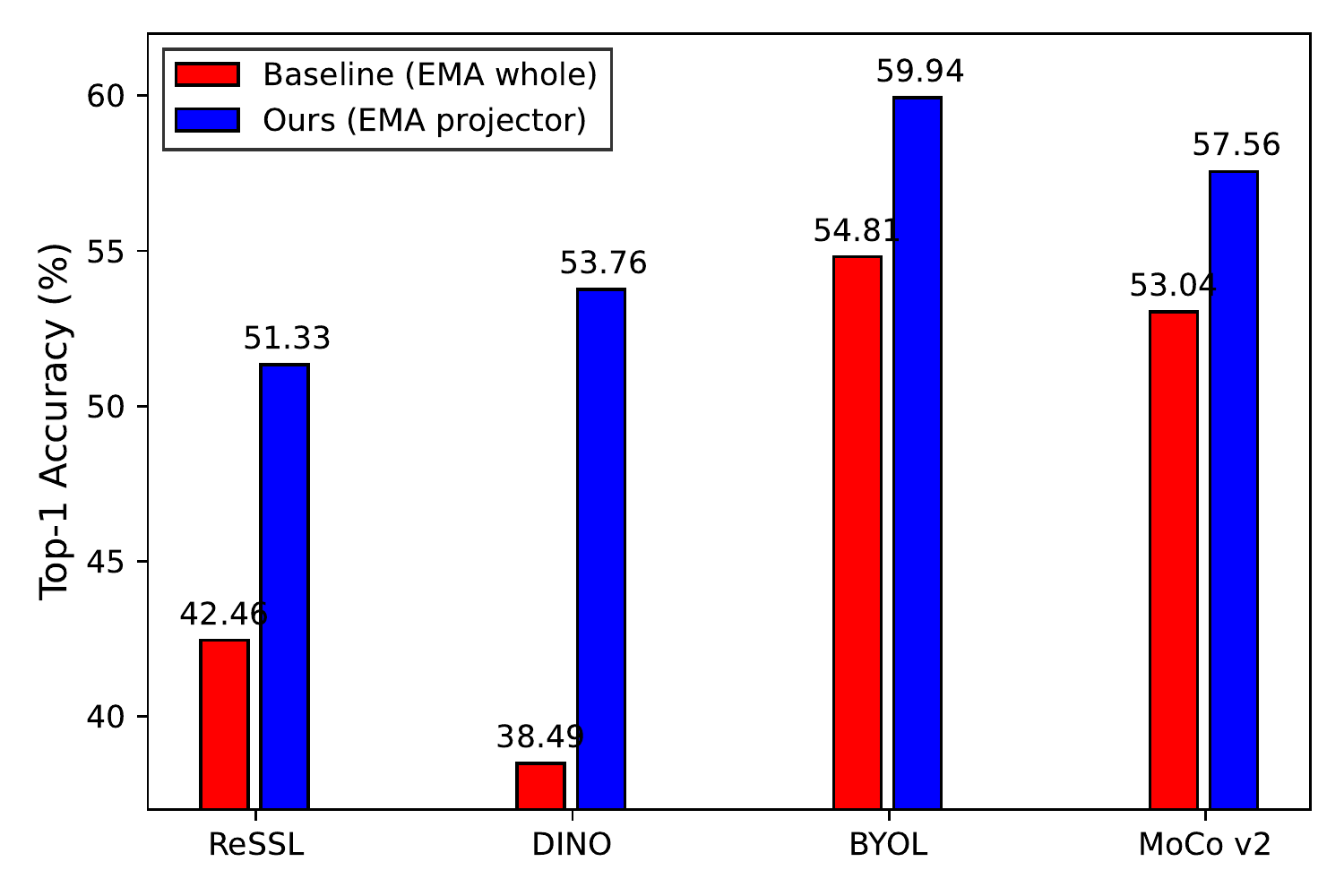}}
  \vspace{-5pt}
  \caption{Compare EMA updating to the whole encoder (baseline) and partial update EMA in two cases: (a) with the optimal parameter of momentum coefficient $\beta$ it shows that EMA in the \textit{projector} can give the competitive accuracy compared to baseline, and (b) with a more aggressive update, only updating EMA to the projector (ours) performs much better than baselines of all the considered frameworks, especially ReSSL and DINO give a big gap lower than our scheme.}
  \label{fig:compare_tau}
\end{figure}
We consider two scenarios: 1) EMA updates for the whole network in a conventional way, and 2) EMA partially updates to the only final part of the network encoder, \ie \textit{projector}. And we set $\beta=0.9995$ for all SSL baselines, which is slow enough to observe the differences. We find that with the highly aggressive update of EMA, all SSL frameworks drop dramatically in performance as shown in Fig. \ref{fig:compare_tau}a), and Fig. \ref{fig:compare_tau}b).
Specifically, ReSSL \cite{zheng2021ressl} degrades \textcolor{red}{$12.20$}\% (from $54.66\%$ to $42.46\%$), DINO \cite{caron2021emerging} impairs \textcolor{red}{$19.63$}\% compares to baseline (from $58.12\%$ to $38.49\%$), BYOL \cite{grill2020bootstrap} decreases \textcolor{red}{$7.20$}\% (from $62.01\%$ to $54.81\%$), and MoCo v2 \cite{chen2020improved} drops \textcolor{red}{$9.30$}\% (from $62.34\%$ to $53.04\%$). 

By contrast, the proposed partial EMA update witnesses the slightly drop on the final accuracy such as ReSSL, DINO, BYOL, and MoCo v2 only decrease \textcolor{blue}{$3.33$}\%, \textcolor{blue}{$4.36$}\%, \textcolor{blue}{$2.07$}\%, and \textcolor{blue}{$4.78$}\%, respectively.
The dramatically dropped performance of the conventional EMA because of the fact that a very high $\beta$ causes the network to learn very slowly so that the encoder cannot retrieve the new information from the new input augmentations promptly.

By contrast, with the new scheme, EMA updates only to the \textit{projector}, while keeping the backbone of the \textit{teacher}'s weight the same as \textit{student} which can leverage the encoder to learn quickly in the first stages while making sure the entire \textit{target} encoder not updated too fast which helps to stabilize the final part to improve performance.

Note that, given the highly aggressive EMA only to \textit{projector}, the accuracy slightly decreases from baseline but still is much higher than the cases using EMA to the whole network.
This shows that with the new scheme, EMA is still exploited to get its benefit in extreme cases which the traditional EMA approach cannot.

These experiments also show that the \textit{EMA projector} is less sensitive to the hyper-parameters of momentum coefficient compared to the traditional SSL approaches.

\begin{wraptable}[11]{ht}{0.3\textwidth}
    \caption{Freezing the target projector $g(.)$, we report top-1 accuracy with ResNet-18 backbone on CIFAR-100 trained with 200 epochs.} 
    \label{tab:large_learning_rate} 
    \centering
    \resizebox{1.0\hsize}{!}{
    \begin{tabular}{ccc}
    \toprule
    \multirow{2}{*}{Method} & \multicolumn{2}{c}{W/ predictor} \\ 
     & $\eta=1.0$ & $\eta=4.0$ \\ 
    \midrule
    BYOL \cite{grill2020bootstrap} & 62.01 & 58.01 \\ 
    Fixed $g(.)$ & 60.27 & \textbf{62.32} \\ 
     \bottomrule
    \end{tabular} }
\end{wraptable}
\textbf{Learning Rate.} The SSL encoders work with the appropriate small learning rate, \ie BYOL \cite{grill2020bootstrap} uses $\eta=1.0$ as the optimal parameter in setting of \cite{turrisi2021sololearn}. With larger learning rate, \ie $\eta=4.0$, BYOL's performance drops from baseline 62.01\% to 58.01\% ($\color{red}-4\%$) as shown in Tab. \ref{tab:large_learning_rate}. With our analysis above, the most instability in the encoder comes from the final part, \ie the projector. We keep the \textit{projector} of the \textit{target} network freeze during training, \ie $\beta=1.0$, while sharing parameters in backbone. We find that the encoder can learn well with a much larger learning rate. 
As shown in Tab. \ref{tab:large_learning_rate}, with $\eta=4.0$ keeping projector fixed can boost accuracy from 60.27\% to 62.32\% ($\color{blue}+2.05\%$) which is slightly better than BYOL. Note that BYOL requires a cost of double forward propagation to the backbone momentum compared to ours that shares backbone but fixed \textit{projector}.

\section{Related Works}
\textbf{Exponential Moving Average.} EMA/Momentum has been studied deeply for smoothing the original sequence signal \cite{klinker2011exponential,appel2005technical,pring2002technical}. It becomes the widely used technique in practices for most of fields ranging from optimization \cite{kingma2014adam,sutskever2013importance,ma2018quasi}, reinforcement learning \cite{vieillard2020momentum,korkmaz2020nesterov,haarnoja2018soft}, knowledge distillation \cite{tian2019contrastive,lee2022prototypical}, recent semi-supervised learning frameworks \cite{tarvainen2017mean,chen2020bigSimclrv2,cai2021exponential,li2021momentum}, and self-supervised learning methods \cite{grill2020bootstrap,he2020momentum,caron2021emerging,wang2022importance}. Momentum is interpreted as an average of consecutive $q$-functions in reinforcement learning \cite{vieillard2020momentum}, or is used in SSL frameworks preventing model collapse \cite{caron2021emerging,grill2020bootstrap,von2021self}. 

\textbf{Self-Supervised Learning.} Recently, SSL has been extensively studied in many filed applications \cite{PAIXAO2020107535, Eun_2020_CVPR, Zhuang_2020_CVPR, Wu_2021_CVPR, Pan_2021_CVPR, Wang_2021_CVPR, Hu_2021_ICCV, Diba_2021_ICCV,nozawa2021understanding, Zeng_2020_CVPR, Vasudeva_2021_ICCV, ZHANG2022108784, iscen2018mining, chuang2020debiased, wu2020conditional, NEURIPS2020_f7cade80,robinson2021contrastive,kaku2021intermediate}. SSL recently becomes an emerging paradigm and key research field thanks to its unique advantages over the traditional supervised learning approaches that does not require any human manual supervision and costly labelling process while showing the superior capability in learning representations for downstream tasks such as classification, segmentation, and object detection \cite{chen2020simple,wang2020DenseCL,he2020momentum,chen2020improved,chen2021mocov3,caron2021emerging,grill2020bootstrap,tian2021divide,caron2020unsupervised,ermolov2021whitening}. 

Amongst those seminal works, MoCo \cite{chen2020improved,chen2021mocov3}, DINO \cite{caron2021emerging}, ReSSL \cite{zheng2021ressl}, and BYOL \cite{grill2020bootstrap} are the state-of-the-art SSL frameworks that use the EMA technique in the target encoder. 
There are several SSL frameworks such as SimCLR \cite{chen2020simple}, SimSiam \cite{chen2021exploring}, and Barlow Twin \cite{zbontar2021barlow} do not use EMA in its architecture. 

Although those non-momentum-based methods exhibit the inferior performance in classification \ie in ImageNet-1K \cite{krizhevsky2012imagenet} or other downstream tasks compared to the momentum-based approaches \cite{chen2020improved,wang2020DenseCL,caron2021emerging}, they are still considered in practice for certain cases due to their simpler design, \ie without the need for storing parameters and forward of the momentum encoder \cite{madaan2022representational}.

\vspace{-5pt}
\section{Conclusion}
\vspace{-5pt}
In this paper, we investigate a new perspective on the benefit of momentum and show that its empirical benefit for boosting performance can be at least partly attributed to the stability effect. We reveal an intriguing observation that the benefit of applying momentum in the later stages of the SSL encoder is significantly higher than that in the early stage and provide a gradient analysis to justify the phenomenon. We note that the benefit of momentum comes at the cost of causing double forward computation. Motivated by the above observation, we propose a projector-only momentum method for maintaining the pros of momentum while mitigating its cons in the SSL frameworks, which is a potential direction for future research in solving the bottleneck of SSL in terms of computational cost.

Nevertheless, in this work, we focus on analyzing the behaviour of momentum with the ResNet backbones for image classification. The other complicated backbones such as ViT as well as other modalities such as text, audio, and video are also interesting to investigate. We let it for future work.





\appendix

\section{Experimental Details}
We directly clone the official code from the open library for SSL \cite{turrisi2021sololearn} and keep everything as default settings such as augmentation settings, projectors, predictors, and hyper-parameters in the bash files.

For datasets CIFAR-10, and CIFAR-100, we use a single GPU of NVIDIA TITAN Xp with batch size 256. For ImageNet-100, we use two GPUs of NVIDIA TITAN Xp with batch size 128 for each GPU. For ImageNet-1K, we use two GPUs of NVIDIA RTX A6000, batch size 128 for each GPU. Except for other notices, all experiments are conducted for 200 epochs.

We use a cosine learning rate schedule with the warm-up in 10 first epochs \cite{grill2020bootstrap}. All methods are trained without using labels. The representation quality is evaluated by training an online linear classifier with the label on the train set (frozen on backbone features) and evaluated with the label on the test set. As shown in the solo-learn \cite{turrisi2021sololearn} frameworks, the performance gap between online and offline linear evaluation is not significant. For the convenience to avoid the need of retraining a linear classifier after the encoder pre-training, we directly report top-1 accuracy (\%) on the test dataset with the online linear evaluation. This also facilitates observing the performance during the training.

For one experiment with ImageNet-1K \cite{krizhevsky2012imagenet}, it took 7 days 16h 56m 36s to train DINO \cite{caron2021emerging} with ResNet-50 backbone for 200 epochs (2 NVIDIA RTX A6000 GPUs). For ImageNet-100, it took 19h 33m 58s to train DINO with ResNet-18 backbone for 200 epochs (2 NVIDIA TITAN Xp GPUs).

\section{Code for Momentum-only Projector}
For all SSL methods, instead of using the momentum output of the momentum backbone as baselines, we use the output from the online backbone (with stop gradient). It helps to reduce half of the forward propagation to the backbone which has more computation overhead compared to a light projector. Therefore, our proposed projector-only momentum help maintain the pros of momentum while mitigating its cons. Below we attach a PyTorch \cite{NEURIPS2019_pytorch} pseudo-code for BYOL \cite{grill2020bootstrap} with our projector-only momentum, in Alg. \ref{alg:projector-only-byol}. Other SSL methods can be done in the same way.

\begin{algorithm}[h]
\caption{Pytorch-like Pseudo-code: Projector-only Momentum}
\label{alg:projector-only-byol}
\definecolor{codeblue}{rgb}{0.25,0.5,0.5}
\definecolor{codekw}{rgb}{0.85, 0.18, 0.50}
\lstset{
  backgroundcolor=\color{white},
  basicstyle=\fontsize{7.5pt}{7.5pt}\ttfamily\selectfont,
  columns=fullflexible,
  breaklines=true,
  captionpos=b,
  commentstyle=\fontsize{7.5pt}{7.5pt}\color{codeblue},
  keywordstyle=\fontsize{7.5pt}{7.5pt}\color{codekw},
}
\centering
\begin{lstlisting}[language=python]
# f: backbone (ResNet)
# g, g_m: online and target projector
# h: predictor

for x in loader:  # load a minibatch x with n samples
    x_1, x_2 = aug(x), aug(x)  # augmentations
    f_1, f_2 = f(x_1), f(x_2)  # backbone features
    z_1, z_2 = g(f_1), g(f_2)  # online projections
    z_1m, z_2m = g_m(f_1), g_m(f_2)  # momentum projections
    p_1, p_2 = h(z_1), h(z_2)  # predictions
    
    L = D(p_1, z_2m)/2 + D(p_2, z_1m)/2 # loss
 
    L.backward() # back-propagate
    update(f, g, h) # SGD update
    ema_update(g_m) # update parameters of the momentum projector, g_m := beta*g_m + (1-beta)*g

def D(p, z):
    z = z.detach() # stop gradient
    p = normalize(p, dim=1) # l2-normalize
    z = normalize(z, dim=1) # l2-normalize
    return -(p*z).sum(dim=1).mean()  # negative cosine similarity

\end{lstlisting}
\end{algorithm}

\section{Learning Curves}

\begin{wraptable}[12]{ht}{0.5\textwidth}
    \caption{Linear classification accuracy in CIFAR-100. All methods are trained in 200 epochs using ResNet-18 as the backbone. BN1, BN2 denote the BN applying to the output of \textit{block 4} and \textit{projector}, respectively.}
    \label{tab:byol_ema_bn}
    \centering
    \resizebox{1.0\hsize}{!}{
    \begin{tabular}{ccc}
    \toprule
    Method & Top-1 (\%) & KNN-1 (\%) \\
    \midrule
    BYOL \cite{grill2020bootstrap} & 62.01 & 54.84 \\
    w/o momentum & \textcolor{red}{43.48}  & \textcolor{red}{31.58} \\
    w/o momentum + BN1 + BN2 & \textcolor{red}{49.59}  & \textcolor{red}{42.92} \\
    EMA projector & \textcolor{blue}{59.61} & \textcolor{blue}{54.02} \\
    EMA projector + BN1 & \textcolor{blue}{60.75} & \textcolor{blue}{53.32} \\
    EMA projector + BN2 & \textcolor{blue}{61.97} & \textcolor{blue}{55.42} \\
    EMA projector + BN1 + BN2 & \textcolor{blue}{62.48} & \textcolor{blue}{55.66} \\
    \bottomrule
    \end{tabular}}
\end{wraptable}
\textbf{Learning Curves of DINO on ImageNet-1K.}
For ImageNet-1K, we train DINO With 200 epochs, using ResNet-50 as the backbone. We ran the DINO baseline that uses EMA for the entire target encoder and ran DINO with EMA only to the \textit{projector} similar to Alg. \ref{alg:projector-only-byol}. We report the top-1 and top-5 test accuracy for the \textit{val} set. For both cases, EMA applied on the projector (3 layers) is almost comparable.

\begin{figure}[!htbp]
  \vspace{-12pt}
  \centering
  \subfloat[Linear train accuracy] {\includegraphics[width=0.5\linewidth]{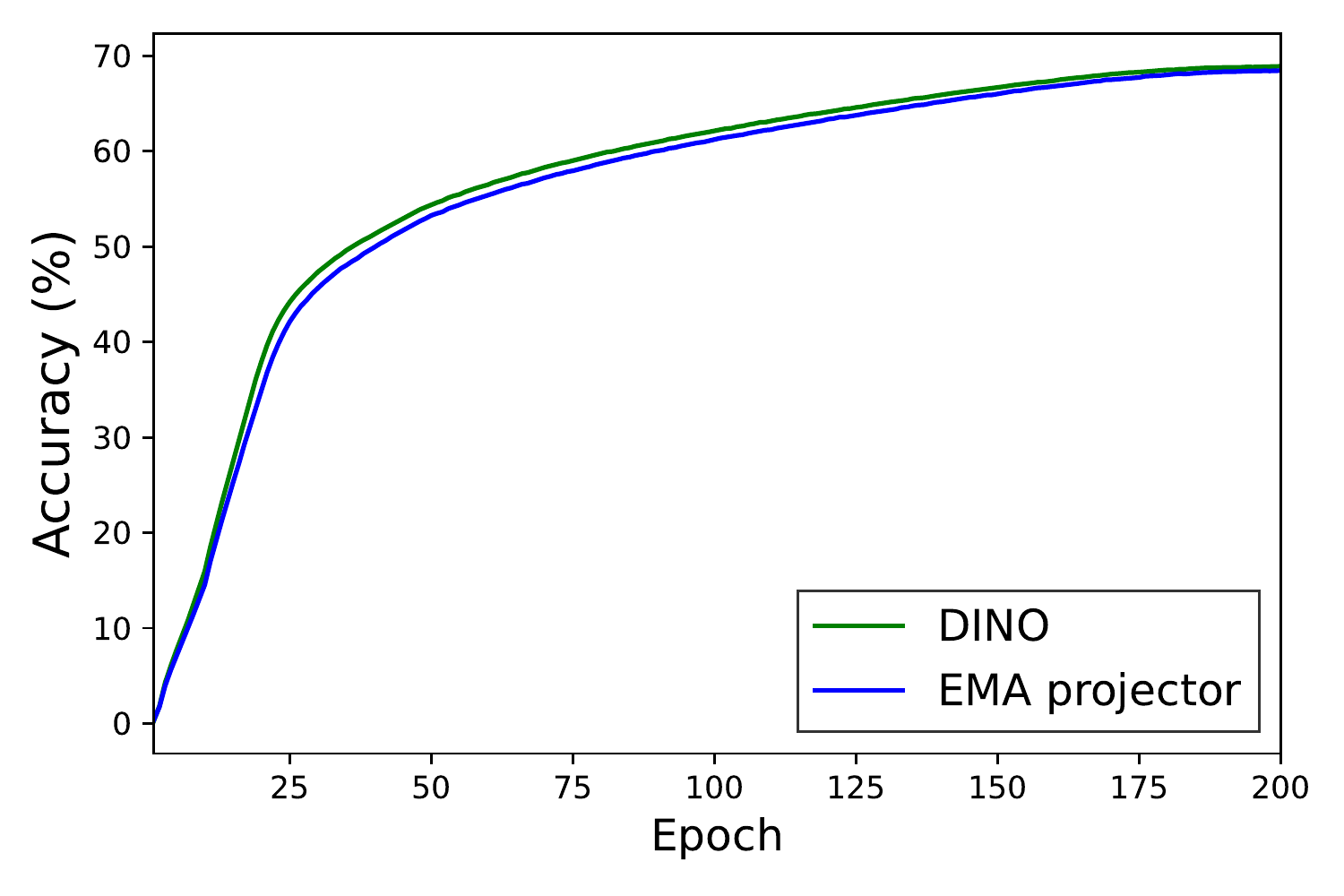}}
  \hfill
  \subfloat[Linear test accuracy] {\includegraphics[width=0.5\linewidth]{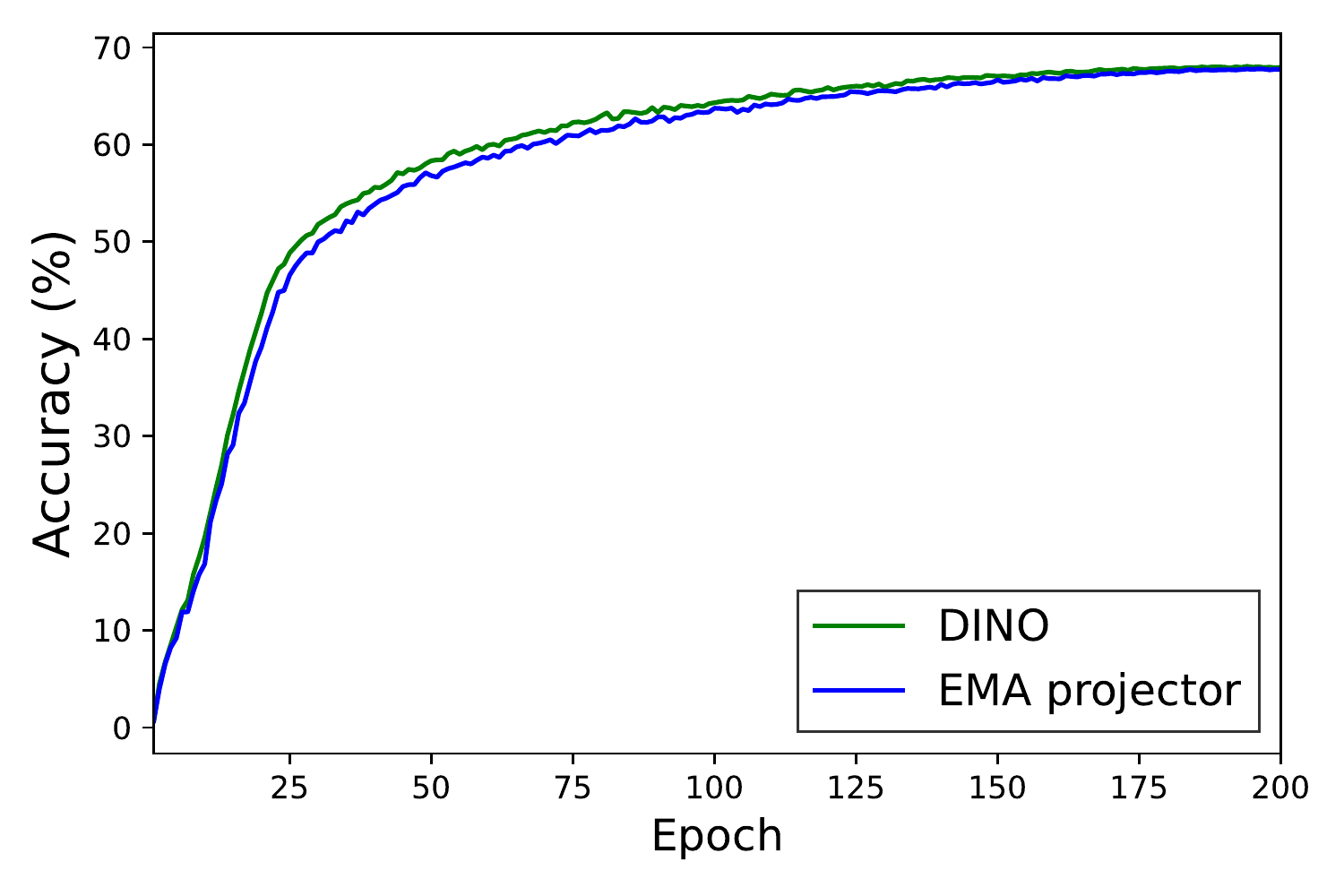}}
  \caption{\textbf{ImageNet-1K classification accuracy.} Linear accuracy (top-1 \%) for DINO and projector-only EMA in DINO. The accuracy gap is almost close to zero. It is best viewed in color.}
  \label{fig:dino_imagenet}
  \vspace{-10pt}
\end{figure}

Due to the resource constraint, we did not run for 1000 epochs, however, we believe, the gap for the longer training is almost zero as the trend in Fig. \ref{fig:dino_imagenet} shows that using EMA in the deep backbone is unnecessary, but EMA on the \textit{projector} is sufficient.

\textbf{Learning Curves of BYOL on CIFAR-100.}

\begin{figure}[!htbp]
  \vspace{-12pt}
  \centering
  \subfloat[Linear train accuracy] {\includegraphics[width=0.5\linewidth]{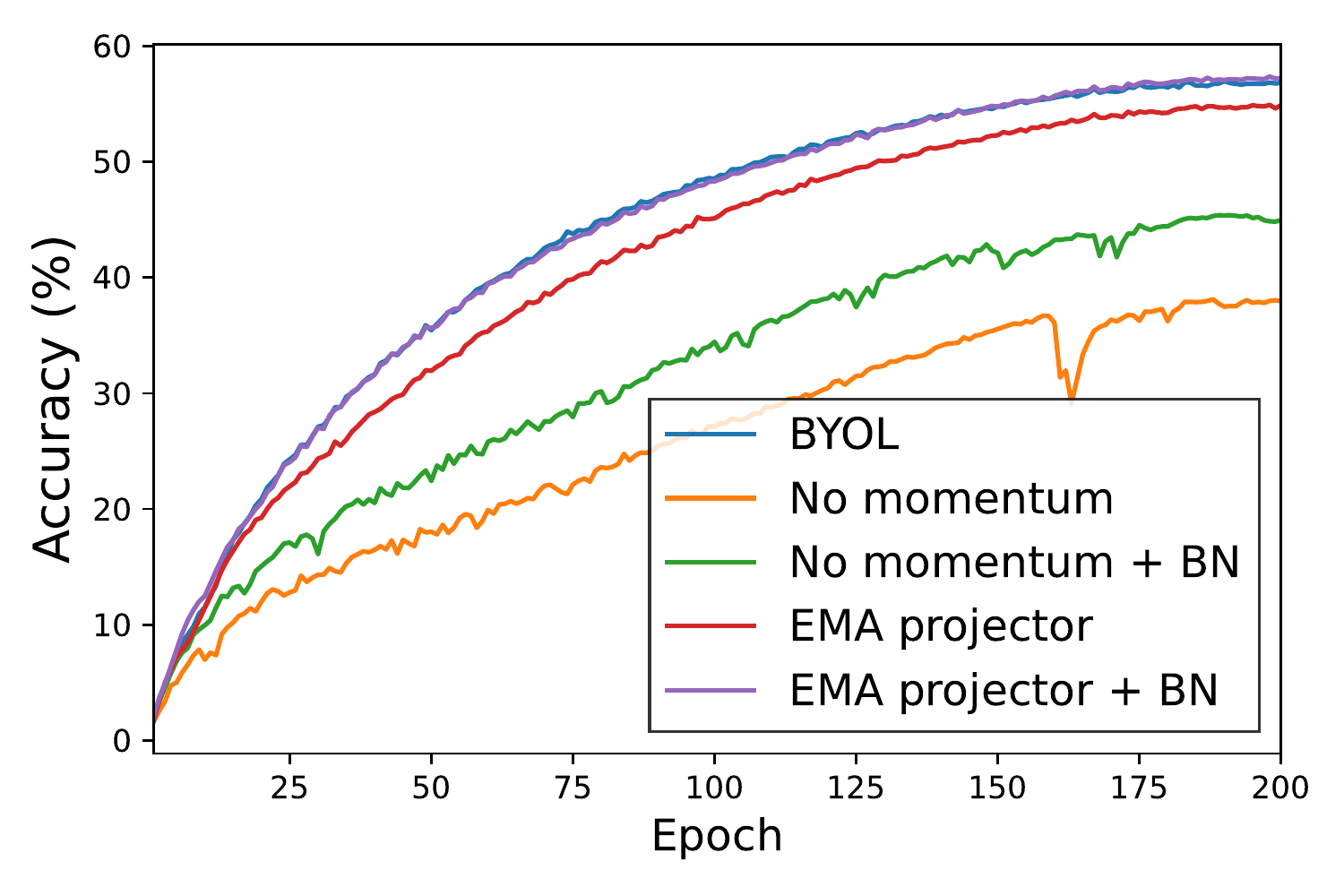}}
  \hfill
  \subfloat[Linear test accuracy] {\includegraphics[width=0.5\linewidth]{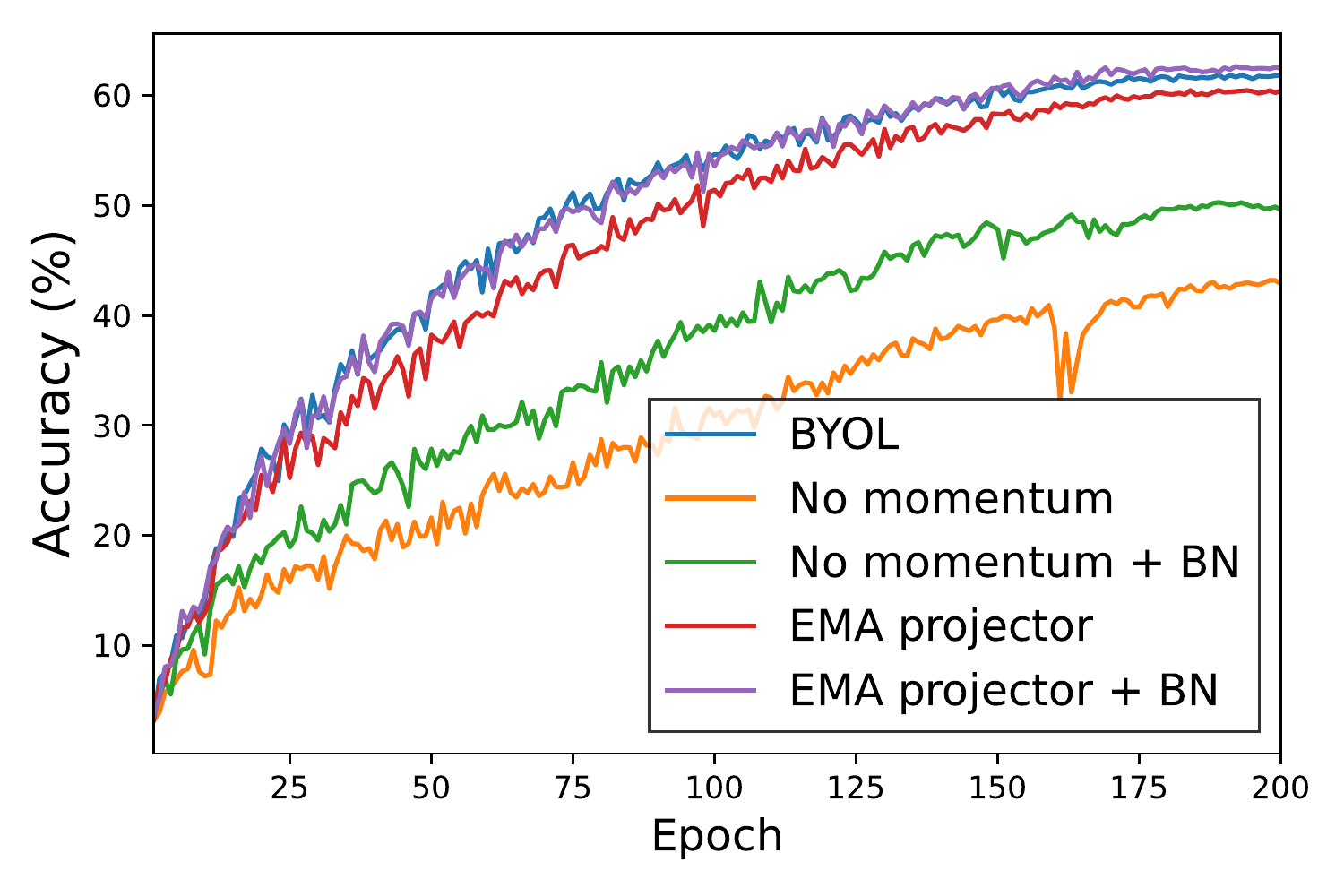}}
  \caption{\textbf{Increasing stability with BN.} Linear accuracy (top-1 \%) for BYOL and projector-only EMA in CIFAR-100. Here BN is applied to both outputs of \textit{block 4} and projector. It is best viewed in color.}
  \label{fig:byol_bn_ema}
  \vspace{-10pt}
\end{figure}

We show the case when applying projector-only momentum in the BYOL framework. As the analysis in the main paper, the final parts of the encoder have more fluctuation and less stability. We can use additionally a technique as BN to the output of block 4 or projector to help training more stable. As shown in Fig. \ref{fig:byol_bn_ema}, with \textit{no momentum} (\textcolor{orange}{orange line}) there always has the unstable point and showing the worst accuracy. Using BN to the final layers (block 4 + projector) can help to stabilize the training as the \textcolor{Green}{green line} that improves 6.7\% compared to the version without BN and EMA.

Our proposed projector-only EMA can help to stabilize and boost performance to be competitive with baseline BYOL to a smaller gap. Adding BN to the output of block 4 and the output of the projector can make projector-only EMA (\textcolor{violet}{purple line}) almost match the BYOL baseline as clearly shown in Fig. \ref{fig:byol_bn_ema}. In this figure, EMA projector + BN means that BN is applied to both \textit{block 4} and the \textit{projector}.

Tab. \ref{tab:byol_ema_bn} shows a more detail about the important of making the \textit{block 4} or \textit{projector} stable with BN. It demonstrates that our proposed \textit{projector-only EMA} can match the BYOL baseline while avoiding the double forward computation overhead to the deep backbone. It shows more evidence for our analysis in the main paper that the final layers (block 4, projector) have much more fluctuation, \ie less stability. Making the final layers learn from the past (with EMA) to stabilize the training is crucial, and a simple technique such as BN can further help.

\section{More Results on CIFAR-10}
We show more results on CIFAR-10 where BYOL is compared with our \textit{projector-only momentum}. Tab. \ref{tab:ame_components_cifar10} shows that our method can also match baseline without using EMA to the deep backbone that can maintain the benefit of EMA while avoiding the double computation overhead to the backbone as the traditional method.

\begin{table}[!htbp]
    \caption{\textbf{Linear classification accuracy on CIFAR-10}. We train all SSL methods for 200 epochs, with the ResNet-18 backbone. ``-'' denotes \textit{sharing} weight. BN1, BN2 denote the BN applying to the output of \textit{block 4} and \textit{projector}, respectively. We report top-1 accuracy on the test set.}
    \label{tab:ame_components_cifar10}
    \centering
    \smallskip
    \resizebox{1.0\hsize}{!}{
    \begin{tabular}{cccc}
    \toprule
    Method & Backbone Momentum (18 layers) & Projector Momentum (2 layers) & Acc (\%) \\
    \midrule
    BYOL (baseline) & \checkmark & \checkmark & 88.37 \\
    w/o momentum & - & - & \textcolor{red}{80.34} \\
    EMA Projector & - & \checkmark & \textcolor{blue}{87.73} \\
    EMA Projector + BN1 & - & \checkmark & \textcolor{blue}{87.91} \\
    EMA Projector + BN2 & - & \checkmark & \textcolor{blue}{88.03} \\
    EMA Projector + BN1 + BN2 & - & \checkmark & \textcolor{blue}{88.45} \\
    \bottomrule
    \end{tabular} }
\end{table}

\section{More Visualizations of Filters}
Beyond the filters that are visualized in the main paper, we also presented 16 more filters in 2D and 3D visualizations out of 256 filters (weights) for the experiments in section 3.2 where EMA is applied only to the final FC layer of the \textit{target} encoder. Here, \textit{start point, end} in the figures meaning that are the beginning and final point of the training process (a total of 40k steps), respectively.

\subsection{3D visualization}
We show more 3D visualizations where \textcolor{Green}{\textit{green}} line denotes \textit{no momentum $\beta=0$}; \textcolor{blue}{\textit{blue}} and \textcolor{red}{\textit{red}} lines show the \textit{student} and \textit{teacher} with momentum $\beta=0.9$; \textit{black} and \textcolor{orange}{\textit{orange}} lines illustrate the \textit{student} and \textit{teacher} with momentum $\beta=0.99$. We can see several filters are stable for both curves and there are filters that have the unstable point (at step around 10k) for the \textit{no momentum} line, such as \textit{filter 001, 002, 005, 008, 011, 013, 014} (Fig. 3 - Fig. 10). We also provide 2D plots (Fig. 11 - Fig. 18) below which can show more clearly than some cases of 3D plots.

We observe that if using momentum, all filters show a mostly stable curve, \ie no `spike' or abnormal point suddenly changes as the case \textit{no momentum}.

\begin{figure}[!htbp]
\captionsetup[subfigure]{labelformat=empty}
  \vspace{-12pt}
  \centering
  \subfloat[Filter 001] {\includegraphics[width=0.48\linewidth]{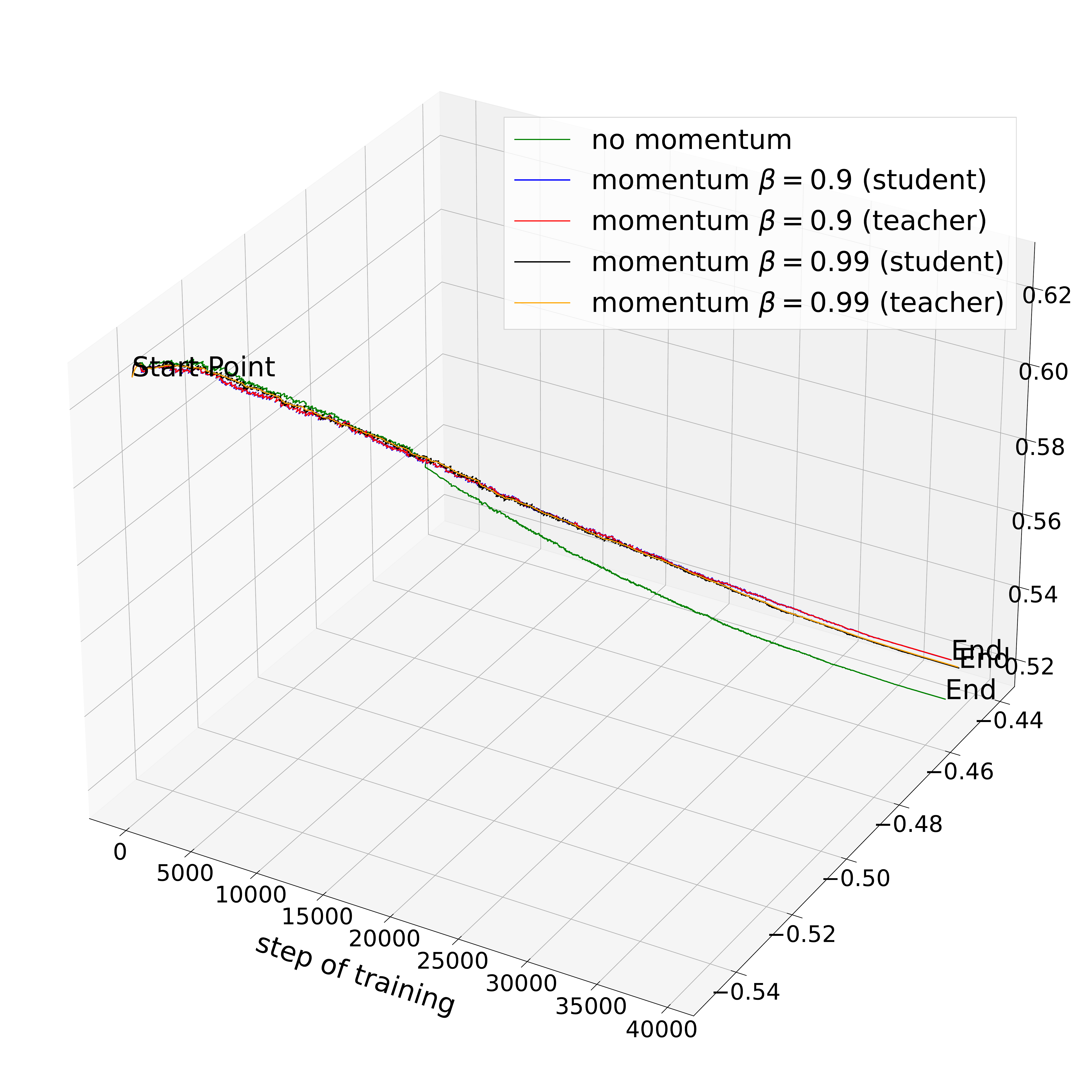}}
  \hfill
  \subfloat[Filter 002] {\includegraphics[width=0.48\linewidth]{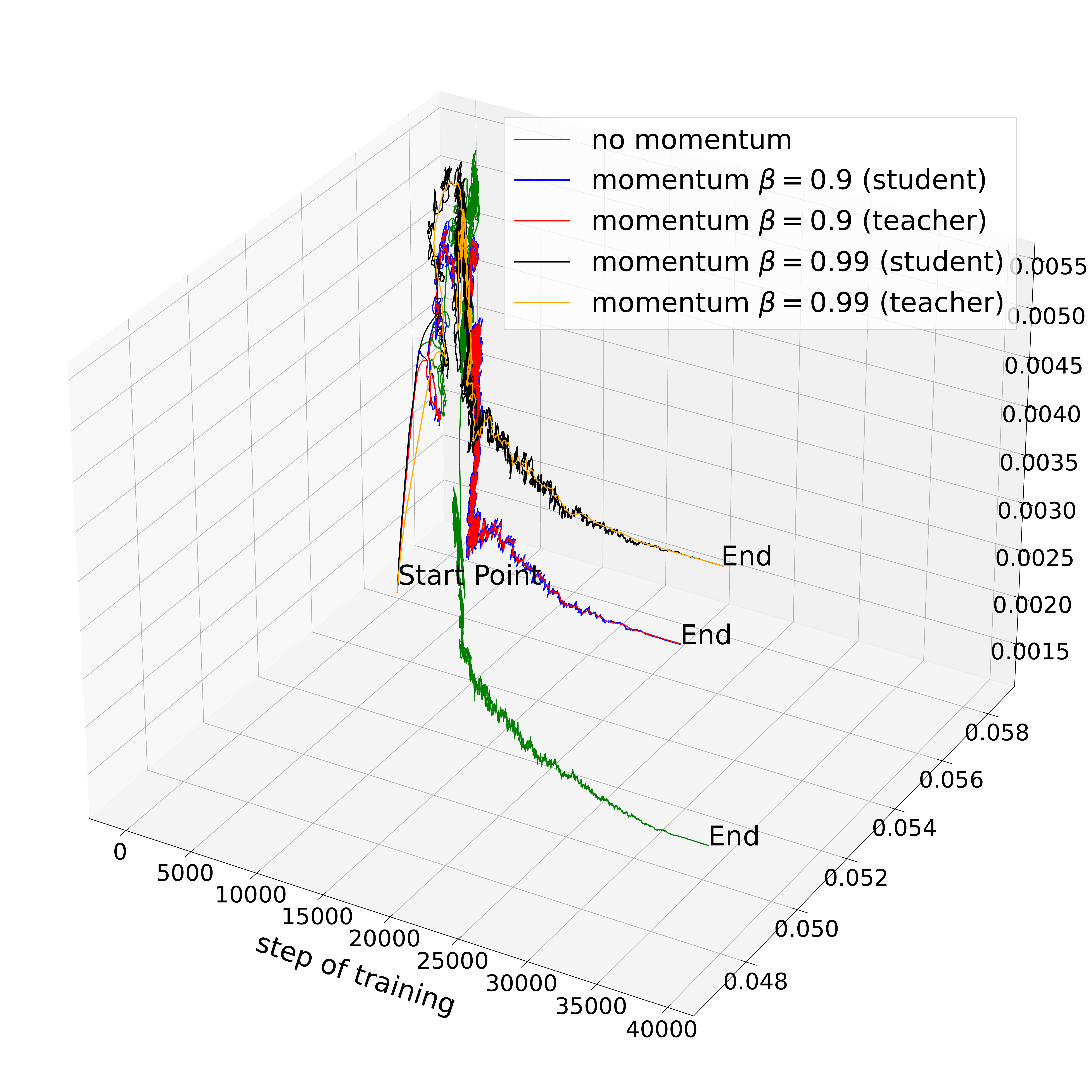}}
  \caption{}
  \vspace{-10pt}
\end{figure}

\begin{figure}[!htbp]
\captionsetup[subfigure]{labelformat=empty}
  \vspace{-20pt}
  \centering
  \subfloat[Filter 003] {\includegraphics[width=0.48\linewidth]{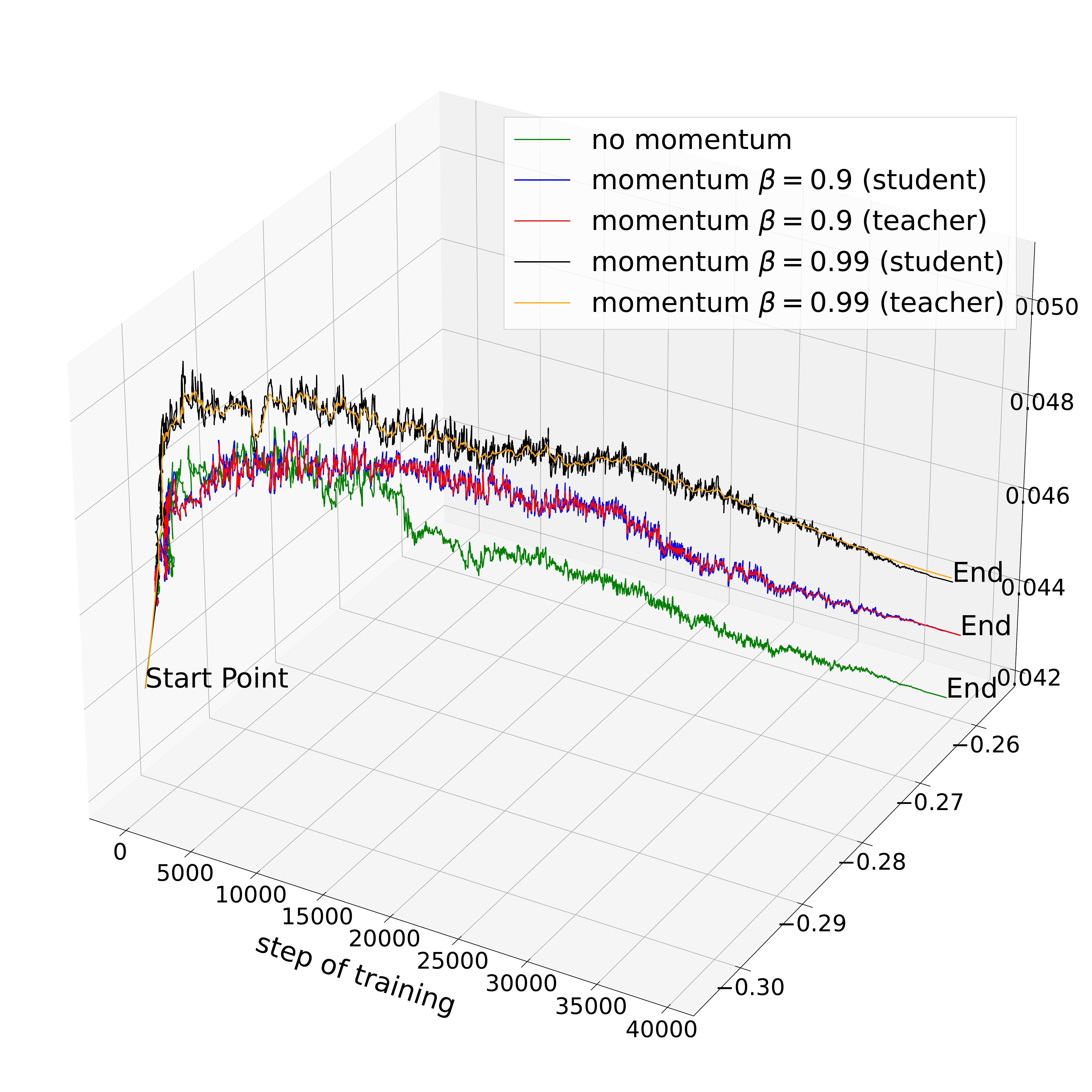}}
  \hfill
  \subfloat[Filter 004] {\includegraphics[width=0.48\linewidth]{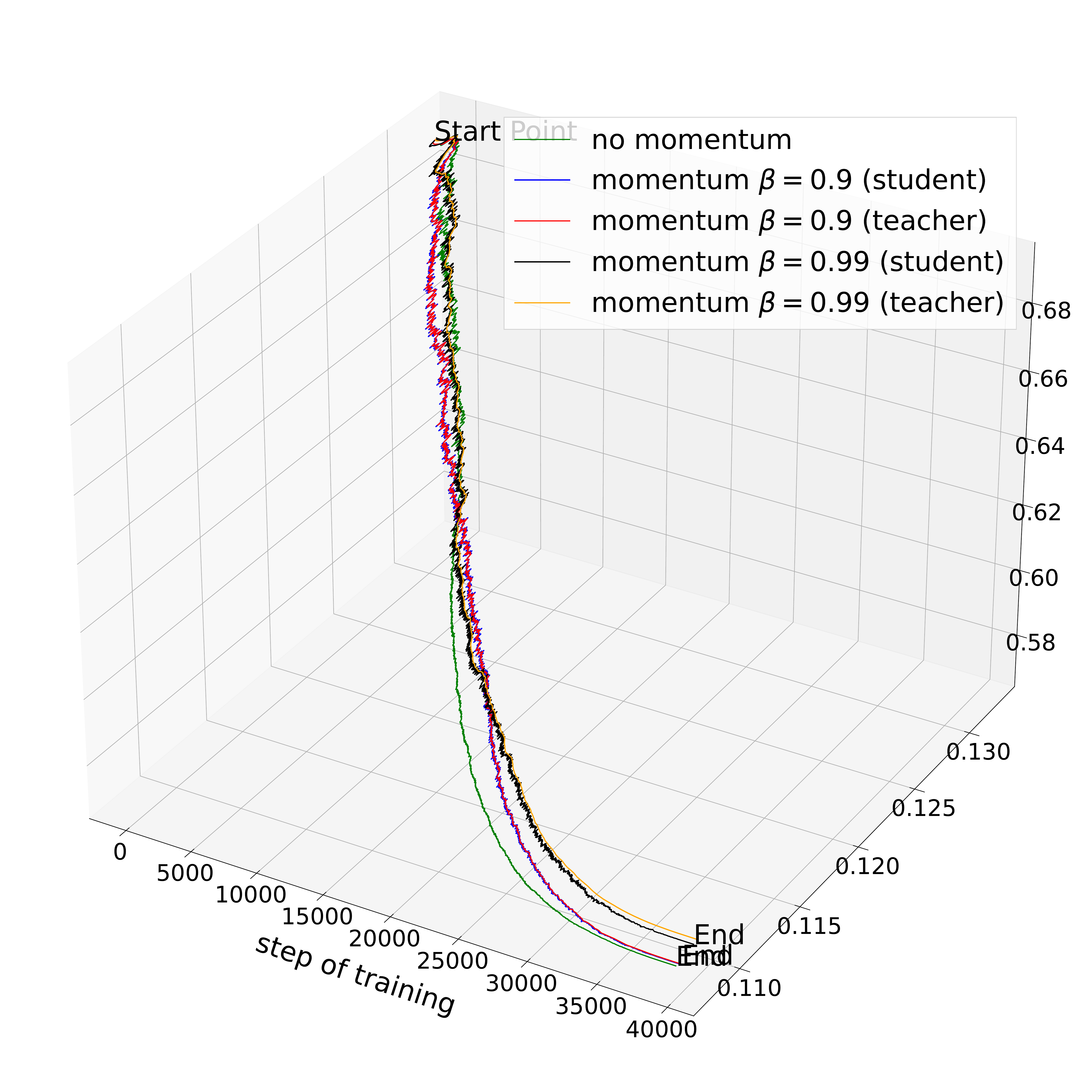}}
  \caption{}
  \vspace{-20pt}
\end{figure}


\begin{figure}[!htbp]
  \vspace{-20pt}
  \centering
  \subfloat[Filter 007] {\includegraphics[width=0.48\linewidth]{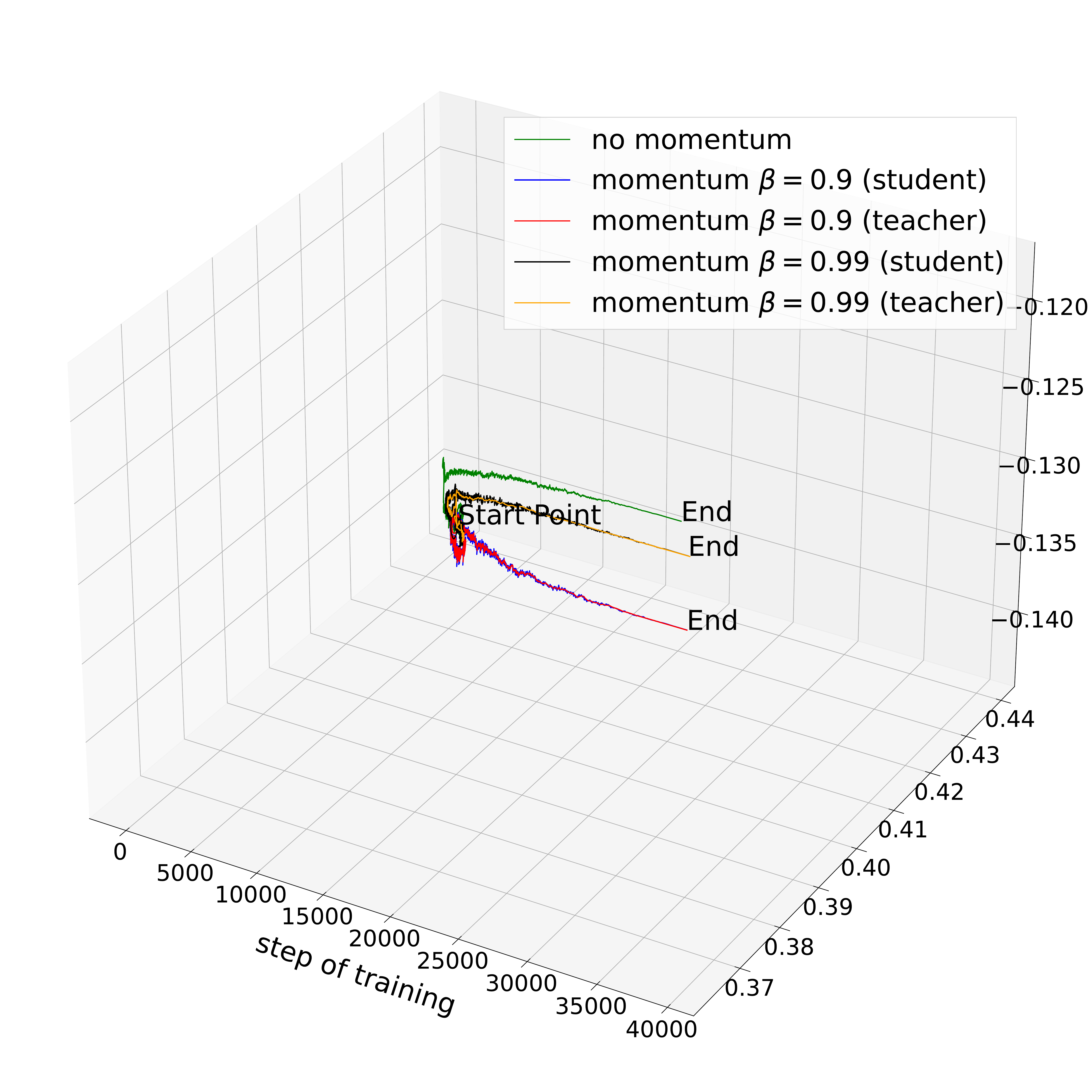}}
  \hfill
  \subfloat[Filter 008] {\includegraphics[width=0.48\linewidth]{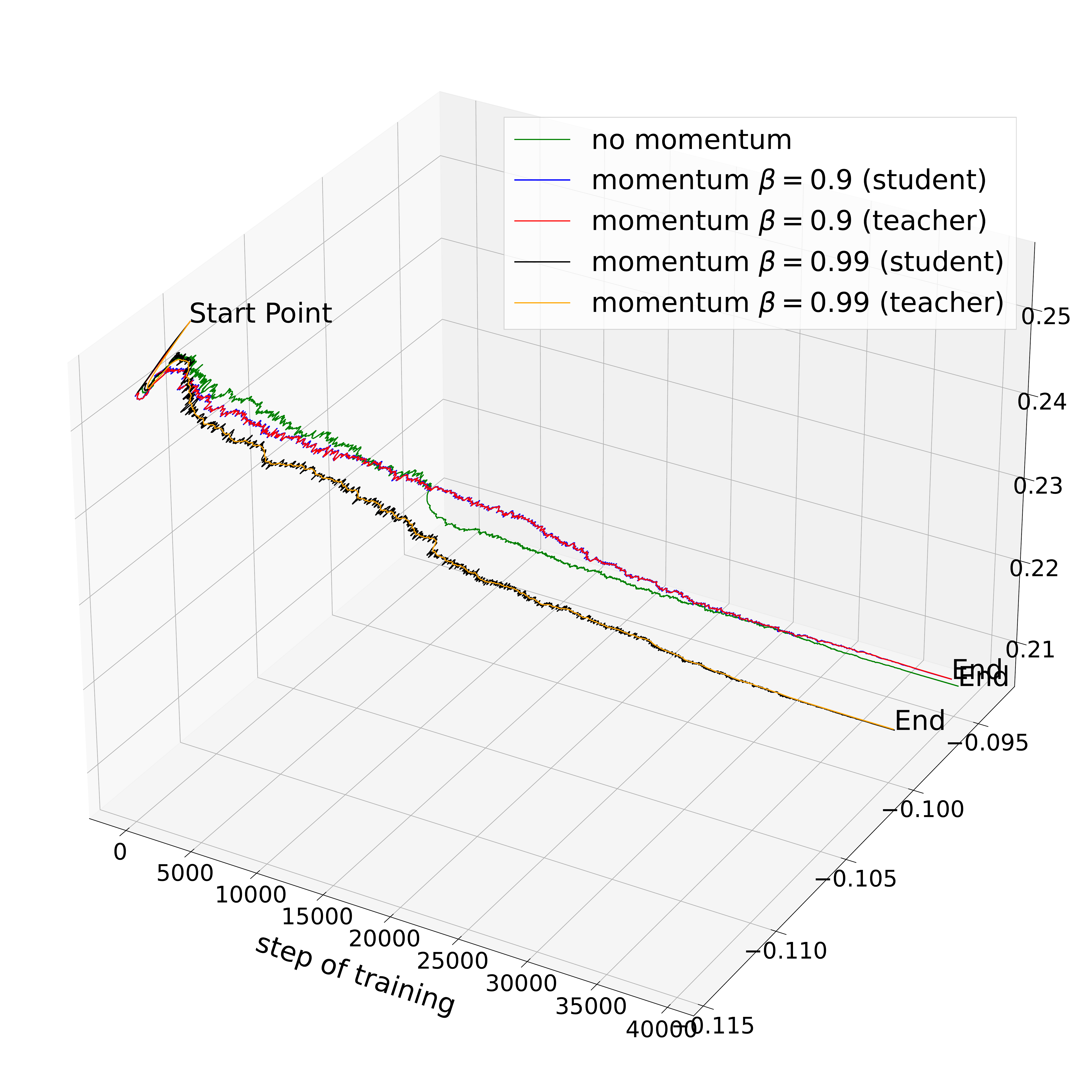}}
  \caption{}
  \vspace{-10pt}
\end{figure}


\begin{figure}[!htbp]
\captionsetup[subfigure]{labelformat=empty}
  \vspace{-12pt}
  \centering
  \subfloat[Filter 011] {\includegraphics[width=0.48\linewidth]{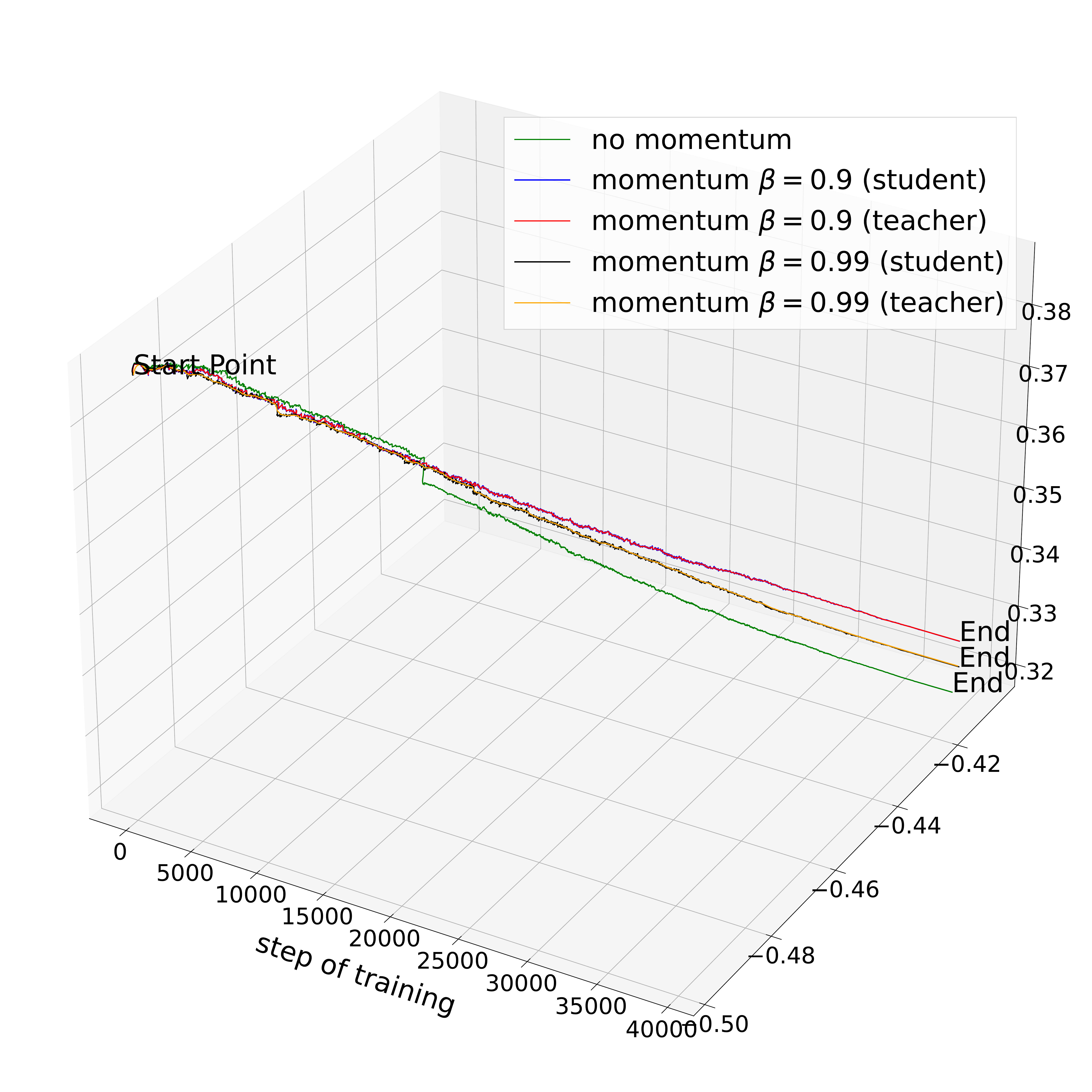}}
  \hfill
  \subfloat[Filter 012] {\includegraphics[width=0.48\linewidth]{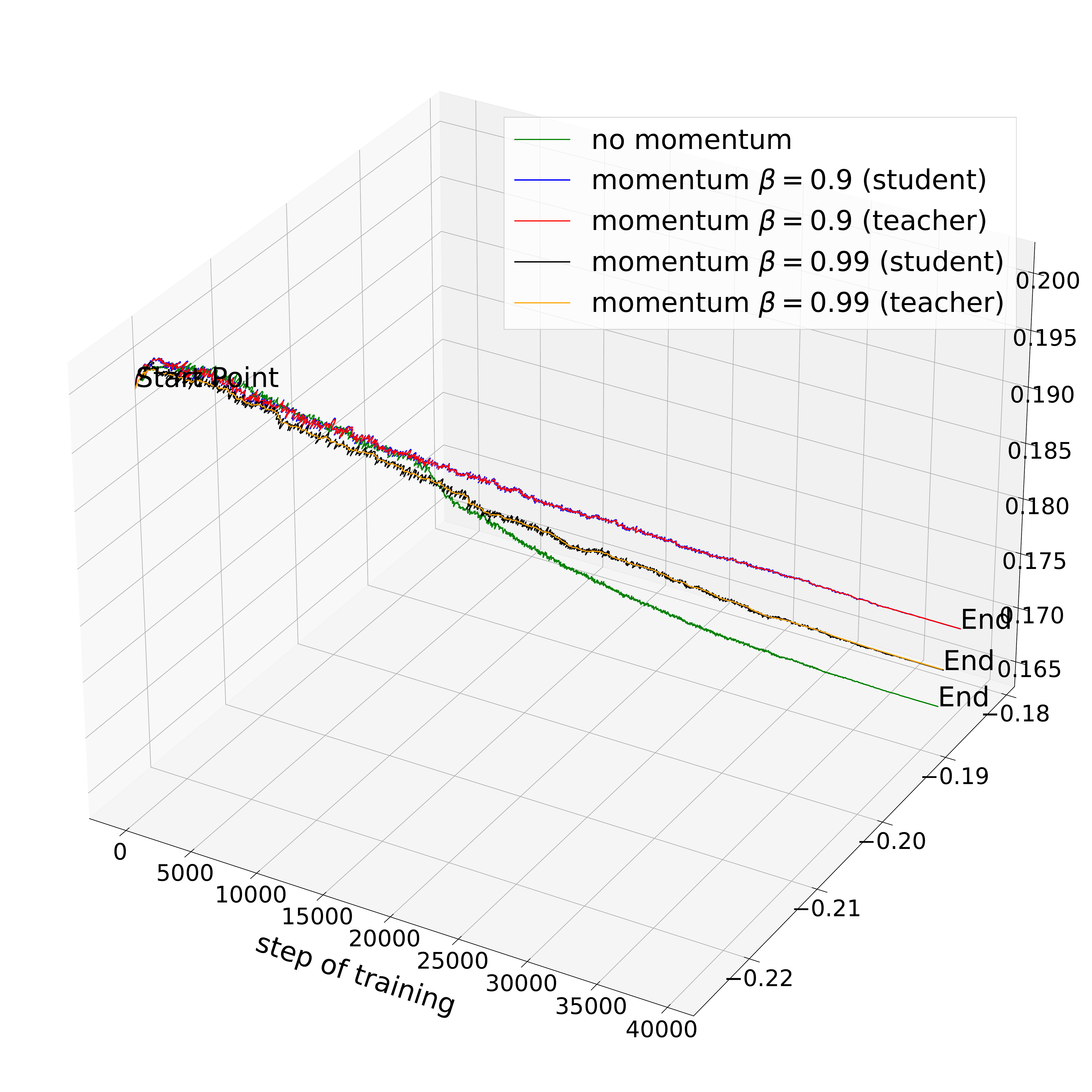}}
  \caption{}
  \vspace{-10pt}
\end{figure}

\begin{figure}[!htbp]
\captionsetup[subfigure]{labelformat=empty}
  \vspace{-12pt}
  \centering
  \subfloat[Filter 013] {\includegraphics[width=0.48\linewidth]{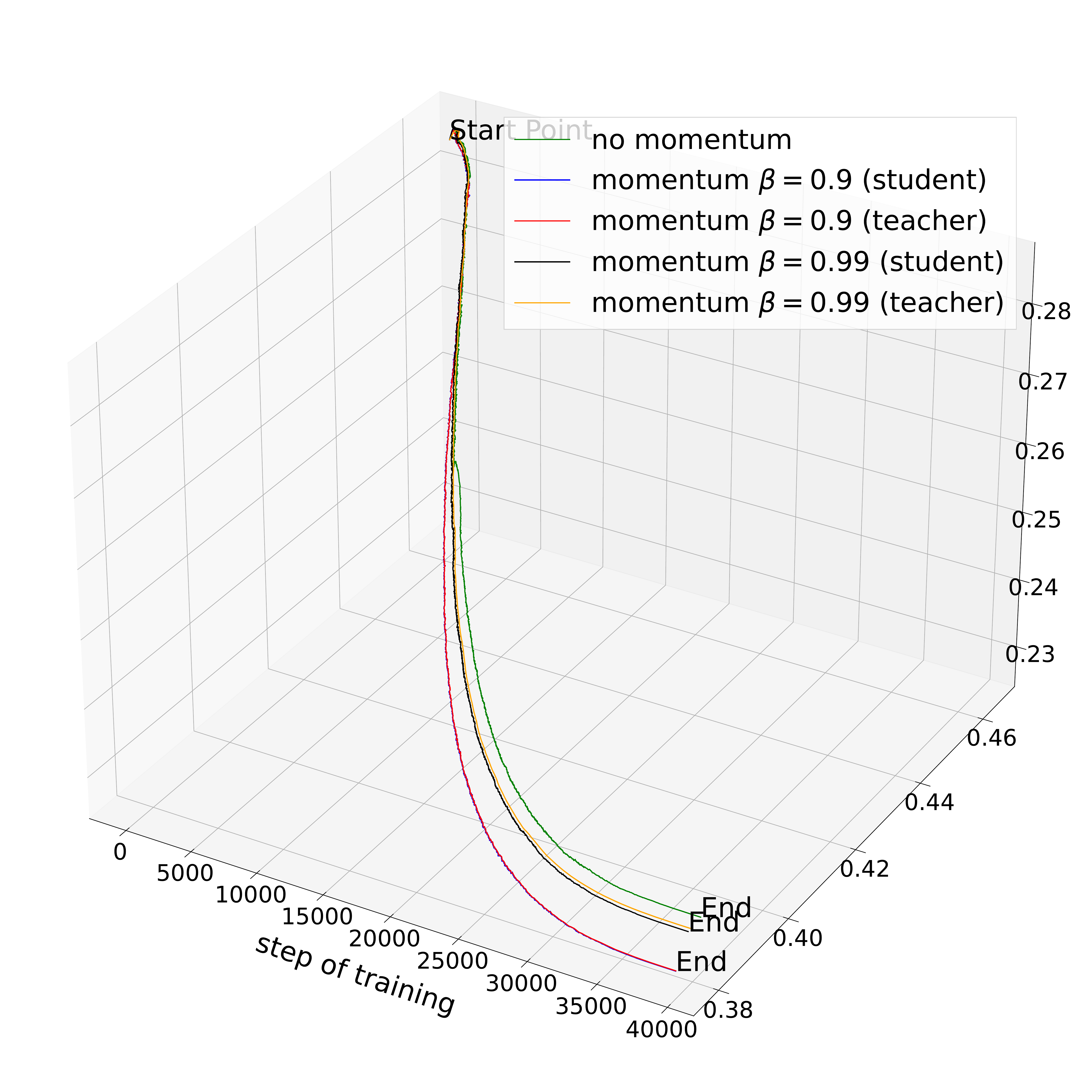}}
  \hfill
  \subfloat[Filter 014] {\includegraphics[width=0.48\linewidth]{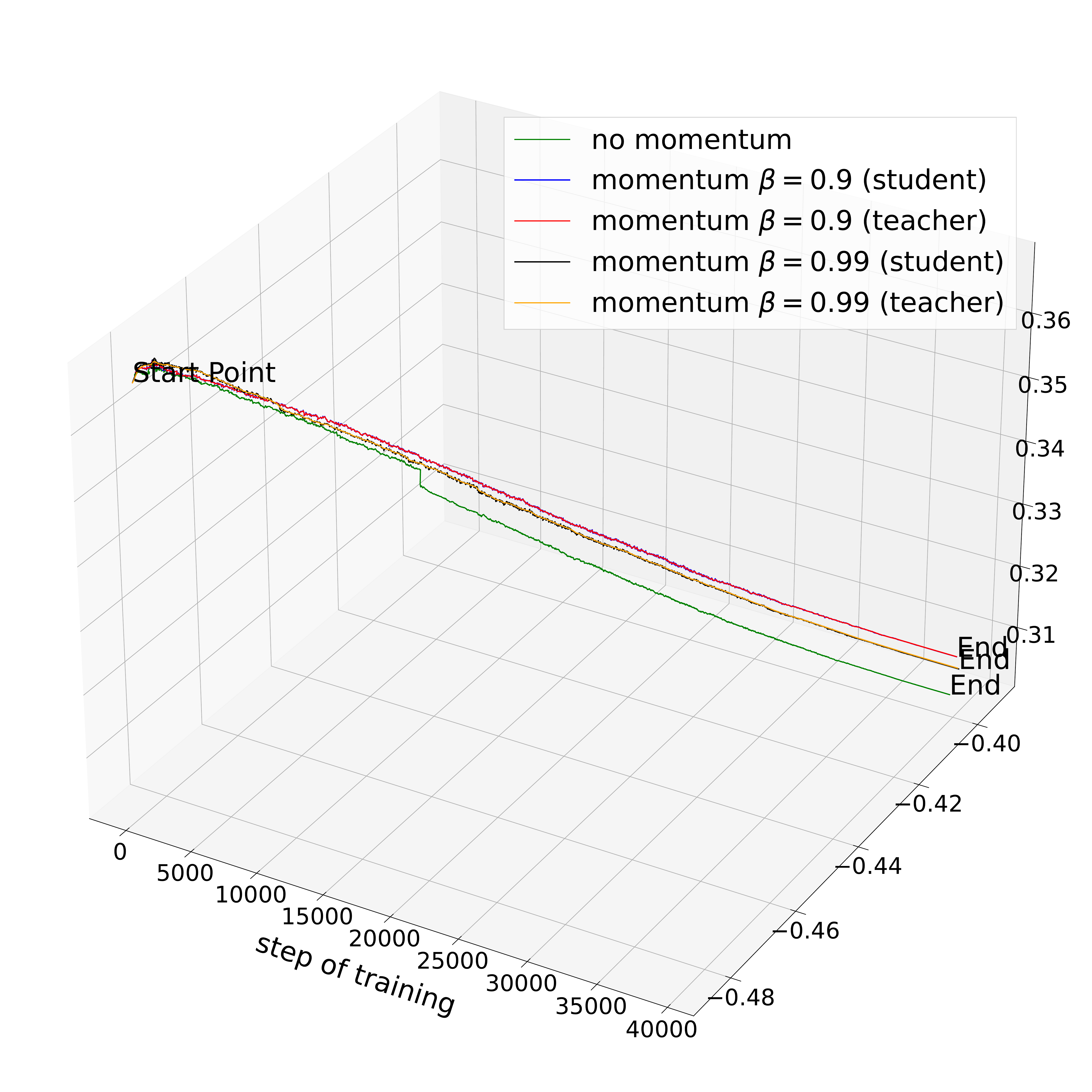}}
  \caption{}
  \vspace{-10pt}
\end{figure}


\subsection{2D visualization}
We also plot filters in 2D where the time dimension (\textit{training step}) is omitted for the more simple view as shown in Fig. 11 - Fig. 18 below. The 2D plots also show clearly that the \textcolor{Green}{green line} of \textit{no momentum} exhibits the unstable learning curves, such as \textit{Filter 001, 002, 007, 010, 011, 014}. By contrast, learning with EMA shows much more stable trends for every filter.

\begin{figure}[!htbp]
\captionsetup[subfigure]{labelformat=empty}
  \vspace{-12pt}
  \centering
  \subfloat[Filter 001] {\includegraphics[width=0.48\linewidth]{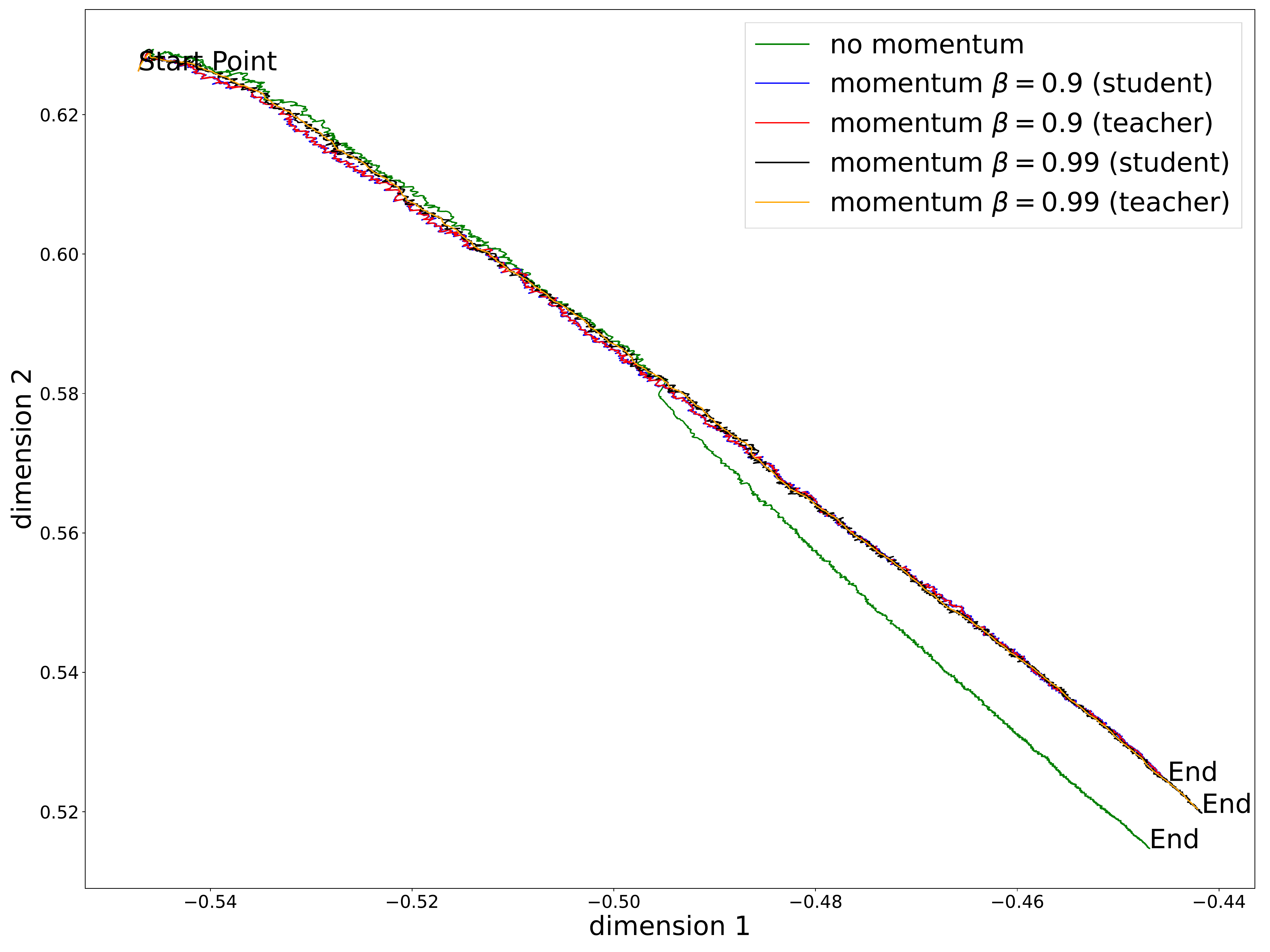}}
  \hfill
  \subfloat[Filter 002] {\includegraphics[width=0.48\linewidth]{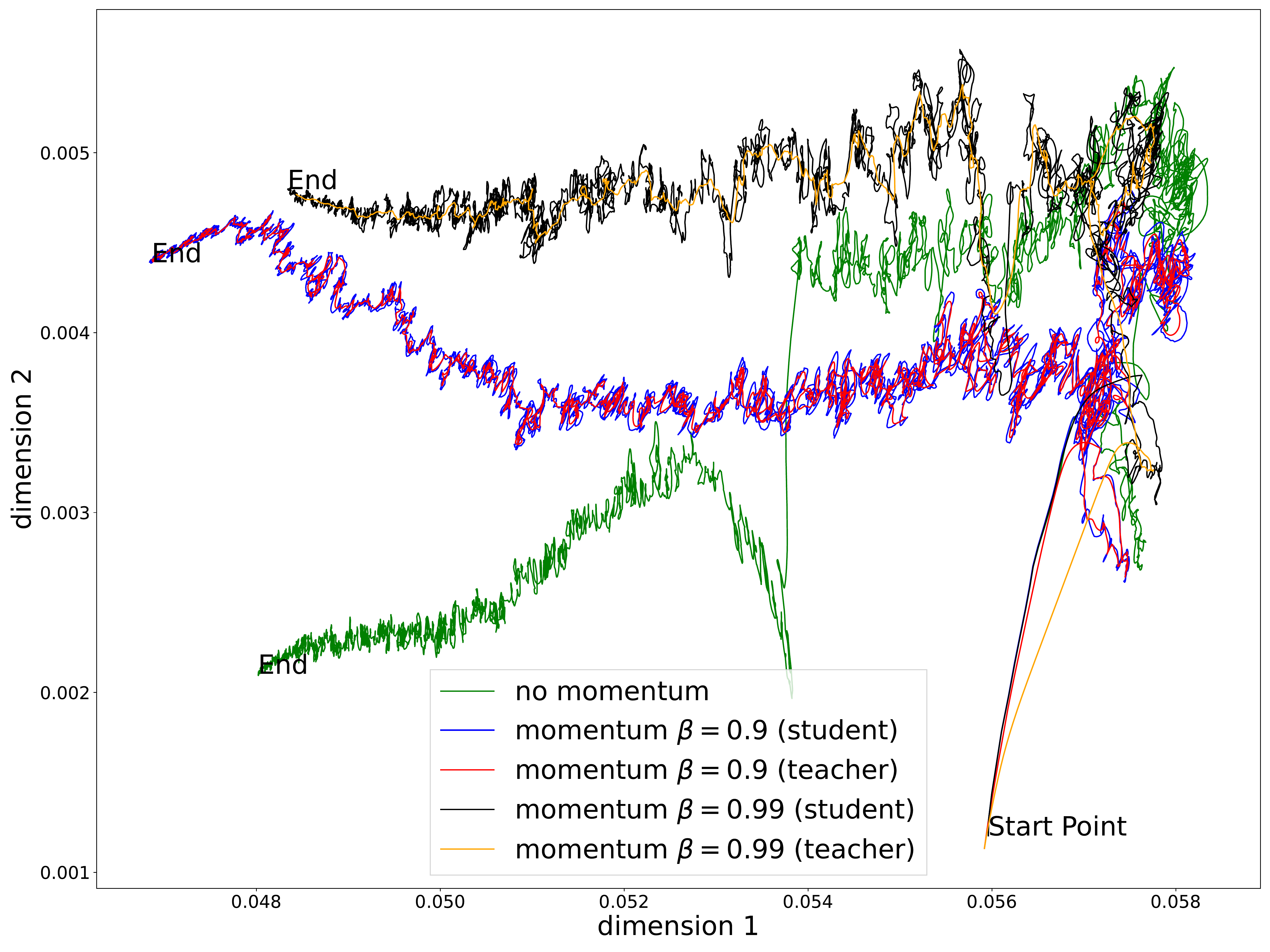}}
  \caption{}
  \vspace{-10pt}
\end{figure}


\begin{figure}[!htbp]
\captionsetup[subfigure]{labelformat=empty}
  \vspace{-12pt}
  \centering
  \subfloat[Filter 005] {\includegraphics[width=0.48\linewidth]{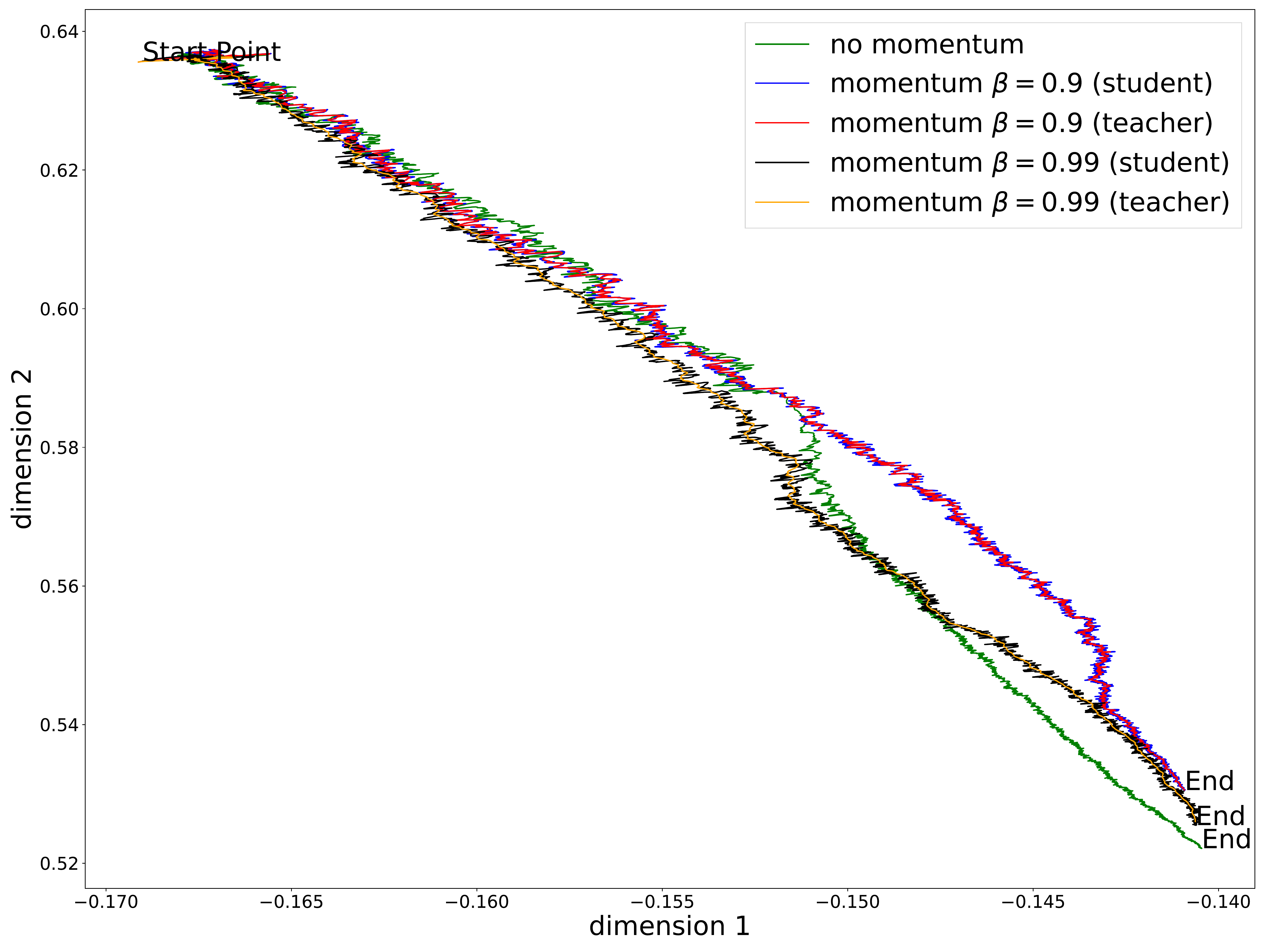}}
  \hfill
  \subfloat[Filter 006] {\includegraphics[width=0.48\linewidth]{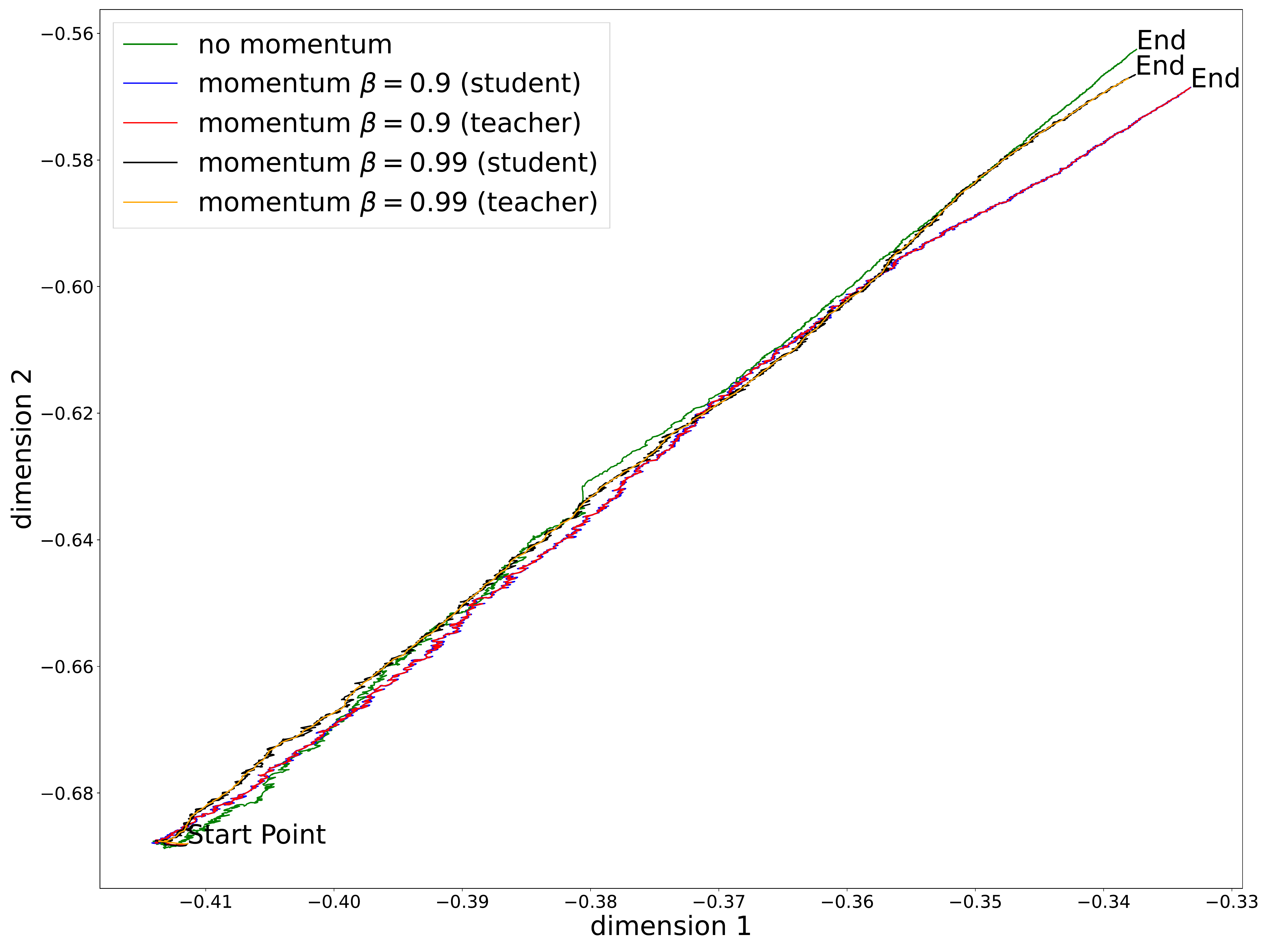}}
  \caption{}
  \vspace{-10pt}
\end{figure}

\begin{figure}[!htbp]
\captionsetup[subfigure]{labelformat=empty}
  \vspace{-12pt}
  \centering
  \subfloat[Filter 007] {\includegraphics[width=0.48\linewidth]{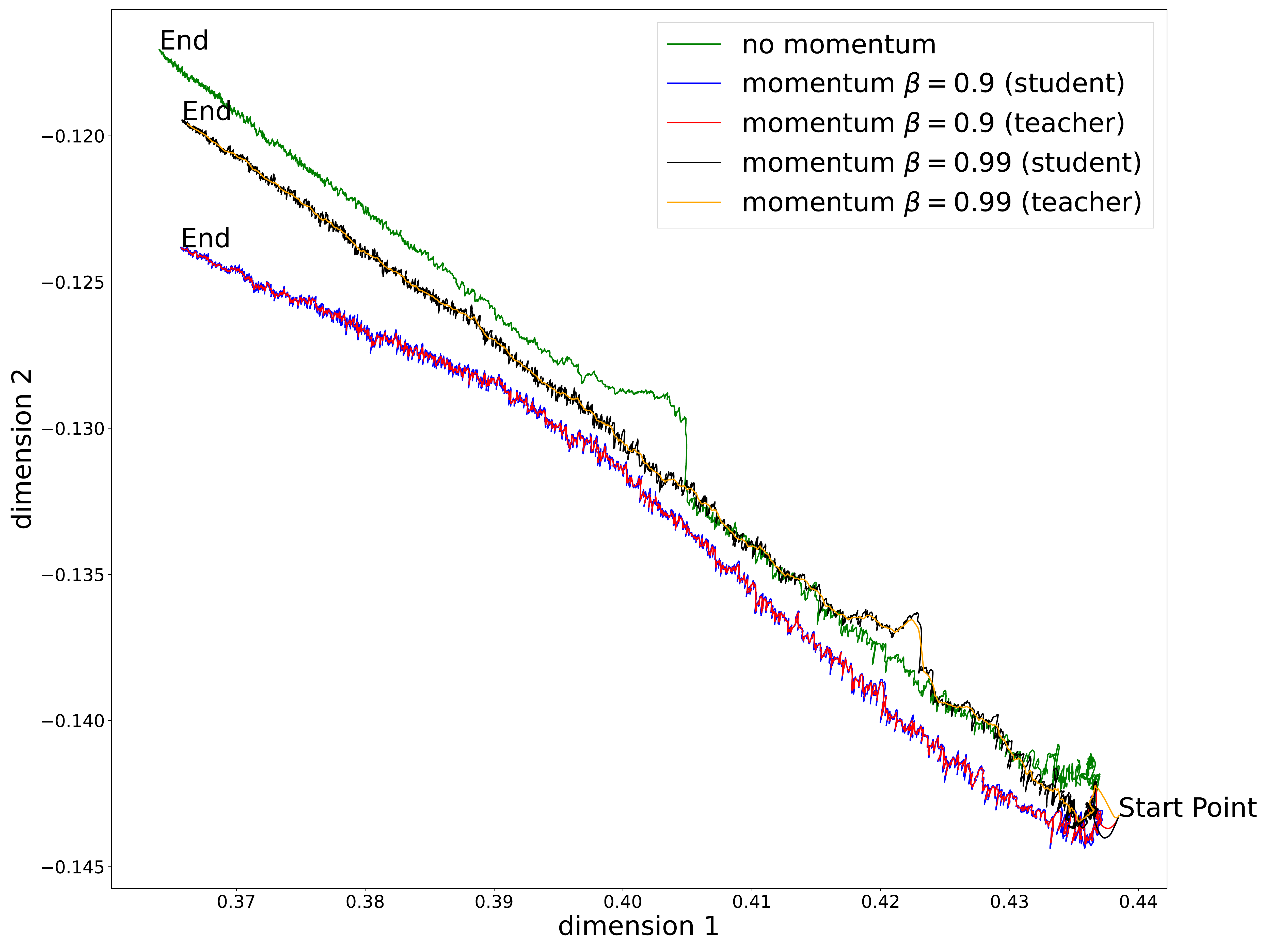}}
  \hfill
  \subfloat[Filter 008] {\includegraphics[width=0.48\linewidth]{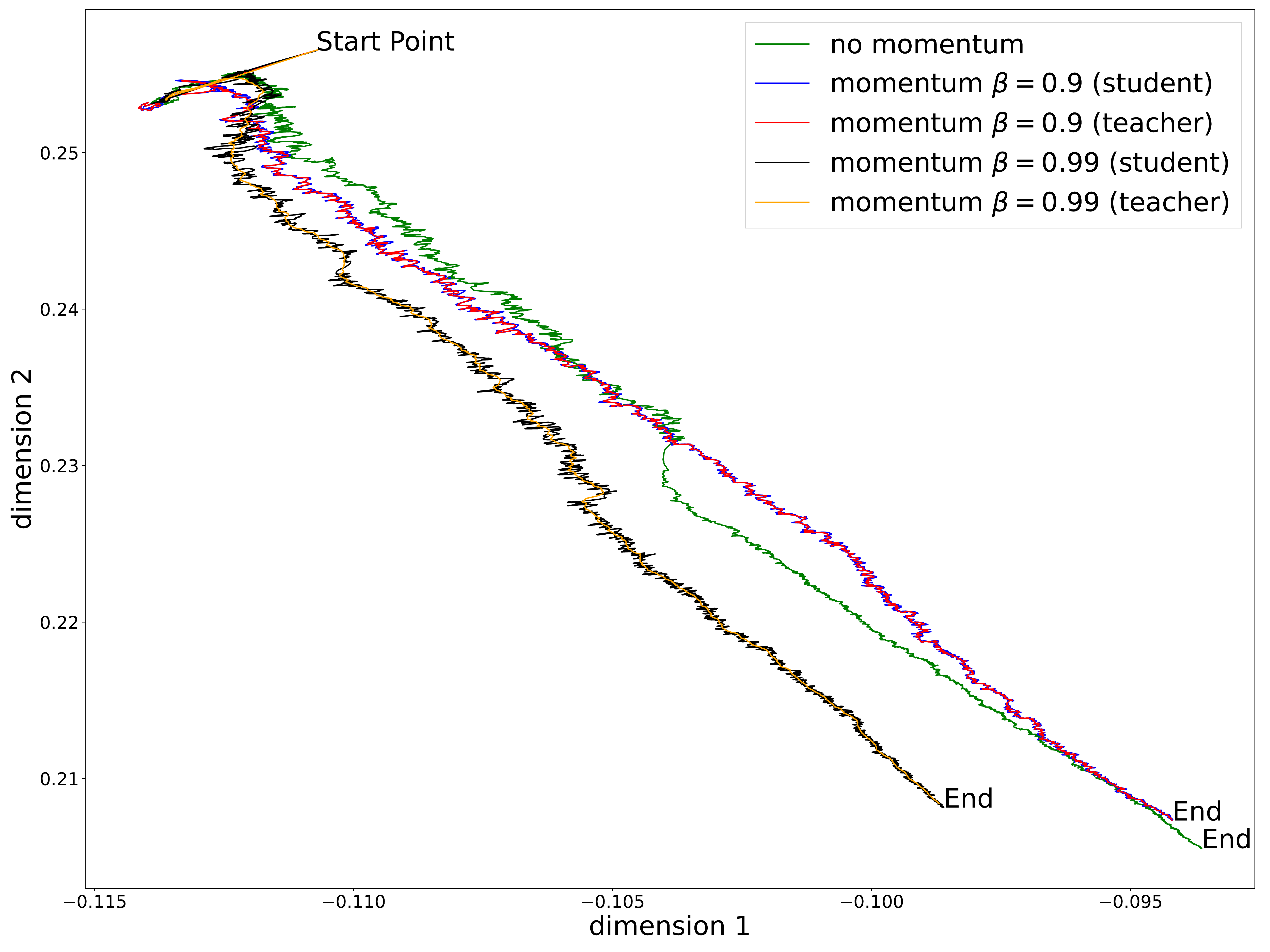}}
  \caption{}
  \vspace{-10pt}
\end{figure}

\begin{figure}[!htbp]
\captionsetup[subfigure]{labelformat=empty}
  \vspace{-12pt}
  \centering
  \subfloat[Filter 009] {\includegraphics[width=0.48\linewidth]{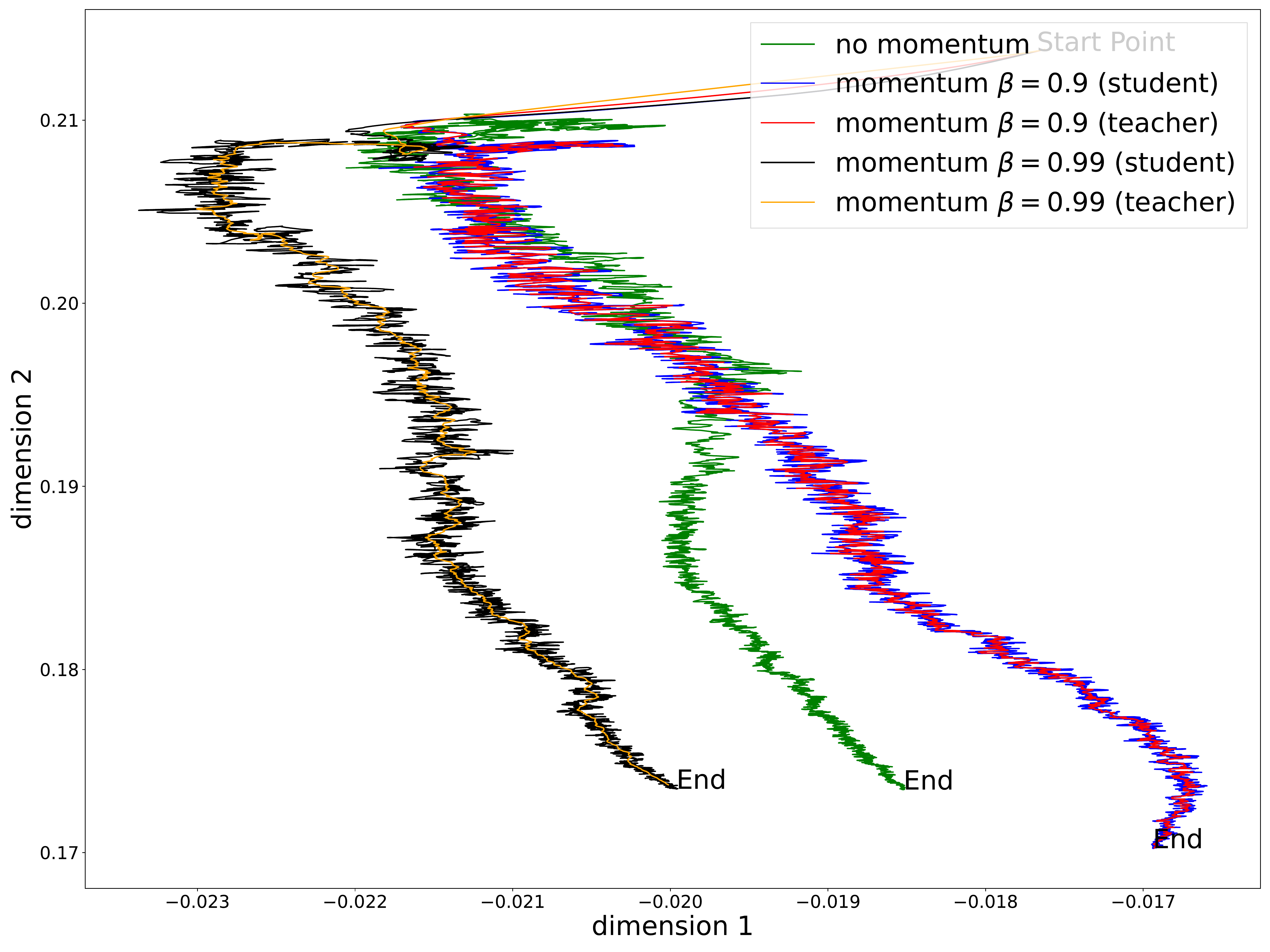}}
  \hfill
  \subfloat[Filter 010] {\includegraphics[width=0.48\linewidth]{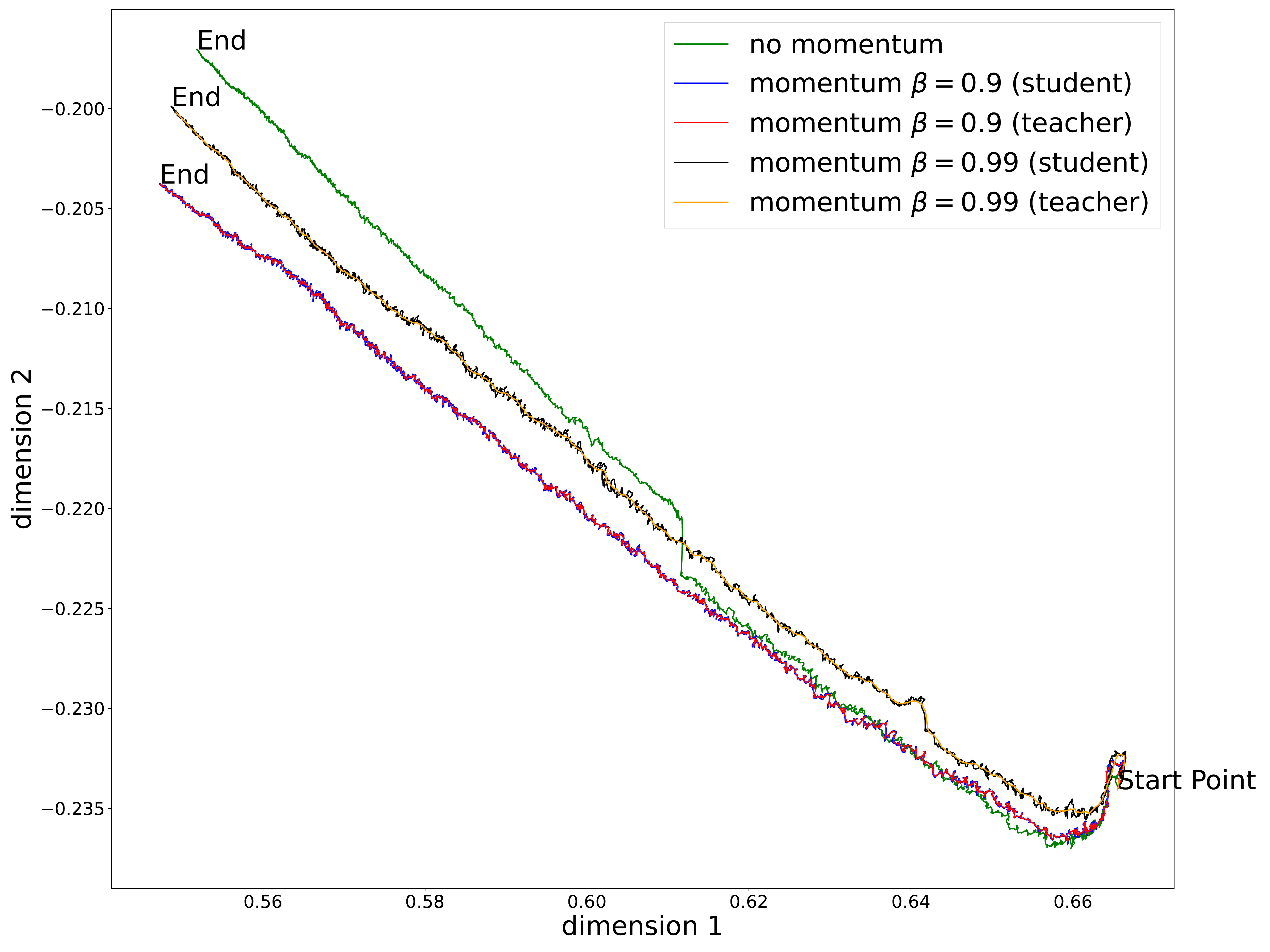}}
  \caption{}
  \vspace{-10pt}
\end{figure}

\begin{figure}[!htbp]
\captionsetup[subfigure]{labelformat=empty}
  \vspace{-12pt}
  \centering
  \subfloat[Filter 011] {\includegraphics[width=0.48\linewidth]{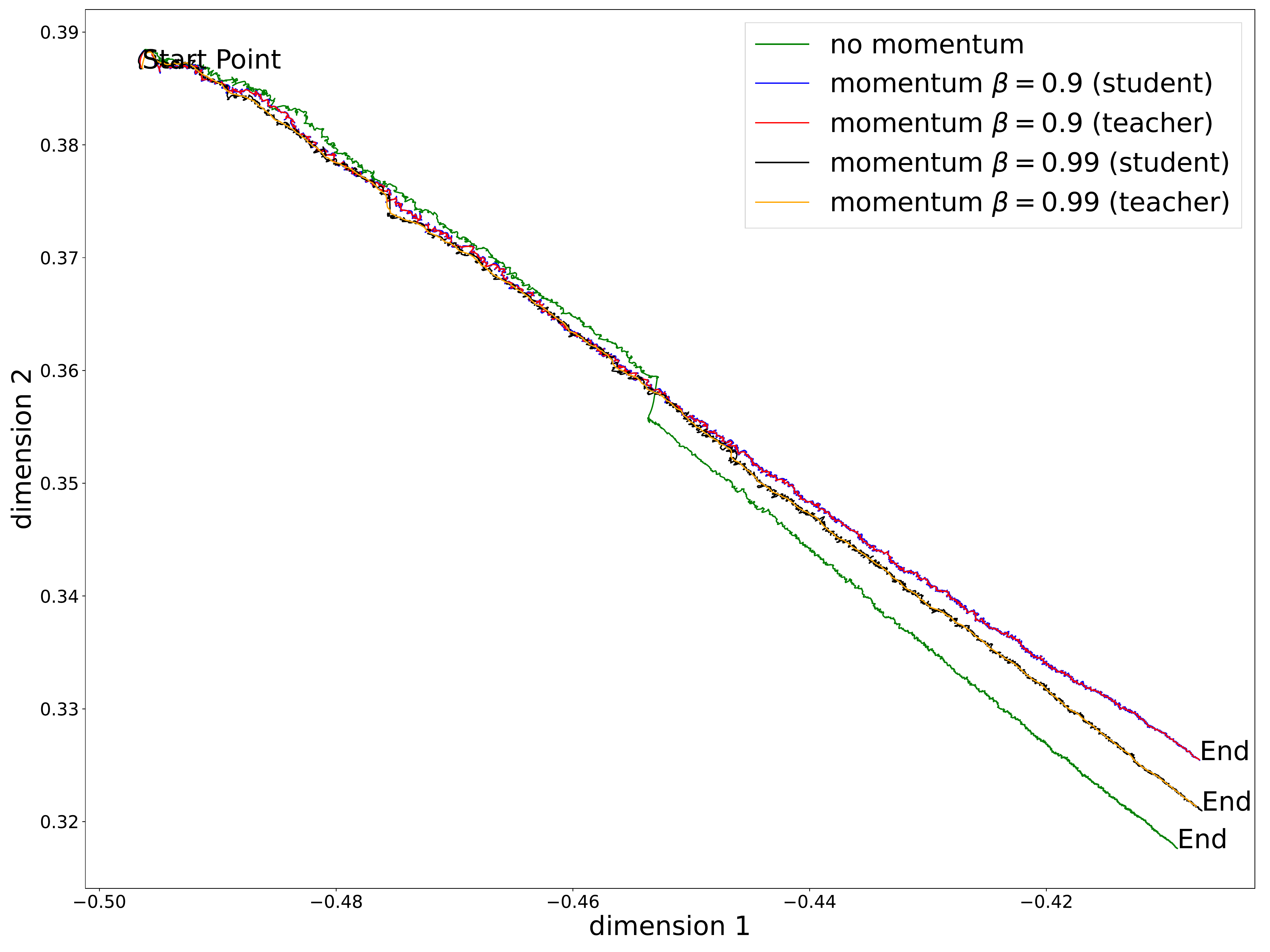}}
  \hfill
  \subfloat[Filter 012] {\includegraphics[width=0.48\linewidth]{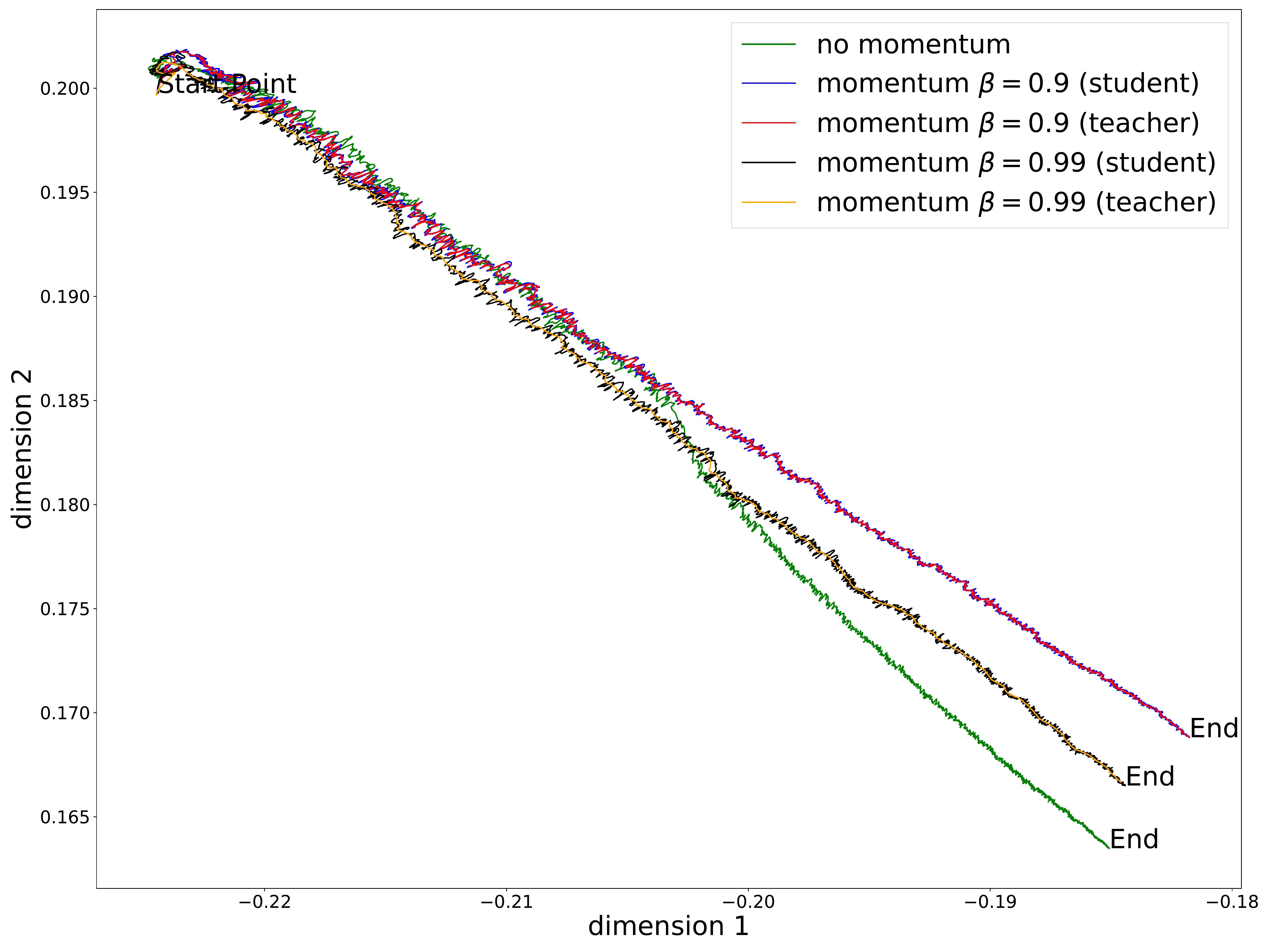}}
  \caption{}
  \vspace{-10pt}
\end{figure}

\begin{figure}[!htbp]
\captionsetup[subfigure]{labelformat=empty}
  \vspace{-12pt}
  \centering
  \subfloat[Filter 013] {\includegraphics[width=0.48\linewidth]{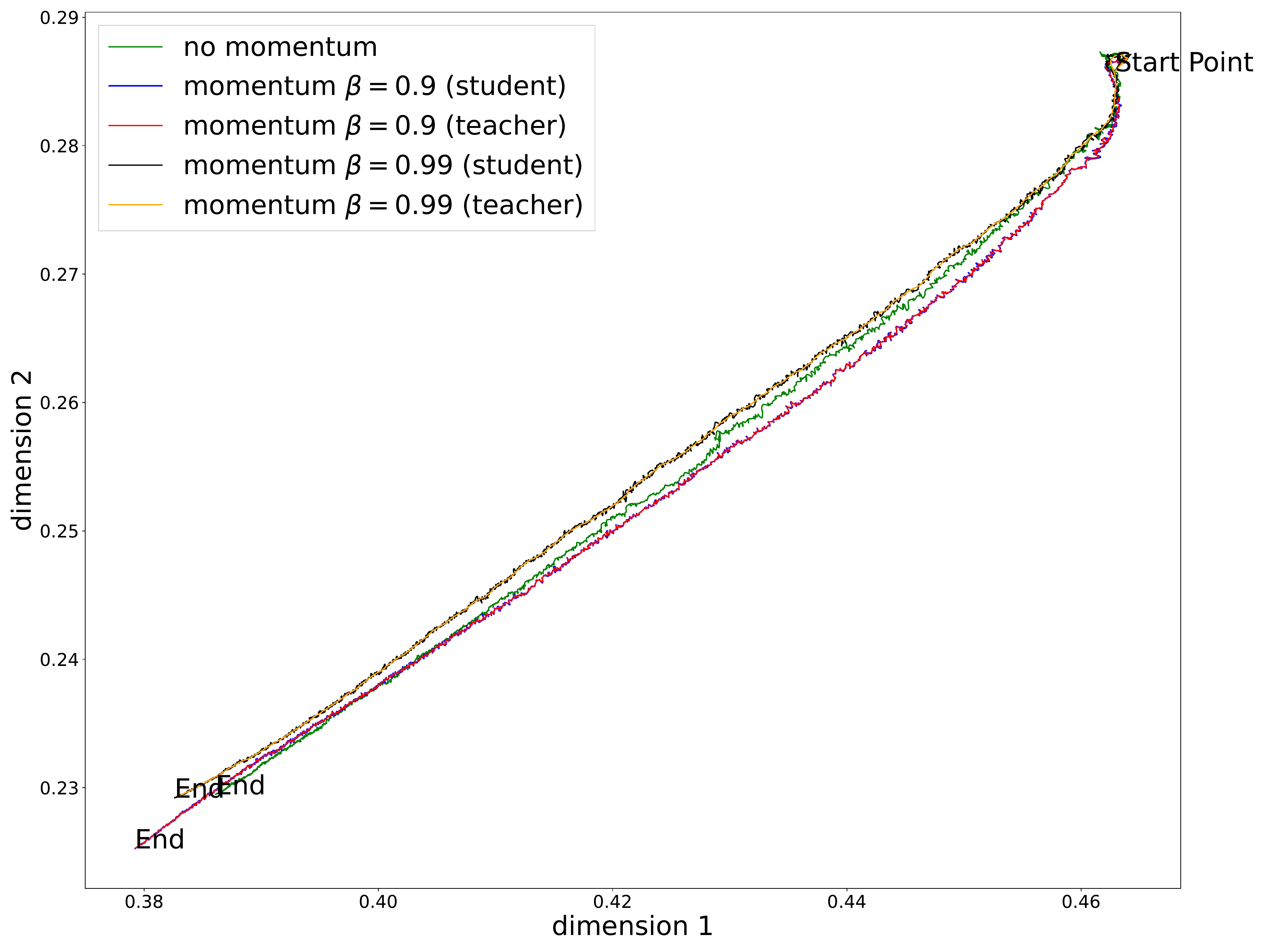}}
  \hfill
  \subfloat[Filter 014] {\includegraphics[width=0.48\linewidth]{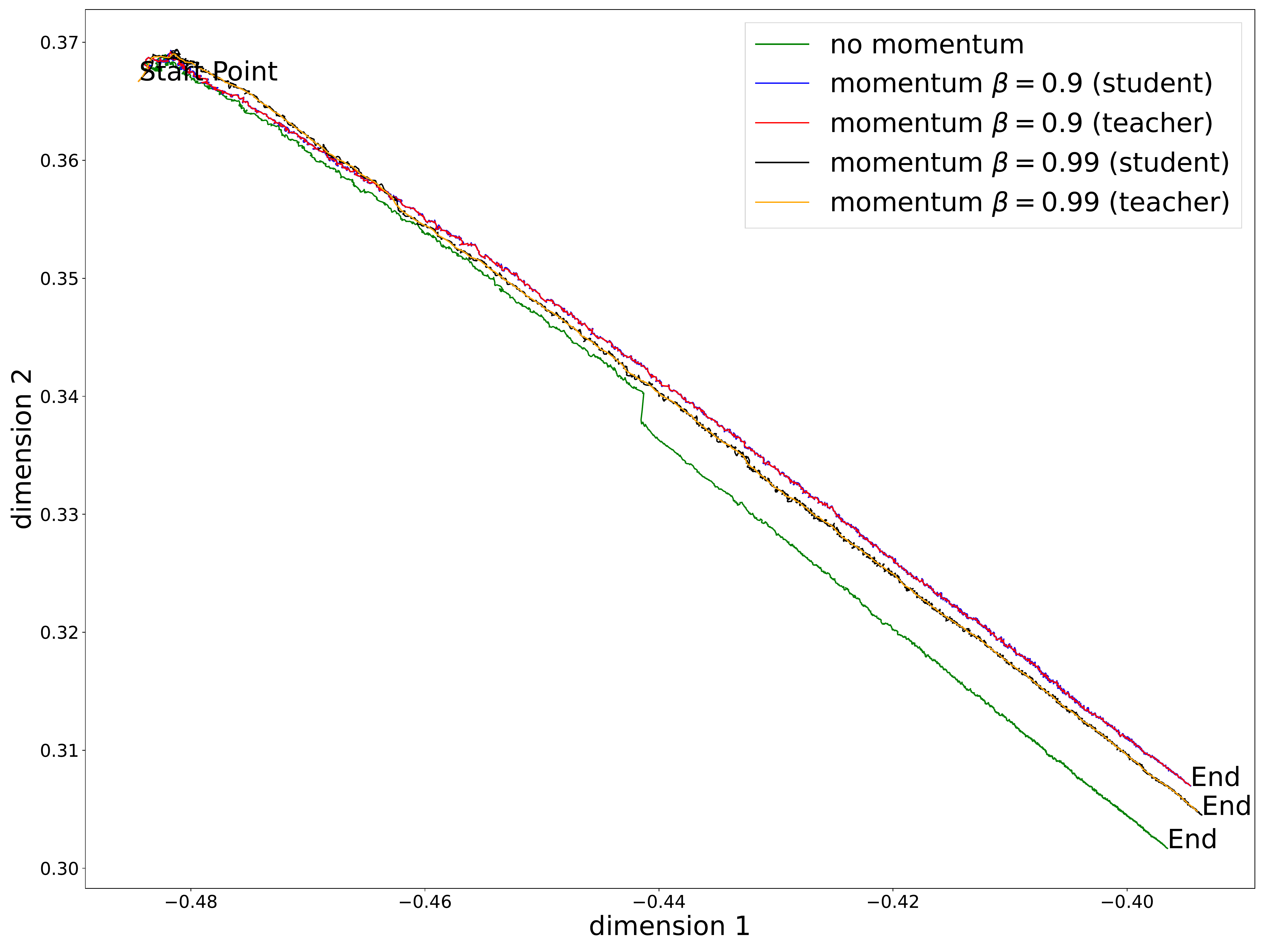}}
  \caption{}
  \vspace{-10pt}
\end{figure}

\begin{figure}[!htbp]
\captionsetup[subfigure]{labelformat=empty}
  \vspace{-12pt}
  \centering
  \subfloat[Filter 015] {\includegraphics[width=0.48\linewidth]{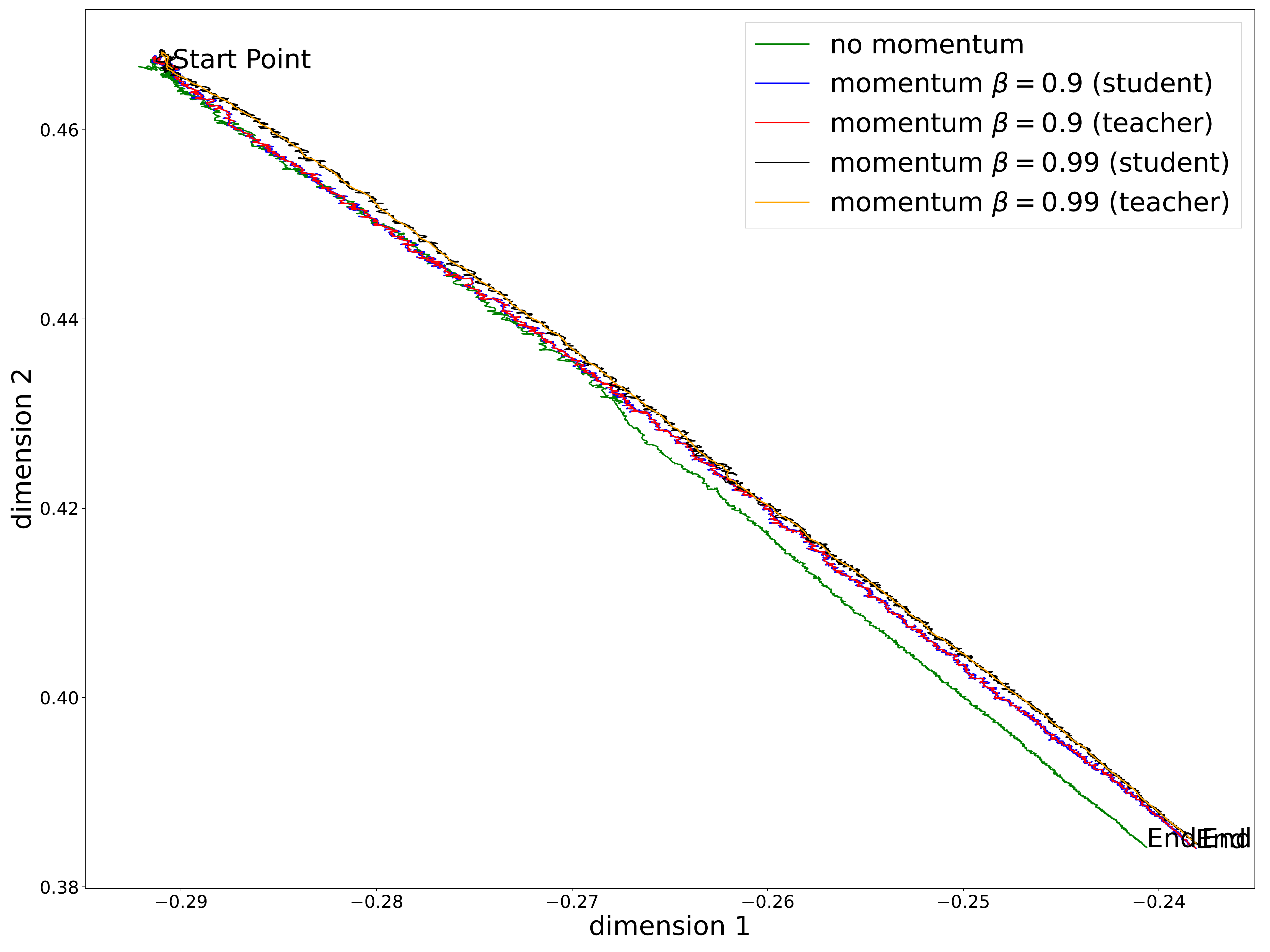}}
  \hfill
  \subfloat[Filter 016] {\includegraphics[width=0.48\linewidth]{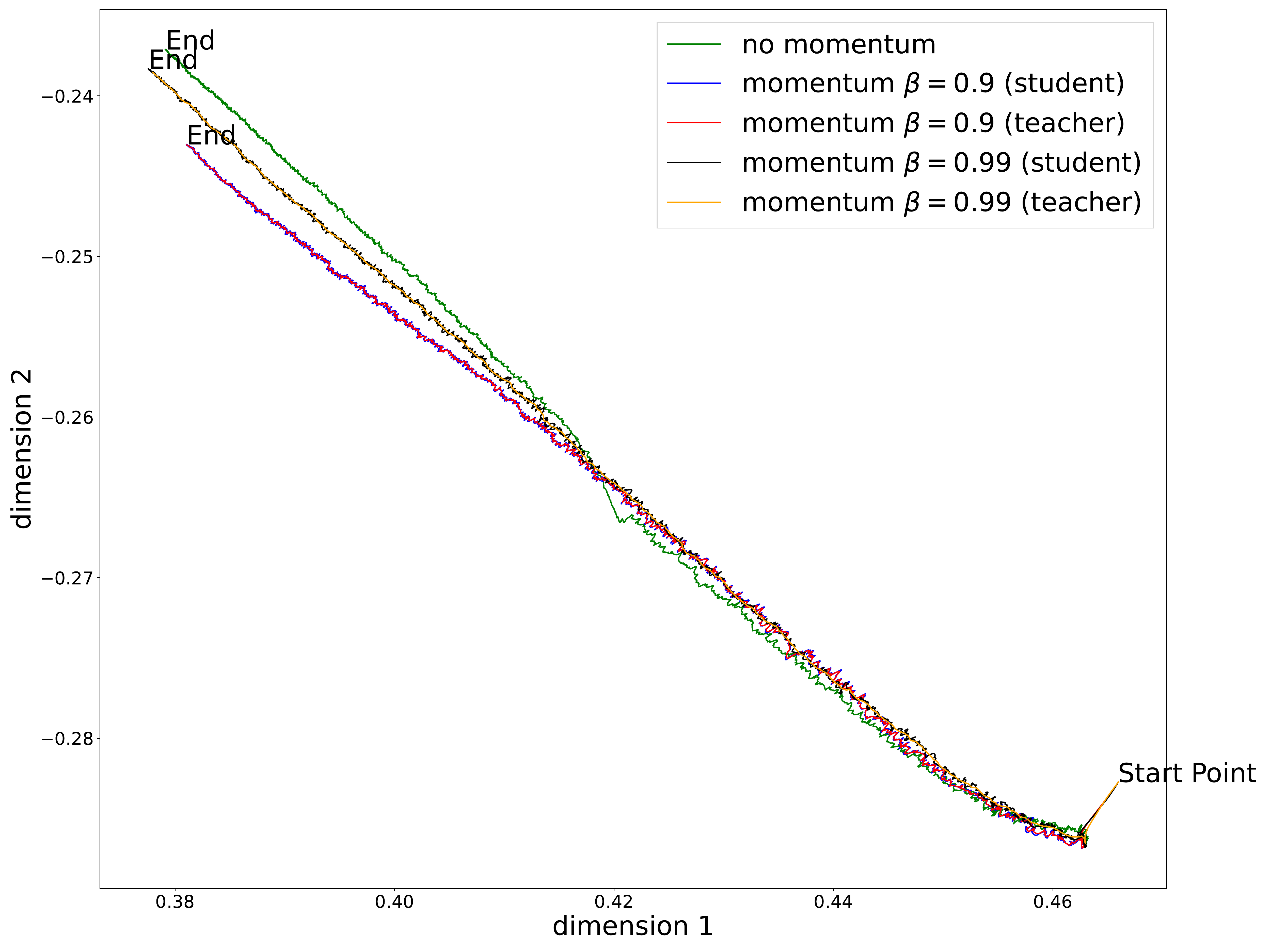}}
  \caption{}
  \vspace{-10pt}
\end{figure}

\bibliographystyle{elsarticle} 


\bibliography{elsarticle}

\begin{thebibliography}{10}
\expandafter\ifx\csname url\endcsname\relax
  \def\url#1{\texttt{#1}}\fi
\expandafter\ifx\csname urlprefix\endcsname\relax\def\urlprefix{URL }\fi
\expandafter\ifx\csname href\endcsname\relax
  \def\href#1#2{#2} \def\path#1{#1}\fi

\bibitem{brown2020language}
T.~Brown, B.~Mann, N.~Ryder, M.~Subbiah, J.~D. Kaplan, P.~Dhariwal,
  A.~Neelakantan, P.~Shyam, G.~Sastry, A.~Askell, et~al., Language models are
  few-shot learners, Advances in neural information processing systems 33
  (2020) 1877--1901.

\bibitem{Lan2020ALBERT}
Z.~Lan, M.~Chen, S.~Goodman, K.~Gimpel, P.~Sharma, R.~Soricut, Albert: A lite
  bert for self-supervised learning of language representations, in:
  International Conference on Learning Representations, 2020.

\bibitem{radford2019language}
A.~Radford, J.~Wu, R.~Child, D.~Luan, D.~Amodei, I.~Sutskever, Language models
  are unsupervised multitask learners, OpenAI blog (2019).

\bibitem{su2020vlbert}
W.~Su, X.~Zhu, Y.~Cao, B.~Li, L.~Lu, F.~Wei, J.~Dai, {\{}VL{\}}-{\{}bert{\}}:
  Pre-training of generic visual-linguistic representations, in: International
  Conference on Learning Representations, 2020.

\bibitem{nie2020dc}
P.~Nie, Y.~Zhang, X.~Geng, A.~Ramamurthy, L.~Song, D.~Jiang, Dc-bert:
  Decoupling question and document for efficient contextual encoding, in:
  Proceedings of the 43rd International ACM SIGIR Conference on Research and
  Development in Information Retrieval, 2020, pp. 1829--1832.

\bibitem{LIU2022108767}
J.~Liu, Z.~Qi, B.~Wang, Y.~Tian, Y.~Shi, Self-llp: Self-supervised learning
  from label proportions with self-ensemble, Pattern Recognition 129 (2022)
  108767.
\newblock \href {https://doi.org/https://doi.org/10.1016/j.patcog.2022.108767}
  {\path{doi:https://doi.org/10.1016/j.patcog.2022.108767}}.

\bibitem{ZHANG2022108234}
X.~Zhang, J.~Mu, X.~Zhang, H.~Liu, L.~Zong, Y.~Li, Deep anomaly detection with
  self-supervised learning and adversarial training, Pattern Recognition 121
  (2022) 108234.
\newblock \href {https://doi.org/https://doi.org/10.1016/j.patcog.2021.108234}
  {\path{doi:https://doi.org/10.1016/j.patcog.2021.108234}}.

\bibitem{komodakis2018unsupervised}
N.~Komodakis, S.~Gidaris, Unsupervised representation learning by predicting
  image rotations, in: International Conference on Learning Representations
  (ICLR), 2018.

\bibitem{oord2018representation}
A.~v.~d. Oord, Y.~Li, O.~Vinyals, Representation learning with contrastive
  predictive coding, arXiv preprint arXiv:1807.03748 (2018).

\bibitem{he2020momentum}
K.~He, H.~Fan, Y.~Wu, S.~Xie, R.~Girshick, Momentum contrast for unsupervised
  visual representation learning, in: IEEE/CVF Conference on Computer Vision
  and Pattern Recognition (CVPR), 2020.

\bibitem{chen2020simple}
T.~Chen, S.~Kornblith, M.~Norouzi, G.~Hinton, A simple framework for
  contrastive learning of visual representations, in: International Conference
  on Machine Learning, 2020.

\bibitem{madaan2022representational}
D.~Madaan, J.~Yoon, Y.~Li, Y.~Liu, S.~J. Hwang,
  \href{https://openreview.net/forum?id=9Hrka5PA7LW}{Representational
  continuity for unsupervised continual learning}, in: International Conference
  on Learning Representations, 2022.
\newline\urlprefix\url{https://openreview.net/forum?id=9Hrka5PA7LW}

\bibitem{Vasudeva_2021_ICCV}
B.~Vasudeva, P.~Deora, S.~Bhattacharya, U.~Pal, S.~Chanda, Loop: Looking for
  optimal hard negative embeddings for deep metric learning, in: Proceedings of
  the IEEE/CVF International Conference on Computer Vision (ICCV), 2021, pp.
  10634--10643.

\bibitem{zhang2022how}
C.~Zhang, K.~Zhang, C.~Zhang, T.~X. Pham, C.~D. Yoo, I.~S. Kweon, How does
  simsiam avoid collapse without negative samples? a unified understanding with
  self-supervised contrastive learning, in: International Conference on
  Learning Representations, 2022.

\bibitem{zhang2022dual}
C.~Zhang, K.~Zhang, T.~X. Pham, A.~Niu, Z.~Qiao, C.~D. Yoo, I.~S. Kweon, Dual
  temperature helps contrastive learning without many negative samples: Towards
  understanding and simplifying moco, arXiv preprint arXiv:2203.17248 (2022).

\bibitem{BAYKAL2022108244}
G.~Baykal, F.~Ozcelik, G.~Unal, Exploring deshufflegans in self-supervised
  generative adversarial networks, Pattern Recognition 122 (2022) 108244.
\newblock \href {https://doi.org/https://doi.org/10.1016/j.patcog.2021.108244}
  {\path{doi:https://doi.org/10.1016/j.patcog.2021.108244}}.

\bibitem{chen2020improved}
X.~Chen, H.~Fan, R.~Girshick, K.~He, Improved baselines with momentum
  contrastive learning, moco v2, arXiv preprint arXiv:2003.04297 (2020).

\bibitem{chen2021mocov3}
X.~Chen, S.~Xie, K.~He, An empirical study of training self-supervised vision
  transformers, Proceedings of the IEEE/CVF International Conference on
  Computer Vision (ICCV) (2021).

\bibitem{grill2020bootstrap}
J.-B. Grill, F.~Strub, F.~Altch{\'e}, C.~Tallec, P.~Richemond, E.~Buchatskaya,
  C.~Doersch, B.~Avila~Pires, Z.~Guo, M.~Gheshlaghi~Azar, et~al., Bootstrap
  your own latent-a new approach to self-supervised learning, Advances in
  Neural Information Processing Systems 33 (2020) 21271--21284.

\bibitem{caron2021emerging}
M.~Caron, H.~Touvron, I.~Misra, H.~J{\'e}gou, J.~Mairal, P.~Bojanowski,
  A.~Joulin, Emerging properties in self-supervised vision transformers, in:
  Proceedings of the IEEE/CVF International Conference on Computer Vision,
  2021, pp. 9650--9660.

\bibitem{zheng2021ressl}
M.~Zheng, S.~You, F.~Wang, C.~Qian, C.~Zhang, X.~Wang, C.~Xu, Re{SSL}:
  Relational self-supervised learning with weak augmentation, in: Thirty-Fifth
  Conference on Neural Information Processing Systems, 2021.

\bibitem{tarvainen2017mean}
A.~Tarvainen, H.~Valpola, Mean teachers are better role models: Weight-averaged
  consistency targets improve semi-supervised deep learning results, Advances
  in neural information processing systems 30 (2017).

\bibitem{chen2021exploring}
X.~Chen, K.~He, Exploring simple siamese representation learning, in: IEEE/CVF
  Conference on Computer Vision and Pattern Recognition (CVPR), 2021.

\bibitem{caron2020unsupervised}
M.~Caron, I.~Misra, J.~Mairal, P.~Goyal, P.~Bojanowski, A.~Joulin, Unsupervised
  learning of visual features by contrasting cluster assignments, Advances in
  Neural Information Processing Systems 33 (2020) 9912--9924.

\bibitem{NEURIPS2019_pytorch}
A.~Paszke, S.~Gross, F.~Massa, A.~Lerer, J.~Bradbury, G.~Chanan, T.~Killeen,
  Z.~Lin, N.~Gimelshein, L.~Antiga, A.~Desmaison, A.~Kopf, E.~Yang, Z.~DeVito,
  M.~Raison, A.~Tejani, S.~Chilamkurthy, B.~Steiner, L.~Fang, J.~Bai,
  S.~Chintala, Pytorch: An imperative style, high-performance deep learning
  library, in: H.~Wallach, H.~Larochelle, A.~Beygelzimer, F.~d\textquotesingle
  Alch\'{e}-Buc, E.~Fox, R.~Garnett (Eds.), Advances in Neural Information
  Processing Systems 32, Curran Associates, Inc., 2019, pp. 8024--8035.

\bibitem{turrisi2021sololearn}
V.~G.~T. da~Costa, E.~Fini, M.~Nabi, N.~Sebe, E.~Ricci,
  \href{http://jmlr.org/papers/v23/21-1155.html}{solo-learn: A library of
  self-supervised methods for visual representation learning}, Journal of
  Machine Learning Research 23~(56) (2022) 1--6.
\newline\urlprefix\url{http://jmlr.org/papers/v23/21-1155.html}

\bibitem{he2016deep}
K.~He, X.~Zhang, S.~Ren, J.~Sun, Deep residual learning for image recognition,
  in: IEEE/CVF Conference on Computer Vision and Pattern Recognition (CVPR),
  2016.

\bibitem{santurkar2018does}
S.~Santurkar, D.~Tsipras, A.~Ilyas, A.~Madry, How does batch normalization help
  optimization?, in: NeurIPS, 2018.

\bibitem{krizhevsky2009learning}
A.~Krizhevsky, et~al., Learning multiple layers of features from tiny images, .
  (2009).

\bibitem{tian2019contrastive}
Y.~Tian, D.~Krishnan, P.~Isola, Contrastive representation distillation, in:
  International Conference on Learning Representations, 2019.

\bibitem{krizhevsky2012imagenet}
A.~Krizhevsky, I.~Sutskever, G.~E. Hinton, Imagenet classification with deep
  convolutional neural networks, in: Advances in Neural Information Processing
  Systems, 2012.

\bibitem{tian2020contrastive}
Y.~Tian, D.~Krishnan, P.~Isola, Contrastive multiview coding, in: Proceedings
  of the European conference on computer vision (ECCV), Springer, 2020, pp.
  776--794.

\bibitem{klinker2011exponential}
F.~Klinker, Exponential moving average versus moving exponential average,
  Mathematische Semesterberichte 58~(1) (2011) 97--107.

\bibitem{appel2005technical}
G.~Appel, Technical analysis: power tools for active investors, FT Press, 2005.

\bibitem{pring2002technical}
M.~J. Pring, Technical analysis explained: The successful investor's guide to
  spotting investment trends and turning points, McGraw-Hill Professional,
  2002.

\bibitem{kingma2014adam}
D.~P. Kingma, J.~Ba, Adam: A method for stochastic optimization, in:
  International Conference on Learning Representations, 2015.

\bibitem{sutskever2013importance}
I.~Sutskever, J.~Martens, G.~Dahl, G.~Hinton, On the importance of
  initialization and momentum in deep learning, in: International conference on
  machine learning, PMLR, 2013, pp. 1139--1147.

\bibitem{ma2018quasi}
J.~Ma, D.~Yarats, Quasi-hyperbolic momentum and adam for deep learning, in:
  International Conference on Learning Representations, 2018.

\bibitem{vieillard2020momentum}
N.~Vieillard, B.~Scherrer, O.~Pietquin, M.~Geist, Momentum in reinforcement
  learning, in: International Conference on Artificial Intelligence and
  Statistics, PMLR, 2020, pp. 2529--2538.

\bibitem{korkmaz2020nesterov}
E.~Korkmaz, Nesterov momentum adversarial perturbations in the deep
  reinforcement learning domain, in: International Conference on Machine
  Learning, ICML 2020, Inductive Biases, Invariances and Generalization in
  Reinforcement Learning Workshop, Vol.~1, 2020.

\bibitem{haarnoja2018soft}
T.~Haarnoja, A.~Zhou, P.~Abbeel, S.~Levine, Soft actor-critic: Off-policy
  maximum entropy deep reinforcement learning with a stochastic actor, in:
  International conference on machine learning, PMLR, 2018, pp. 1861--1870.

\bibitem{lee2022prototypical}
K.~Lee, \href{https://openreview.net/forum?id=8la28hZOwug}{Prototypical
  contrastive predictive coding}, in: International Conference on Learning
  Representations, 2022.
\newline\urlprefix\url{https://openreview.net/forum?id=8la28hZOwug}

\bibitem{chen2020bigSimclrv2}
T.~Chen, S.~Kornblith, K.~Swersky, M.~Norouzi, G.~E. Hinton, Big
  self-supervised models are strong semi-supervised learners, Advances in
  neural information processing systems 33 (2020) 22243--22255.

\bibitem{cai2021exponential}
Z.~Cai, A.~Ravichandran, S.~Maji, C.~Fowlkes, Z.~Tu, S.~Soatto, Exponential
  moving average normalization for self-supervised and semi-supervised
  learning, in: Proceedings of the IEEE/CVF Conference on Computer Vision and
  Pattern Recognition, 2021, pp. 194--203.

\bibitem{li2021momentum}
Z.~Li, S.~Liu, J.~Sun, Momentum\^{} 2 teacher: Momentum teacher with momentum
  statistics for self-supervised learning, arXiv preprint arXiv:2101.07525
  (2021).

\bibitem{wang2022importance}
X.~Wang, H.~Fan, Y.~Tian, D.~Kihara, X.~Chen, On the importance of asymmetry
  for siamese representation learning, arXiv preprint arXiv:2204.00613 (2022).

\bibitem{von2021self}
J.~Von~K{\"u}gelgen, Y.~Sharma, L.~Gresele, W.~Brendel, B.~Sch{\"o}lkopf,
  M.~Besserve, F.~Locatello, Self-supervised learning with data augmentations
  provably isolates content from style, Advances in Neural Information
  Processing Systems 34 (2021).

\bibitem{PAIXAO2020107535}
T.~M. Paixão, R.~F. Berriel, M.~C. Boeres, A.~L. Koerich, C.~Badue, A.~F. {De
  Souza}, T.~Oliveira-Santos, Self-supervised deep reconstruction of mixed
  strip-shredded text documents, Pattern Recognition 107 (2020) 107535.
\newblock \href {https://doi.org/https://doi.org/10.1016/j.patcog.2020.107535}
  {\path{doi:https://doi.org/10.1016/j.patcog.2020.107535}}.

\bibitem{Eun_2020_CVPR}
H.~Eun, J.~Moon, J.~Park, C.~Jung, C.~Kim, Learning to discriminate information
  for online action detection, in: IEEE/CVF Conference on Computer Vision and
  Pattern Recognition (CVPR), 2020.

\bibitem{Zhuang_2020_CVPR}
C.~Zhuang, T.~She, A.~Andonian, M.~S. Mark, D.~Yamins, Unsupervised learning
  from video with deep neural embeddings, in: IEEE/CVF Conference on Computer
  Vision and Pattern Recognition (CVPR), 2020.

\bibitem{Wu_2021_CVPR}
H.~Wu, Y.~Qu, S.~Lin, J.~Zhou, R.~Qiao, Z.~Zhang, Y.~Xie, L.~Ma, Contrastive
  learning for compact single image dehazing, in: Proceedings of the IEEE/CVF
  Conference on Computer Vision and Pattern Recognition (CVPR), 2021, pp.
  10551--10560.

\bibitem{Pan_2021_CVPR}
T.~Pan, Y.~Song, T.~Yang, W.~Jiang, W.~Liu, Videomoco: Contrastive video
  representation learning with temporally adversarial examples, in: Proceedings
  of the IEEE/CVF Conference on Computer Vision and Pattern Recognition (CVPR),
  2021, pp. 11205--11214.

\bibitem{Wang_2021_CVPR}
F.~Wang, H.~Liu, Understanding the behaviour of contrastive loss, in:
  Proceedings of the IEEE/CVF Conference on Computer Vision and Pattern
  Recognition (CVPR), 2021, pp. 2495--2504.

\bibitem{Hu_2021_ICCV}
H.~Hu, J.~Cui, L.~Wang, Region-aware contrastive learning for semantic
  segmentation, in: Proceedings of the IEEE/CVF International Conference on
  Computer Vision (ICCV), 2021, pp. 16291--16301.

\bibitem{Diba_2021_ICCV}
A.~Diba, V.~Sharma, R.~Safdari, D.~Lotfi, S.~Sarfraz, R.~Stiefelhagen,
  L.~Van~Gool, Vi2clr: Video and image for visual contrastive learning of
  representation, in: ICCV, 2021.

\bibitem{nozawa2021understanding}
K.~Nozawa, I.~Sato, Understanding negative samples in instance discriminative
  self-supervised representation learning, arXiv preprint arXiv:2102.06866
  (2021).

\bibitem{Zeng_2020_CVPR}
K.~Zeng, M.~Ning, Y.~Wang, Y.~Guo, Hierarchical clustering with hard-batch
  triplet loss for person re-identification, in: IEEE/CVF Conference on
  Computer Vision and Pattern Recognition (CVPR), 2020.

\bibitem{ZHANG2022108784}
Z.~Zhang, J.~Sun, Y.~Dai, D.~Zhou, X.~Song, M.~He, Self-supervised rigid
  transformation equivariance for accurate 3d point cloud registration, Pattern
  Recognition 130 (2022) 108784.
\newblock \href {https://doi.org/https://doi.org/10.1016/j.patcog.2022.108784}
  {\path{doi:https://doi.org/10.1016/j.patcog.2022.108784}}.

\bibitem{iscen2018mining}
A.~Iscen, G.~Tolias, Y.~Avrithis, O.~Chum, Mining on manifolds: Metric learning
  without labels, in: Proceedings of the IEEE Conference on Computer Vision and
  Pattern Recognition, 2018, pp. 7642--7651.

\bibitem{chuang2020debiased}
C.-Y. Chuang, J.~Robinson, Y.-C. Lin, A.~Torralba, S.~Jegelka, Debiased
  contrastive learning, in: NeurIPS, 2020.

\bibitem{wu2020conditional}
M.~Wu, M.~Mosse, C.~Zhuang, D.~Yamins, N.~Goodman, Conditional negative
  sampling for contrastive learning of visual representations, in:
  International Conference on Learning Representations, 2020.

\bibitem{NEURIPS2020_f7cade80}
Y.~Kalantidis, M.~B. Sariyildiz, N.~Pion, P.~Weinzaepfel, D.~Larlus, Hard
  negative mixing for contrastive learning, in: H.~Larochelle, M.~Ranzato,
  R.~Hadsell, M.~F. Balcan, H.~Lin (Eds.), Advances in Neural Information
  Processing Systems, Vol.~33, Curran Associates, Inc., 2020, pp. 21798--21809.

\bibitem{robinson2021contrastive}
J.~D. Robinson, C.-Y. Chuang, S.~Sra, S.~Jegelka, Contrastive learning with
  hard negative samples, in: International Conference on Learning
  Representations, 2021.

\bibitem{kaku2021intermediate}
A.~Kaku, S.~Upadhya, N.~Razavian, Intermediate layers matter in momentum
  contrastive self supervised learning, Advances in Neural Information
  Processing Systems 34 (2021).

\bibitem{wang2020DenseCL}
X.~Wang, R.~Zhang, C.~Shen, T.~Kong, L.~Li, Dense contrastive learning for
  self-supervised visual pre-training, in: Proc. IEEE Conf. Computer Vision and
  Pattern Recognition (CVPR), 2021.

\bibitem{tian2021divide}
Y.~Tian, O.~J. Henaff, A.~van~den Oord, Divide and contrast: Self-supervised
  learning from uncurated data, in: Proceedings of the IEEE/CVF International
  Conference on Computer Vision, 2021, pp. 10063--10074.

\bibitem{ermolov2021whitening}
A.~Ermolov, A.~Siarohin, E.~Sangineto, N.~Sebe, Whitening for self-supervised
  representation learning, in: International Conference on Machine Learning,
  PMLR, 2021.

\bibitem{zbontar2021barlow}
J.~Zbontar, L.~Jing, I.~Misra, Y.~LeCun, S.~Deny, Barlow twins: Self-supervised
  learning via redundancy reduction, International Conference on Machine
  Learning (2021).

\end{thebibliography}

\end{document}